\documentclass[11pt,a4paper,abstract=on]{scrartcl}

\usepackage[utf8]{inputenc}
\usepackage[english]{babel}

\usepackage[paper=a4paper,left=25mm,right=25mm,top=25mm,bottom=30mm]{geometry}
\usepackage{amsmath,amssymb,amsthm,mathrsfs,bbm,amsfonts,bm}
\usepackage{color,graphicx,cases,caption,enumerate}
\usepackage{float,multicol,authblk}
\usepackage[dvipsnames]{xcolor}
\usepackage{colortbl}
\usepackage{booktabs}
\usepackage{multirow}
\usepackage{subcaption}
\usepackage{algorithmic}
\usepackage[linesnumbered,ruled]{algorithm2e}
\usepackage{setspace}
\usepackage{booktabs}
\usepackage{bbm}
\usepackage{capt-of}
\usepackage{array}
\usepackage[title]{appendix}
\newcolumntype{C}[1]{>{\centering\let\newline\\\arraybackslash\hspace{0pt}}m{#1}}
\usepackage{natbib}
\usepackage{placeins} 
\usepackage[breaklinks, colorlinks=true, citecolor=black, urlcolor=black, linkcolor=black]{hyperref}
\usepackage{comment}

\title{Learning low-dimensional representations of ensemble forecast fields using autoencoder-based methods}

\author[1]{Jieyu Chen}
\author[2]{Kevin Höhlein}
\author[1,3]{Sebastian Lerch}
\affil[1]{Institute of Statistics, Karlsruhe Institute of Technology}
\affil[2]{TUM School of Computation, Information and Technology, Technical University of Munich}
\affil[3]{Heidelberg Institute for Theoretical Studies}

\date{\today}

\begin{document}

\maketitle

\begin{abstract}
\noindent

Large-scale numerical simulations often produce high-dimensional gridded data that is challenging to process for downstream applications.
A prime example is numerical weather prediction, where atmospheric processes are modeled using discrete gridded representations of the physical variables and dynamics. 
Uncertainties are assessed by running the simulations multiple times, yielding ensembles of simulated fields as a high-dimensional stochastic representation of the forecast distribution. 
The high-dimensionality and large volume of ensemble datasets imposes major computing challenges for subsequent forecasting stages. 
Data-driven dimensionality reduction techniques could help to reduce the data volume before further processing by learning meaningful and compact representations.
However, existing dimensionality reduction methods are typically designed for deterministic and single-valued inputs, and thus cannot handle ensemble data from multiple randomized simulations.
In this study, we propose novel dimensionality reduction approaches specifically tailored to the format of ensemble forecast fields. 
We present two alternative frameworks, which yield low-dimensional representations of ensemble forecasts while respecting their probabilistic character. 
The first approach derives a distribution-based representation of an input ensemble by applying standard dimensionality reduction techniques in a member-by-member fashion and merging the member representations into a joint parametric distribution model. 
The second approach achieves a similar representation by encoding all members jointly using a tailored variational autoencoder. 
We evaluate and compare both approaches in a case study using 10 years of temperature and wind speed forecasts over Europe. 
The approaches preserve key spatial and statistical characteristics of the ensemble and enable probabilistic reconstructions of the forecast fields.

\end{abstract}

\section{Introduction}\label{intro}

Large-scale physics-based models are used across environmental sciences for prediction and modeling. 
A particularly important example is numerical weather prediction (NWP) models, where atmospheric processes are represented via partial differential equations. 
The forecast quality of NWP models has improved tremendously in recent decades due to continued scientific and technological advances \citep{bauer2015the}.
Nowadays, NWP models are often run in an ensemble mode to quantify forecast uncertainty.
Thereby, a collection of predictions of future weather states is obtained by running the model several times with varying initial conditions and perturbed model physics.
Conceptually, these ensemble members can be considered equally likely realizations of an unknown probability distribution. 

Due to their continuously increasing spatial and temporal resolution, ensemble weather forecasting models produce large amounts of data. 
However, such high-dimensional and complex data can be challenging to process in applications relying on weather predictions as inputs.
Examples include weather forecasting applications such as post-processing and analog forecasting, and downstream applications such as hydrological and energy forecasting models.
Therefore, summarizing relevant information from meteorological input data across space and time via learning low-dimensional representations is of interest beyond just reducing the amount of data that needs to be stored.

One example is ensemble post-processing, which aims at correcting systematic errors of NWP ensemble predictions via statistical or machine learning (ML) models \citep{vannitsem2021statistical}. Post-processing models use ensemble predictions of relevant meteorological variables as inputs, and produce corrected probabilistic forecasts in the form of probability distributions as their output.
While recent ML-based approaches have enabled the incorporation of many predictor variables \citep{rasp2018neural,schulz2022machine,chen2024generative} 
and there exist first spatial post-processing approaches \citep{Scheuerer2020,gronquist2021,Veldkamp2021wind,Chapman2022,HoratLerch2024}, most post-processing models still tend to operate on localized predictions at individual stations or grid point locations.
However, the restriction to localized predictions prevents the incorporation of predictability information from large-scale spatial structures, including flow-dependent error characteristics and weather regimes, which are inherent in physically consistent forecast fields \citep{rodwell2018flow,AllenEtAl2021}.
To address this limitation, \citet{lerch2022convolutional} propose the use of convolutional autoencoders to learn low-dimensional latent representations of the spatial forecast fields and demonstrate that using the learned representations as additional predictors to augment a NN-based post-processing model with information about the spatial structure of relevant forecast fields helps to improve predictive performance. 
However, \citet{lerch2022convolutional} only utilize learned representation of the mean ensemble field, where all ensemble member forecasts are averaged at every grid point.
One potential drawback is that the mean field will be notably smoother than forecast fields from individual members. 
More importantly, however, such approaches ignore the underlying probabilistic information available in the ensemble simulations, which can be seen as samples from a multivariate probability distribution.

Our overarching aim is to propose dimensionality reduction methods to learn low-dimensional representations of ensemble forecast fields, which respect the inherently probabilistic nature of the input data.
A variety of dimensionality reduction methods are available, ranging from classical principal component analysis \citep[PCA;][]{pearson1901pca,jolliffe2016principal} to neural network (NN)-based autoencoder (AE) methods \citep{bourlard1988auto, kramer1991nonlinear, hinton1993autoencoders, hinton2006reducing}.
However, the application of existing dimensionality reduction methods to ensemble forecast fields is not straightforward, since they tend to be tailored to deterministic input data.
To the best of our knowledge, the problem of learning representations of ensemble simulation data has not been considered thus far, potentially since this type of data is somewhat specific to environmental modeling.
The key challenge thus is to develop dimensionality reduction approaches that learn distributional representations in the latent space for an ensemble of forecast fields, allowing for random samples drawn from the latent distribution to be mapped back to a reconstructed forecast field, which ideally would be indistinguishable from a random member from the input ensemble.

To achieve this, we propose two approaches, one based on existing dimensionality reduction methods and one utilizing variational autoencoder \citep[VAE;][]{kingma2022autoencoding} architectures.
The former is an extension of existing dimensionality reduction models with deterministic latent code and can be summarized as a two-step framework.
In the first step, a dimensionality reduction model (e.g., PCA or an AE model) is employed to learn low-dimensional representations for each member of the ensemble forecast fields.
In the second step, a multivariate Gaussian distribution in the latent space is fitted to the learned representations of all ensemble members.
This distribution serves as a learned probabilistic representation of the entire ensemble and can be used to reconstruct ensemble members that are statistically indistinguishable from the original members.
This is achieved by reverting the encoding process, i.e., drawing independent samples from the fitted distribution and applying the reverse step of the dimensionality reduction model (e.g., inverse PCA transform or the decoder of AE). 
In the latter VAE-based framework, on the other hand, we utilize a tailored VAE model that jointly considers all ensemble members as input and provides a distributional ensemble representation as the encoder posterior distribution defined on the VAE's latent space.
A key design consideration is that the proposed VAE model should respect the interpretation of ensemble members as interchangeable samples from an unknown, multivariate probability distribution. 
Notably, the obtained distributional representation should be independent of any (arbitrary) ordering in which the ensemble members are sampled, held in memory, and supplied to the VAE. 
To this end, we use an invariant VAE (iVAE) architecture designed to be invariant to the reordering of ensemble members.

We systematically compare the two approaches in two case studies on ensemble forecast fields covering a region that roughly corresponds to Europe. We focus on 2-day ahead forecasts of temperature and wind speed, utilizing 10 years of daily forecasts from the European Centre for Medium-Range Weather Forecasts (ECMWF). To that end, we discuss appropriate evaluation approaches for the problem at hand, and consider an exemplary analysis of the learned representations.

The remainder of the paper is structured as follows.
Section \ref{data} provides an overview of the dataset, and Section \ref{methods} introduces the proposed two-step and iVAE approaches to learn distributional representations of ensemble forecast fields.
The evaluation methods and main results are presented in Section \ref{results}, followed by conclusions and discussions in Section \ref{conclusion}.
Python code with implementations of all approaches is available online (\url{https://github.com/jieyu97/invariantVAE}).

\section{Data}\label{data}

We focus on daily ensemble forecasts from the ECMWF's 50-member ensemble on a spatial domain roughly covering the European continent ($-10$E to $30$E and $30$N to $70$N). The forcasts are available as gridded fields with regular $0.5^\circ \times 0.5^\circ$ resolution in latitude and longitude. This results in $81 \times 81 (= 6561)$ grid points over Europe.
The forecasts are initialized daily at 00 UTC with a forecast lead time of 48 hours. We retrieve forecast data for all days in the time period from January 3, 2007 to January 2, 2017, and split the data into non-overlapping parts for training (03.01.2007--31.12.2014), validation (01.01.2015--31.12.2015), and testing (remainder). 

For brevity, we select four exemplary meteorological variables as the basis of our evaluation\footnote{The datasets for the four variables considered are based on the underlying data from \citet{rasp2018neural}, and are available at \url{https://doi.org/10.6084/m9.figshare.28151213.v2} (\texttt{t2m}), \url{https://doi.org/10.6084/m9.figshare.28151372.v1} (\texttt{u10}), \url{https://doi.org/10.6084/m9.figshare.28151411.v1} (\texttt{v10}), \url{https://doi.org/10.6084/m9.figshare.28151444.v1} (\texttt{z500}).}: 2-m temperature (\texttt{t2m}) in Kelvin, U component of wind at 10 meters (\texttt{u10}) in meter per second, V component of wind at 10 meters (\texttt{v10}), and geopotential height at 500 hPa (\texttt{z500}).
In the main paper, we present results exclusively for t2m and u10. 
Results for the remaining variables are available in the Supplemental Material.

For each weather variable, we apply standard normalization to the raw ensemble forecast data for more stable training of neural network models.
The data is standardized by subtracting a global mean and dividing by a global standard deviation. The parameters are computed separately for each weather variable using the data from all grid points in the domain and all samples in the training dataset.

\section{Learning distributional representations of ensemble forecast fields}\label{methods}

This section first introduces required mathematical notation and outlines the problem to be addressed,
and then presents two different frameworks for learning distributional representations of ensembles of spatial fields.

\subsection{Mathematical notations and problem formulation}\label{sec_problem}

Throughout this paper, we aim to find lower-dimensional representations for ensembles of spatial forecast fields that summarize the ensemble's information and reduce the data complexity to simplify subsequent forecasting tasks. The required representations are learned in a data-driven way using suitable statistical models for encoding and decoding the inputs. Due to the stochastic characteristic of ensembles, we focus on distributional representations, which express the ensemble information through a suitably parameterized probability distribution defined on a low-dimensional latent space. The problem can thus be considered a dimensionality reduction task with low-dimensional distribution-based embeddings. The intended distributional representations encapsulate both the general spatial structure of the forecast fields and the variability among ensemble members. The field information of the original or new ensemble members can thus be reconstructed by sampling from the parametric distribution and decoding the samples.

For a specific weather variable (e.g., 2-m temperature, \texttt{t2m}) and time $t$, we denote the 50-member ensemble forecast by $\bm{X}^{\texttt{t2m}, t} = \{\bm{X}_m^{\texttt{t2m}, t}\}_{m=1}^{50}$, wherein $\bm{X}_m^{\texttt{t2m}, t} \in \mathbb{R}^{d_{\text{data}}}$ represents the $m$-th member forecast field.
The subscripts \texttt{t2m} and $t$ will typically be omitted for brevity. We then write $\bm{X} = \{\bm{X}_m\}_{m=1}^{50}$ to denote the ensemble forecast for a given variable and a given time, e.g., to refer to a data point as one training example. The proposed dimensionality reduction methods process one meteorological variable at a time.
Therefore, each forecast field $\bm{X}_m$ comprises scalar-valued forecast data for $81 \times 81$ grid locations, resulting in $d_{\text{data}} = 6561$.
The 50 ensemble members are interpreted as independent samples from an unknown but identical multivariate probability distribution $\mathcal{P}$, which captures the uncertainty about the predicted weather state:
\begin{equation*}
\bm{X}_m \sim \mathcal{P} \ \text{for } m \in \{1, \ldots, 50\}.
\end{equation*}

Each dimensionality reduction consists of an encoding part \texttt{E}, a decoding part \texttt{D}, and a latent space $\mathbb{R}^{d_{\text{latent}}}$, which hosts the learned representations. The encoding part learns to translate the input ensemble into a representative distribution $\mathcal{D}$ in the latent space, i.e.,
\begin{equation*}
    \mathcal{D} = \texttt{E}(\bm{X})\text{,}
\end{equation*}
and the decoding part is trained to reconstruct an ensemble of forecast fields $\tilde{\bm{X}} = \{\tilde{\bm{X}}_n\}_{n=1}^N$ based on an ensemble of samples $\bm{z} = \{\bm{z}_n\}_{n=1}^{N}$, drawn from $\mathcal{D}$, i.e.,
\begin{equation*}
    \tilde{\bm{X}} = \texttt{D}(\bm{z}), 
\end{equation*}
wherein $\bm{z}_n\sim \mathcal{D}$ for $n \in \{1, ..., N\}$. While $N$ -- the size of the reconstructed ensemble -- can be arbitrary, in general, we mainly consider the case $N = 50$, which matches the number of members in the original ensemble forecast. We note, however, that even in this case $\tilde{\bm{X}}_n$ and $\bm{X}_m$ usually do not correspond, even if they have the same subscript value, since $\tilde{\bm{X}}_n$ is decoded from a random sample $\bm{z}_n$ which is not necessarily the adequate latent representation of $\bm{X}_m$ with $m=n$. The reconstruction of the original ensemble will be denoted as $\hat{\bm{X}} = \{\hat{\bm{X}}_m\}_{m=1}^{50}$.

Additionally, we usually have that the latent dimension $d_{\text{latent}} \ll d_{\text{data}}$. Therefore, $d_{\text{latent}}$  controls the compression level of the methods, i.e., how much information from each ensemble field remains in the latent representations. As a hyperparameter of the proposed methods, the latent dimension can be adapted to the needs of downstream tasks. In the presented case studies, we restrict our focus to relatively low latent dimensions ranging from 2 to 32. The information content of the representation is furthermore affected by the parametric form of the latent distribution $\mathcal{D}$, which is another design choice within the proposed method. We will assume $\mathcal{D}$ to be Gaussian, i.e., $\mathcal{D} = \mathcal{N}(\bm{\mu}, \bm{\Sigma})$ with parameters $\bm{\mu} \in \mathbb{R}^{d_{\text{latent}}}$  and $\bm{\Sigma} \in \mathbb{R}^{d_{\text{latent}} \times d_{\text{latent}}}$ representing the distribution mean and covariance matrix, respectively. 

The ultimate goal of the proposed learning framework is finding suitable mappings \texttt{E} and \texttt{D} such that for multi-samples $\bm{z}$ from $\mathcal{D}$, the reconstructed field ensemble $\texttt{D}(\bm{z})$ becomes statistically indistinguishable from ensemble members sampled directly from the forecast distribution $\mathcal{P}$.
The key challenge therein lies in performing dimensionality reduction on the space of probability distributions, which are represented through stochastically sampled ensembles of forecast fields. In this setting, the representations of each data point (i.e., ensemble $\bm{X}$) only convey incomplete and stochastic information about the underlying data (i.e., forecast distribution $\mathcal{P}$).
This is in stark contrast to the assumption of standard dimensionality reduction problems, which presuppose complete and deterministic data representations. 
We propose two different approaches to address this problem, leveraging statistical and machine learning methods, which will be introduced in the following sections.

\subsection{Two-step dimensionality reduction approaches}\label{sec_ae}

Our input data are 50-member ensembles of spatial forecast fields $\bm{X} = \{\bm{X}_m\}_{m=1}^{50}$.
The most straightforward approach is to treat all members collectively as one input $\bm{X}$ and utilize existing dimensionality reduction methods.
However, treating the ensemble members jointly ignores their nature as samples drawn from an identical distribution, leading to a deterministic latent representation for the entire ensemble.
This conflicts with our goal of learning a representative low-dimensional distribution for the ensemble of forecast fields.
Furthermore, the variabilities among different ensemble members are often less distinct than those among different grid locations in the spatial forecast fields.
Consequently, the learned deterministic representation primarily captures the spatial structure in the data.
Initial experiments indicated that the resulting reconstructed ensemble forecast fields approximate only the mean field, and fail to reproduce any variability between ensemble members.

To address the probabilistic nature of the ensemble forecast fields and preserve uncertainty information, we propose a two-step framework to identify a latent distribution capturing both general spatial structure and variability within the ensemble.
This framework builds on existing methods, which are used to reduce the dimension of each ensemble member separately before merging the per-member representations into a distributional form. Specifically, we assume that a given standard dimensionality reduction approach provides a projection $f$, which maps a data item to its reduced representation, and a reconstruction function $g$, which restores a data item based on its latent code. 

To encode an ensemble, we proceed by projecting each forecast field separately to obtain its low-dimensional representations as 
\begin{equation*}
    \hat{\bm{z}}_m = f(\bm{X}_m) \quad \text{for} \ m \in \{1, \ldots, 50\}\text{.}
\end{equation*}
This yields an ensemble of latent representations to which we can fit a $d_\text{latent}$-dimensional Gaussian distribution:
\begin{equation*}
    \mathcal{D} = \mathcal{N}(\bm{\mu}, \bm{\Sigma}), \quad \text{with} \ \bm{\mu} = \frac{1}{50} \sum_{m=1}^{50} \hat{\bm{z}}_m\text{, and} \ \bm{\Sigma} = \text{Var} (\hat{\bm{z}}_1, \ldots, \hat{\bm{z}}_{50})\text{.}
\end{equation*}
Therein, $\bm{\mu}$ is the estimated mean vector and $\bm{\Sigma}$ is the estimated covariance matrix. This corresponds to the intended low-dimensional representation.

To reconstruct ensemble members, any number $N$ of samples can be drawn from $\mathcal{D}$ and utilized to generate new forecast fields. A reconstructed ensemble is obtained as
\begin{equation*}
    \tilde{\bm{X}} = \{g(\bm{z}_n)\}_{n=1}^{N}, \quad \text{with} \ \bm{z}_n \sim \mathcal{D}, \quad \text{for} \ n \in \{1, \ldots, N\}.
\end{equation*}
These newly generated forecast fields can be considered to follow the same distribution as the reconstructions of the input ensemble members, $\hat{\bm{X}}_m = g(\hat{\bm{z}}_m)$, for $m \in \{1, ..., 50\}$.

\begin{figure}
\centering
    \includegraphics[width=0.95\textwidth]{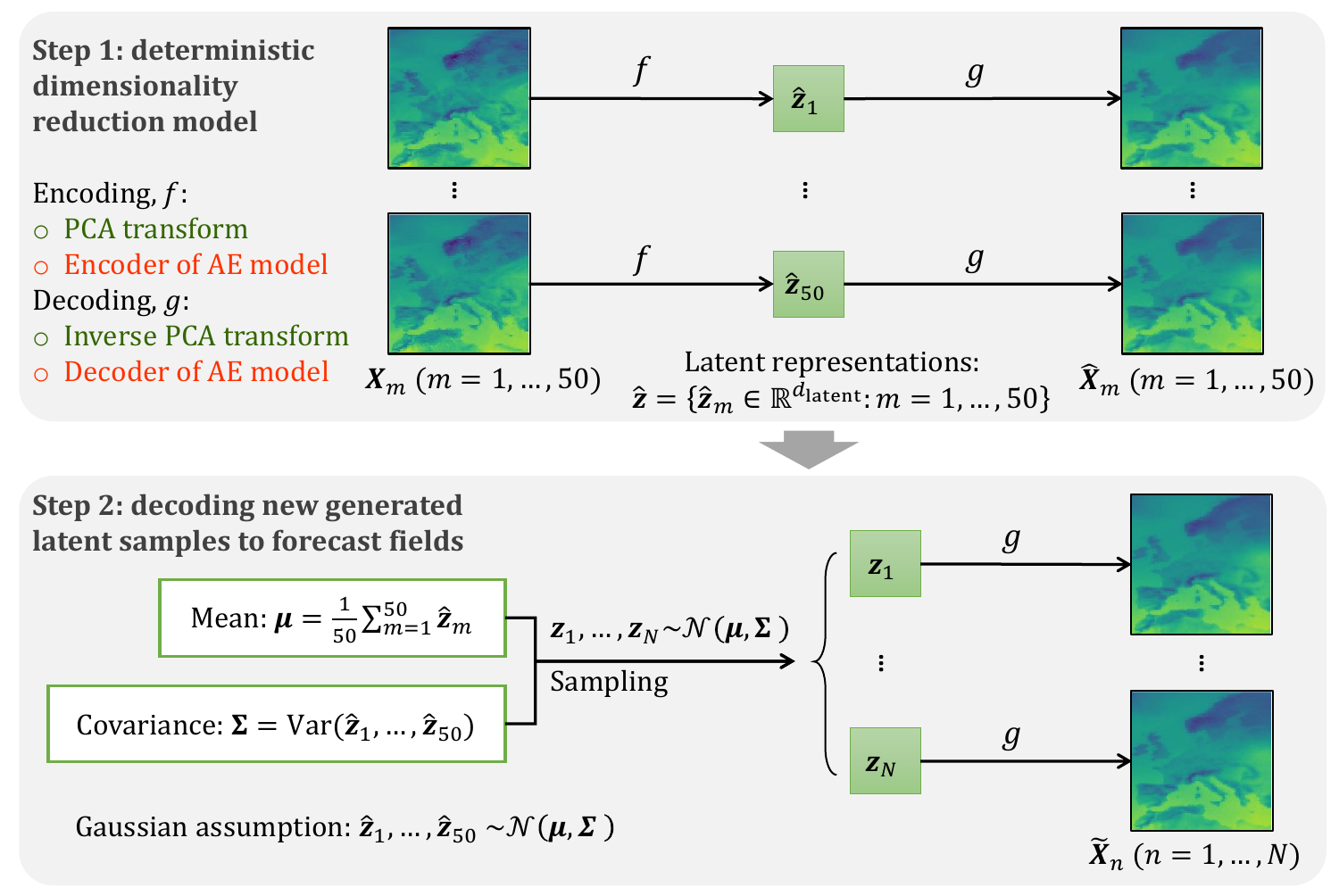}
    \caption{Schematic overview of the two-step dimensionality reduction methods based on PCA and AE models.}
    \label{fig_ae}
\end{figure}

We focus on two practical implementations of this approach using principal component analysis (PCA) and autoencoder (AE) neural networks as the underlying algorithms.
A schematic illustration of the autoencoder approach is provided in Figure \ref{fig_ae}. PCA and AE are introduced in the following sections.

\subsubsection{Principal Component Analysis approach}

Principal Component Analysis (PCA) is a linear method that finds the projections of data onto the principal components, which capture the largest variation in the data.
The basic idea is to project all samples into a new coordinate system, where the axes (principal components) are determined by the direction along which projections have the largest variance in descending order.
For a dataset of $d_{\text{data}}$ dimensions, the number of principal components is also $d_{\text{data}}$, and the dimensionality reduction is conducted by taking the projections onto only the first $d_{\text{latent}}$ principal components.
The reversibility of PCA transform allows data to be reconstructed from the low-dimensional representations.
To provide a benchmark for comparison, we employ PCA in the two-step framework, and refer to it as the PCA-based approach.

In many real-world datasets, linear transformations are not adequate to compress key information, and a variety of nonlinear dimensionality reduction techniques have been proposed.
Examples include kernel PCA \citep{schoelkopf1997kernel}, Isomap \citep{joshua2000global}, and Locally Linear Embedding \citep[LLE;][]{roweis2000nonlinear}.
While these methods provide effective low-dimensional representations, the process of converting them back to the original data space introduces additional challenges.
The inverse transformations for those nonlinear techniques often require additional training procedures, as exemplified by the pre-image problem for kernel PCA \citep{mika1998kernel, kwok2004preimage}.
Autoencoder neural network models, on the other hand, have emerged as a more flexible and now widely used alternative in reconstructing data from latent features.
Since PCA can be considered as a linear case of a simple neural network, it is a natural reference method.

\subsubsection{Autoencoder neural network approach}

Autoencoders are neural network models for unsupervised learning that aim to replicate the input as their output.
A typical autoencoder features an internal bottleneck layer with fewer nodes than the input and output layers, dividing the network into two distinct components, the encoder and the decoder, and induces an information bottleneck that prohibits the autoencoder network from memorizing every detail of the input.
The encoder $f$ and decoder $g$ can be formulated as two mappings, where, following the notations in Section \ref{sec_problem},
\begin{equation*}
    f(\bm{X}_m) = \bm{\hat{z}}_m, \quad g(\bm{\hat{z}}_m) = \hat{\bm{X}}_m, \quad \text{for } \bm{X}_m, \hat{\bm{X}}_m \in \mathbb{R}^{d_\text{data}}, \; \bm{\hat{z}}_m \in \mathbb{R}^{d_\text{latent}} \text{ and } m \in \{1, \ldots, 50\}.
\end{equation*}
The encoder network $f$ maps one input forecast field $\bm{X}_m$ to its latent representation $\bm{\hat{z}}_m$ from the bottleneck layer, with $d_\text{latent}$ typically much smaller than $d_\text{data}$.
The decoder network $g$ maps one latent representation $\bm{\hat{z}}_m$ back to the corresponding reconstruction $\hat{\bm{X}}_m$, the output of the autoencoder, which aims to reproduce the input $\bm{X}_m$.
Training autoencoder neural networks involves minimizing differences between input and output, often using mean square error as a loss function.

The latent representation $\bm{\hat{z}}_m$ obtained from the encoder naturally functions as a compact representation of the input, capturing essential features needed for the decoder to reconstruct the original data.
In addition to dimensionality reduction applications \citep{hinton2006reducing, Wang2014generalized, wang2016autoencoder}, autoencoder models have also found applications in other domains, such as anomaly detection \citep{sakurada2014anomaly, zhou2017anomaly} and image denoising \citep{gondara2016medical}.
Autoencoders exist in many variants, developed for different applications, including, e.g., sparse autoencoders for classification tasks \citep{baccouche2012spatio}.

Our autoencoder model for the AE-based dimensionality reduction approach is a shallow neural network utilizing fully connected dense layers in both the encoder and decoder.
The model is trained to minimize the mean absolute error (MAE) between the input forecast field and the reconstructed field obtained as output.
Hyperparameter tuning is performed using the Bayesian optimization algorithm HyperBand \citep{li2018hyperband} implemented in the \texttt{Ray Tune} Python library \citep{liaw2018tune}.
The final AE model configuration includes layers of sizes ``6561 - 4096 - $d_{\text{latent}}$" in the encoder and ``$d_{\text{latent}}$ - 4096 - 6561" in the decoder.
The LeakyReLU activation function is applied in the hidden layer in both the encoder and the decoder.
We utilize the AdamW optimizer \citep{loshchilov2019decoupled} with a learning rate decay scheduler starting from $1e^{-4}$ to stabilize the training process.
Mini-batch training is employed with a batch size of 1024 to enhance training efficiency, and samples in all batches are randomly shuffled in each training epoch.
To prevent overfitting, an early stopping criterion with a patience of 20 epochs on the validation loss is applied.
We also investigated more sophisticated frameworks for the encoder and decoder during initial experiments, including convolutional layers with residual blocks \citep{he2016deep}, and a vision transformer \citep[ViT;][]{dosovitskiy2021image} based architecture.
However, these more complex frameworks did not yield significant improvements and incurred higher computational costs.
Therefore, we prioritize a simpler framework with only dense layers for our neural network models, allowing efficient computation without requiring a GPU.

\subsection{Invariant variational autoencoder approach}\label{sec_ivae}

The two-step framework developed for ensemble forecast fields can, in principle, be generalized to other dimensionality reduction techniques beyond PCA and AE.
However, a conceptual disadvantage of the approach is the assumption of Gaussian-distributed representations in the latent space, which is somewhat decoupled from the training process.

To address this limitation, we propose an ensemble-invariant framework based on variational autoencoders (VAEs).
VAEs \citep{kingma2022autoencoding} are generative machine learning methods that leverage variational inference to learn a probabilistic representation in latent space.
In contrast to standard autoencoders, VAEs connect an encoder network to its decoder through a probabilistic latent space, which corresponds to the parameters of a pre-specified probability distribution.
Thereby, the encoder network maps input samples to parameters of the latent space distribution, and the decoder network maps samples drawn from the distribution in the latent space back to the data space by generating new data points decoded from the samples.
The reparameterization trick \citep{kingma2019foundations} enables the simultaneous training of the encoder and decoder using backpropagation by transforming the sampling process to make it differentiable.
VAEs have been widely applied in various domains, including image generation, denoising, and inpainting \citep{an2015variational, pu2016variational}.
Moreover, the VAE framework has inspired the development of extensions targeting different aspects of feature representations and applications.
Examples include Importance Weighted Autoencoders \citep{burda2015importance}, the combination of a VAE with a Generative Adversarial Network \citep{larsen2016autoencoding}, Wasserstein Autoencoders \citep{tolstikhin2017wasserstein} and Sinkhorn Autoencoders \citep{patrini2020sinkhorn}.

\begin{figure}
\centering
    \includegraphics[width=\textwidth]{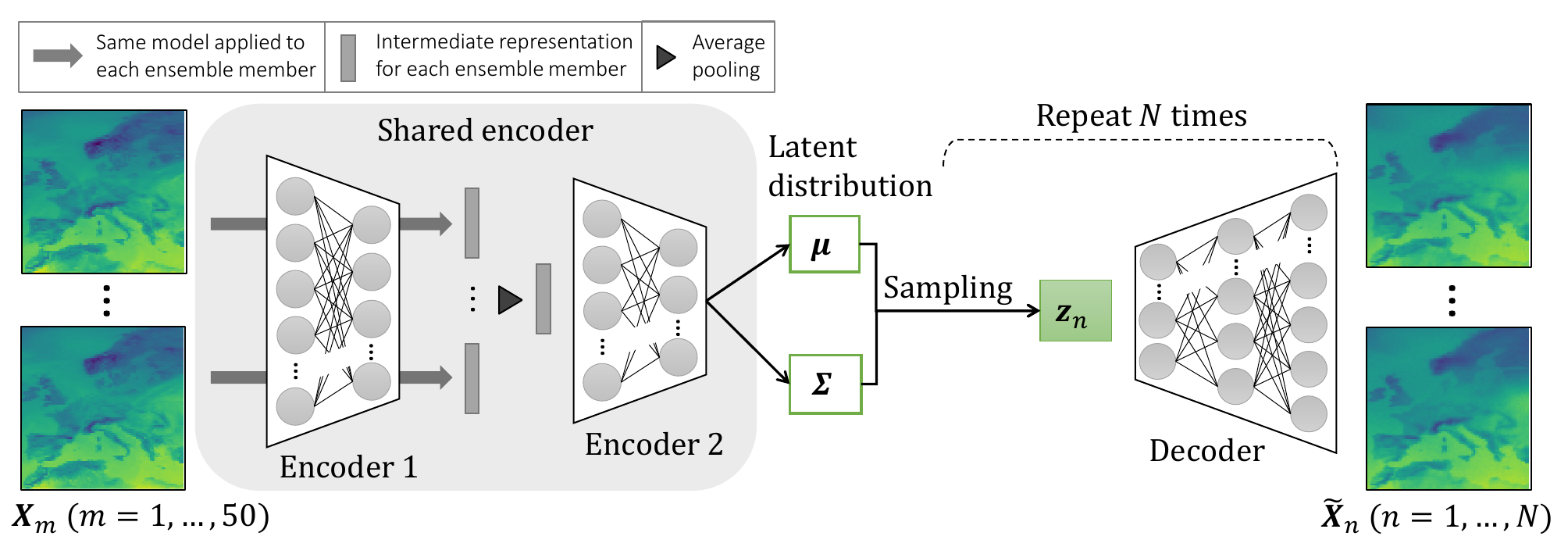}
    \caption{Schematic illustration of the invariant variational autoencoder (iVAE) model.}
    \label{fig_ivae}
\end{figure}

The inherently probabilistic nature of VAEs makes them potentially effective for the problem at hand.
The standard VAE model is trained to learn a latent distribution for a single instance from the input data, where samples drawn from the latent distribution are decoded to data points close to the corresponding instance.
If we consider each ensemble member separately, the VAE model would thus learn different latent distributions for different members from the same forecast case.
Therefore, we need to adapt the VAE framework to jointly learn one latent distribution for all ensemble members, and decode samples from the latent distribution to newly generated members that follow the same distribution as the inputs.
To address this challenge, we propose an invariant VAE (iVAE) model, inspired by the permutation-invariant neural network framework in the Deep Sets architecture \citep{zaheer2017deep}.
The encoder of our iVAE model follows such a permutation-invariant framework and is invariant to any permutation on the order of ensemble members.
A schematic overview of our iVAE model is available in Figure \ref{fig_ivae}. 

The main difference between our iVAE model and a standard VAE is that the encoder is shared for all 50 members of the ensemble forecast fields as input, and is invariant to their order.
The shared encoder comprises two separate encoder parts, which we denote by $e_1$ and $e_2$, see Figure \ref{fig_ivae}.
For a given 50-member ensemble $\bm{X} = \{\bm{X}_m\}_{m=1}^{50}$, the first encoder is applied to each ensemble member forecast field $\bm{X}_m$ iteratively to obtain intermediate representations $\bm{y} = \{\bm{y}_m\}_{m=1}^{50}$, i.e.,
\begin{equation*}
    \bm{y}_m = e_1(\bm{X}_m), \ \text{ for }  m \in \{1, \ldots, 50\}.
\end{equation*}
These intermediate representations are interchangeable since their original input fields are assumed to follow the same distribution.
Next, we average the 50 intermediate representations to summarize key features learned from all ensemble members, i.e.,
\begin{equation*}
\Bar{\bm{y}} = \frac{1}{50} \sum_{m=1}^{50} \bm{y}_m,
\end{equation*}
which ensures that the shared encoder is permutation-invariant.
Subsequently, the second encoder is applied after this average pooling step.
Similar to standard VAEs, a probabilistic encoder $e_2$ with parameters $\phi$ is applied to approximate the posterior distribution $p(z|\bm{X})$ in the latent space using a parameterized distribution $q_\phi(\bm{z} | \bm{X}) = \mathcal{N}(\bm{z}; \bm{\mu}, \mathrm{diag}(\bm{\sigma}^2))$.
After applying the reparametrization trick, we obtain
\begin{align*}
    \bm{\epsilon}_n & \sim \mathcal{N}(\bm{0}, \mathbf{I}), \\
    (\bm{\mu}, \log \bm{\sigma}) &= e_2 \left( \frac{1}{50} \sum_{m=1}^{50} e_1 (\bm{X}_m) \right), \\
    \bm{z}_n &= \bm{\mu} + \bm{\sigma} \odot \bm{\epsilon}_n.
\end{align*}
The latent distribution $\mathcal{N}(\bm{z}; \bm{\mu}, \mathrm{diag}(\bm{\sigma}^2))$ is thus the low-dimensional probabilistic representation of the ensemble forecast fields that we aimed for.
The decoder $d$ with parameters $\theta$ is then applied to the sample $\bm{z}_n$ from the latent distribution to generate reconstructed forecast field $\tilde{\bm{X}}_n$, parameterizing the likelihood $p_\theta(\bm{X} | \bm{z})$ in the data space.       
In contrast to standard VAEs, we decode an arbitrary number $N$ of samples $\bm{z} = \{\bm{z}_n\}_{n=1}^N$ for each data point, producing an ensemble of reconstructed forecast fields as output.

The architecture of our iVAE is built on the AE model discussed above and employs fully-connected dense layers.
The shared encoder adds an average pooling layer to the encoder of the AE model, consisting of $e_1$ with one layer of size ``6561 - 4096" and $e_2$ with two layers of sizes ``4096 - 4096 - $d_{\text{latent}}$", while the decoder follows the same structure as the decoder of the AE with layers of sizes ``$d_{\text{latent}}$ - 4096 - 6561".
The LeakyReLU activation function is again applied in the hidden layers, and the same AdamW optimizer and early stopping criterion are applied.
Due to the significantly increased memory requirements for training the iVAE model, we employ mini-batch training with a batch size of 64.

The training objective of a standard VAE is to maximize the evidence lower bound on the marginal likelihood of the data,
\begin{equation*}
    \mathcal{L} (\theta,\phi) = \log p_\theta (\bm{X}|z) - D_{\text{KL}} ( q_\phi (\bm{z} | \bm{X}) \| p_\theta (\bm{z}) ),
\end{equation*}
which consists of a negative reconstruction error and a regularization term.
The reconstruction error $-\log p_\theta (\bm{X}|\bm{z})$ measures how well the model reconstructs the input data, which is proportional to the mean square error (MSE) with the Gaussian assumption on the data distribution for deterministic input and output.
The regularization term is the Kullback-Leibler divergence between the approximate posterior $q_\phi (\bm{z} | \bm{X})$ from the encoder and the prior $p_\theta (\bm{z})$ of the latent code $\bm{z}$, where standard multivariate Gaussian distributions are often used as the prior.

In our case, however, the probabilistic nature of both input and output of the iVAE necessitates a different notion of the reconstruction error.
Our iVAE model takes an ensemble of forecast fields $\bm{X} = \{\bm{X}_m\}_{m=1}^{50}$ as input and generates an ensemble of reconstructed fields $\tilde{\bm{X}} = \{\tilde{\bm{X}}_n\}_{n=1}^N$ with size $N$ as output.
The number $N$ is not necessarily equal to 50, and each input member $\bm{X}_m$ does not match the corresponding output member $\tilde{\bm{X}}_n$ for $m=n$ due to the random sampling of latent distribution.
Therefore, the MSE between $\bm{X}_m$ and $\tilde{\bm{X}}_n$ is not a suitable choice for estimating the reconstruction error.
Given that both the input and output ensembles of the iVAE can be considered to be multivariate empirical probability distributions, notions of the distance between the two distributions yield a more appropriate choice.
We thus incorporate two such metrics into the iVAE training objective. Specifically, we use the energy distance and the Sinkhorn distance, which will be introduced in Section \ref{evaluation}, for measuring different aspects of distances between multivariate probability distributions.
The reconstruction error of our iVAE is defined as the weighted sum of the energy distance and the Sinkhorn distance, complemented by the Kullback-Leibler divergence as a regularization term.
The three loss components exhibit significantly different scales, necessitating rescaling to ensure that their value ranges are comparable.
By comparing the mean values of different loss components over the first 20 epochs of training, we applied the following adjustments: the KL divergence was divided by 10, the energy distance was multiplied by 2 for both weather variables, and the Sinkhorn distance was divided by 50 for temperature and by 500 for wind speed.
The loss function of our iVAE model for temperature data thus is
\begin{equation}\label{equ_loss}
    \ell (\bm{X}, \tilde{\bm{X}}) = \omega_1 \cdot 2 D(\bm{X}, \tilde{\bm{X}}) + \omega_2 \cdot \frac{1}{50} \text{SD} (\bm{X}, \tilde{\bm{X}}) + \omega_3 \cdot \frac{1}{10} D_{\text{KL}} ( q_\phi (\bm{z} | \bm{X}) \| p_\theta (\bm{z}) ),
\end{equation}
where $D(\cdot)$ represents the energy distance and $\text{SD}(\cdot)$ denotes the Sinkhorn distance.
In our preliminary experiments, we observed that assigning a high weight to the KL divergence component in the loss function restricts the information flow through the bottleneck of the network, which results in outputs that fail to preserve the general spatial patterns of the input forecast fields.
To mitigate this issue and alleviate posterior collapse, we, like many other studies, heuristically selected a small weight $\omega_3 = 0.01$ for the KL divergence component.
For a more comprehensive understanding of the posterior collapse problem, we refer to \citet{lucas2019understanding}.
Regarding the two components used to measure reconstruction error, we assign equal weights of $\omega_1 = \omega_2 = 0.5$. Further discussions on the impact of different weighting schemes on the evaluation results are provided in the ablation studies in Section \ref{sec_ablation}.

\section{Results}\label{results}

In the following, we first briefly introduce the evaluation methods tailored for our specific problem, and then present and discuss the corresponding results for the dimensionality reduction methods introduced above.
Finally, we analyze the learned low-dimensional representations from different methods in an exemplary use case.

\subsection{Evaluation methods}\label{evaluation}

Choosing appropriate evaluation methods for our specific setting presents a challenge.
Two primary perspectives guide our evaluation: assessing the accuracy of the reconstructed ensemble forecast fields in comparison to the original ensemble fields, and analyzing the information content of the learned low-dimensional representations.
Evaluating the latter is particularly challenging as there naturally is no ground truth information for the representations, and the suitability will strongly depend on the application use case.
Therefore, our main focus is on discrepancy measures between the reconstructed output and the original input ensemble fields.
The discrepancy could be evaluated in terms of independent pixel-wise errors at each grid point, and joint, whole-image evaluation, where the entire forecast field is considered at once.
Several evaluation metrics are available for both settings and will be introduced in the following.

All three dimensionality reduction approaches learn a low-dimensional Gaussian distribution for representing an ensemble of forecast fields.
We draw 50 samples from the learned distribution and decode them into reconstructed forecast fields to enable a fair comparison when assessing the discrepancy with the raw 50-member ensemble.
As discussed above, the reconstructed ensemble members do not necessarily match the individual raw ensemble members, which prohibits measuring the pair-wise differences directly.
As an alternative, we compare the mean and standard deviation of all ensemble members between the raw and reconstructed fields at each grid point, providing insight into how well the model captures general characteristics of the input ensemble.
Given an ensemble of spatial forecast fields $\bm{X} = \{\bm{X}_m\}_{m=1}^{50}$ and an ensemble of reconstructed fields $\tilde{\bm{X}} = \{\tilde{\bm{X}}_n\}_{n=1}^{N}$ with $N=50$, we compute the absolute difference of ensemble means and the standard deviation difference at each grid point $(i,j)$,
\begin{equation*}
    e_{(i,j)} = \left|\frac{1}{50}\sum_{m=1}^{50} \bm{X}_{m, (i,j)} - \frac{1}{N}\sum_{n=1}^{N} \tilde{\bm{X}}_{n, (i,j)} \right|, \quad
    \Delta (\sigma)_{(i,j)} = \sigma(\bm{X}_{(i,j)}) - \sigma(\tilde{\bm{X}}_{(i,j)}),
\end{equation*}
where $i,j \in \{1,\ldots,81\}$ and $\sigma(\cdot)$ denotes the standard deviation of the respective ensemble.

We further consider probabilistic measures to quantify the discrepancy between two distributions of the ensembles used as inputs and obtained as outputs.
The evaluation of ensemble forecast fields could be executed pixel-wise, considering the (one-dimensional) univariate distribution at each grid point, or for the whole image, treating all grid points together as a high-dimensional multivariate distribution.
Measuring the divergence between two distributions for evaluating climate models in the univariate setting has been studied by \citet{thorarinsdottir2013using}, and here we consider the distance measures for both univariate and multivariate settings, utilizing the energy distance and optimal transportation distances.
The multivariate measures are also integrated into the reconstruction error component of the loss function for training our iVAE models, as discussed earlier in Section \ref{sec_ivae}.

The energy distance introduced by \citet{szekely2013energy} is a metric that measures the distance between two probability distributions.
Following our notations in Section \ref{sec_problem}, consider an ensemble of forecast fields $\bm{X} = \{\bm{X}_m\}_{m=1}^{50}$ and reconstructed fields $\tilde{\bm{X}} = \{\tilde{\bm{X}}_n\}_{n=1}^{N}$ in $\mathbb{R}^{d_{\text{data}}}$ with $N=50$, with the assumption that the ensemble members follow an idential distribution, i.e., $\bm{X}_m \sim \mathcal{P}$ for $m \in \{1, \ldots, 50\}$ and $\tilde{\bm{X}}_n \sim \tilde{\mathcal{P}}$ for $n \in \{1, \ldots, N\}$. The squared energy distance between $\mathcal{P}$ and $\tilde{\mathcal{P}}$ can be estimated in terms of expected pairwise distances between the two ensembles of samples,
\begin{equation*}
    D^2(\bm{X}, \tilde{\bm{X}}) = \frac{2}{50N} \sum_{m=1}^{50} \sum_{n=1}^N \big\|\bm{X}_{m} - \tilde{\bm{X}}_{n}\big\| - \frac{1}{50^2} \sum_{m_1=1}^{50} \sum_{m_2=1}^{50} \big\|\bm{X}_{m_1} - \bm{X}_{m_2}\big\| - \frac{1}{N^2} \sum_{n_1=1}^N \sum_{n_2=1}^N \big\|\tilde{\bm{X}}_{n_1} - \tilde{\bm{X}}_{n_2}\big\|,
\end{equation*}
where $\|\cdot\|$ is the Euclidean norm in $\mathbb{R}^{d_{\text{data}}}$.
The energy distance $D(\bm{X}, \tilde{\bm{X}})$ is negatively oriented and is zero if and only if the two empirical distributions with samples $\bm{X}$ and $\tilde{\bm{X}}$ coincide.
In the univariate setting of pixel-wise evaluation, the squared energy distance at each grid point $(i,j)$ is thus
\begin{align*}
    D^2_{(i,j)} = \frac{2}{50N} \sum_{m=1}^{50} \sum_{n=1}^N \big|\bm{X}_{m,(i,j)} - \tilde{\bm{X}}_{n,(i,j)}\big| &- \frac{1}{50^2} \sum_{m_1=1}^{50} \sum_{m_2=1}^{50} \big|\bm{X}_{m_1,(i,j)} - \bm{X}_{m_2,(i,j)}\big| \\
    &- \frac{1}{N^2} \sum_{n_1=1}^N \sum_{n_2=1}^N \big|\tilde{\bm{X}}_{n_1,(i,j)} - \tilde{\bm{X}}_{n_2,(i,j)}\big|.
\end{align*}
The univarite squared energy distance is closely related to the Cramér distance \citep{rizzo2016energy}, which is also known as the integrated quadratic distance \citep{thorarinsdottir2013using} for evaluating probabilistic forecasts from climate models.
The computation of univariate energy distance $D_{(i,j)}$ follows existing Python implementations from the \texttt{scikit-learn} library \citep{scikitlearn}, while the multivariate energy distance $D(\bm{X}, \tilde{\bm{X}})$ is implemented by custom code.

The optimal transportation distances, also known as the $p$-Wasserstein distances \citep{kantorovich1960mathematical}, are another type of metric that measure distances between two probability distributions.
The general optimal mass transport problem aims to find the optimal strategy to transport probability mass from one probability measure into another while minimizing the transportation cost, where the $p$-Wasserstein distance is the minimum total cost with the cost function $c(x,y) = |x-y|^p$.
Optimal transportation distances have found broad applications in machine learning in recent years \citep{frogner2015learning, courty2016optimal, arjovsky2017wasserstein}.
For an overview of the theory and methodology of optimal transportation distances and their applications, we refer to \citet{kolouri2017optimal}.
In our univariate setting of pixel-wise evaluation, the empirical format of 1-Wasserstein distance is considered.
The 1-Wasserstein distance between an input ensemble of forecast fields $\bm{X}$ and an output ensemble of reconstructed fields $\tilde{\bm{X}}$ at each grid point $(i,j)$ is estimated based on order statistics,
\begin{equation*}
    W_{1,(i,j)} = \frac{1}{50} \sum_{k=1}^{50} \big|\bm{X}_{(k),(i,j)} - \tilde{\bm{X}}_{(k),(i,j)}\big|,
\end{equation*}
where the subscript $\cdot_{(k)}$ denotes the $k$-th order of values.
We follow existing Python implementations from the \texttt{scikit-learn} library \citep{scikitlearn} for the computation of univariate 1-Wasserstein distance $W_{1, (i,j)}$.
In the multivariate setting of whole-image evaluation, estimating the Wasserstein distance between two high-dimensional distributions requires substantial computational cost and is thus not suitable for both evaluation tasks and particularly the integration into the iVAE training loss.
The Sinkhorn distance proposed by \citet{cuturi2013sinkhorn} provides a computationally efficient approximation of the Wasserstein distance by leveraging entropic regularizations.
We thus utilize the Sinkhorn distance as an alternative, and follow the Sinkhorn algorithm proposed in \citet{eisenberger2022unified} for a memory-efficient estimation.
The MSE is used as the cost function, i.e., $c(x,y) = \|x-y\|^p$ with $p=2$.
Thereby, the estimated Sinkhorn distance $\text{SD}(\bm{X}, \tilde{\bm{X}})$ approximates the 2-Wasserstein distance between $\bm{X}$ and $\tilde{\bm{X}}$,
\begin{equation*}
    W_2 = \inf_{\pi} \left( \sum_{k=1}^{50} \big\|\bm{X}_{k} - \tilde{\bm{X}}_{\pi(k)}\big\|^2 \right)^{1/2},
\end{equation*}
where $\pi$ denotes all possible permutations.

To compare different methods based on the same distance measure with respect to a benchmark, we further compute the associated skill score for analyzing the relative performances.
Among our three dimensionality reduction approaches, the PCA-based approach is naturally taken as the reference baseline method.
For each day in the test set, we first compute either the mean distance over all grid points for the pixel-wise metric, or directly take the distance measure for the entire grid as the score $S_a$ of a certain approach.
The skill score $SS_a$ is then calculated via
\begin{equation*}
    SS_a = \frac{S_{\text{ref}} - S_a}{S_{\text{ref}} - S_{\text{opt}}},
\end{equation*}
where $S_{\text{ref}}$ is the corresponding score of the reference PCA-based approach, and $S_{\text{opt}} = 0$ represents the score of an optimal method, i.e., an ideal model that replicates its input ensemble members perfectly.
Skill scores are positively oriented and have an upper bound of 1, where a positive value indicates better performance than the benchmark, and a value of 0 indicates no improvement over the benchmark.

For brevity, we present only the results for the energy distances in the main text, whereas the results concerning the optimal transport distances are deferred to the Supplemental Material.

\subsection{Reconstruction accuracy}

We compare the accuracy of the reconstructed ensemble forecast fields obtained by the three different approaches (i.e., PCA-based and AE-based two-step methods and the iVAE approach).
We here focus on 2-m temperature and the U component of 10-m wind speed, and present results for the corresponding V component and for the geopotential height at 500 hPa in the Supplemental Material.

\begin{figure}
\centering
    \includegraphics[width=0.95\textwidth]{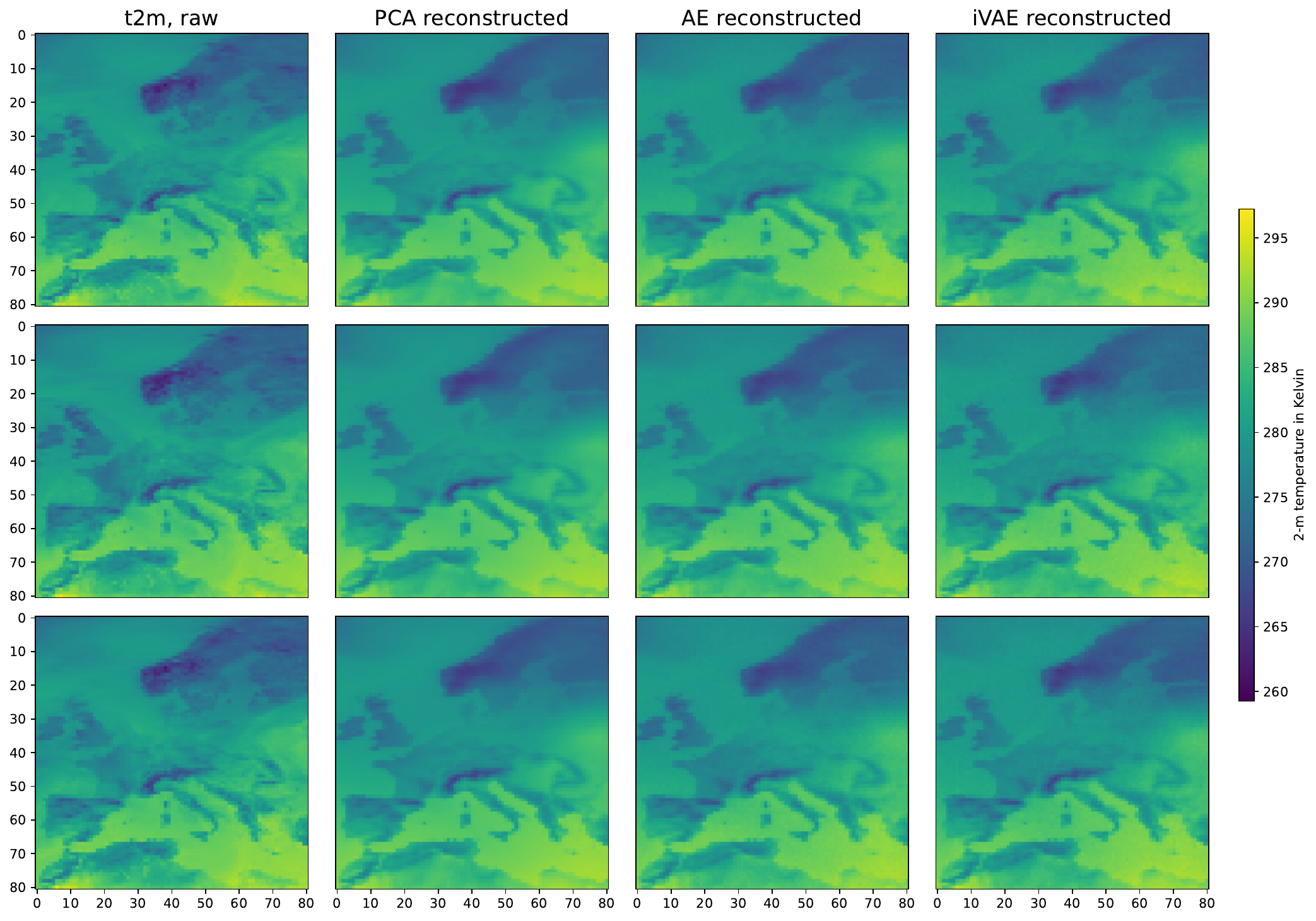}
    \includegraphics[width=0.95\textwidth]{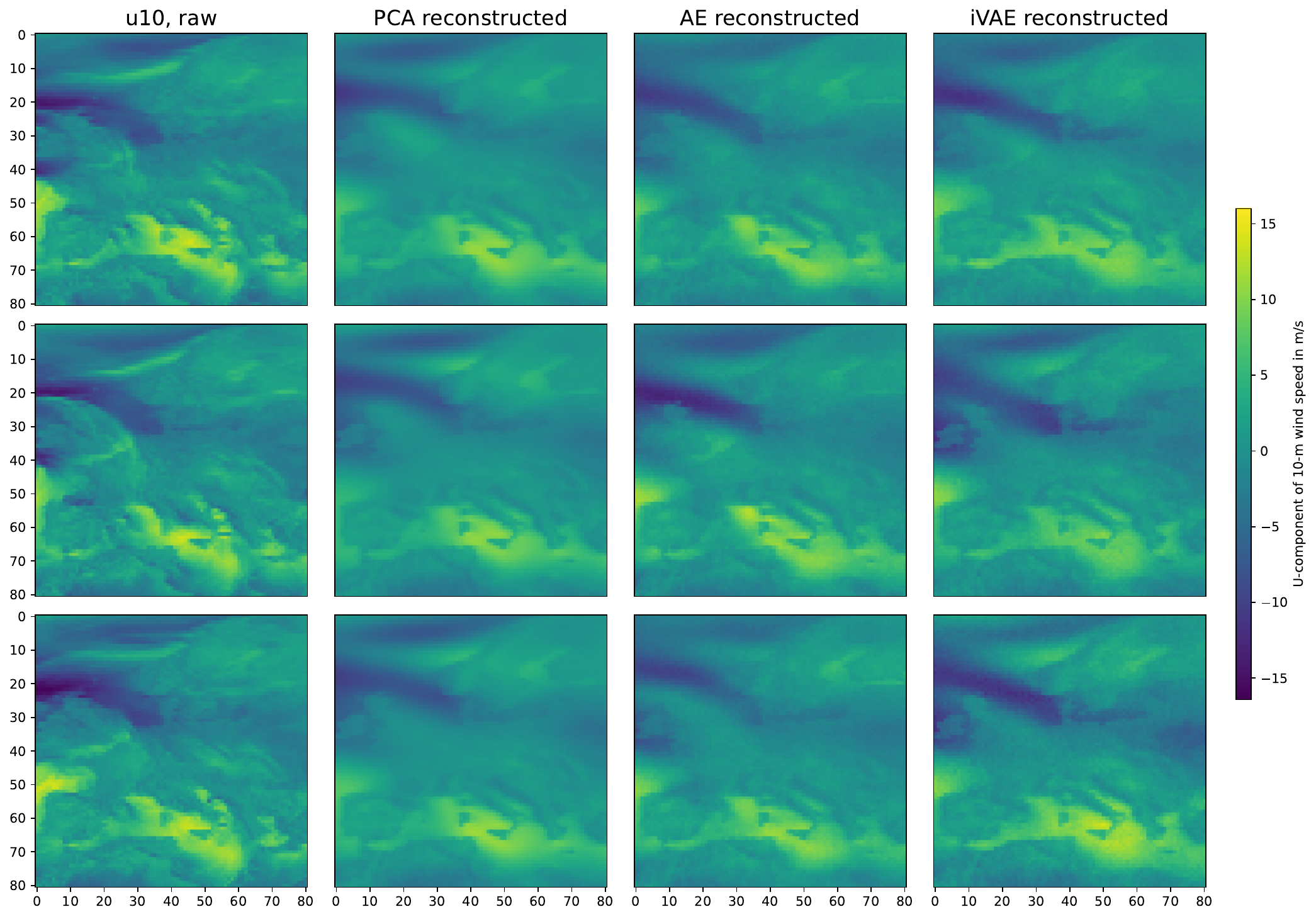}
    \caption{Exemplary raw forecast fields and reconstructed forecast fields of 2-m temperature (top) and the U component of 10-m wind speed (bottom) by different methods, with a latent dimension of 32. The rows correspond to different ensemble members for the same forecast day.}
    \label{fig_example}
\end{figure}

Figure \ref{fig_example} shows examples of input forecast fields along with reconstructed ensemble members produced as output of the different approaches.
As discussed above and in light of our aim of learning representations of the underlying probability distributions, the reconstructed ensemble members should not be expected to perfectly match the input even though the selected fields are from the same forecast day.
For both target variables, all approaches are able to capture the general spatial structure in the raw fields despite reducing the dimensionality from 6561 to 32.
However, neither of them is able to realistically replicate the localized fine patterns in the raw fields, with the AE and iVAE methods showing slightly better, specifically for the U component of wind speed based on visual inspection across ensemble members. 
The reconstruction quality and level of fine-scale details can be improved for higher-dimensional latent representation.

\begin{figure}
\centering
    \includegraphics[width=0.96\textwidth]{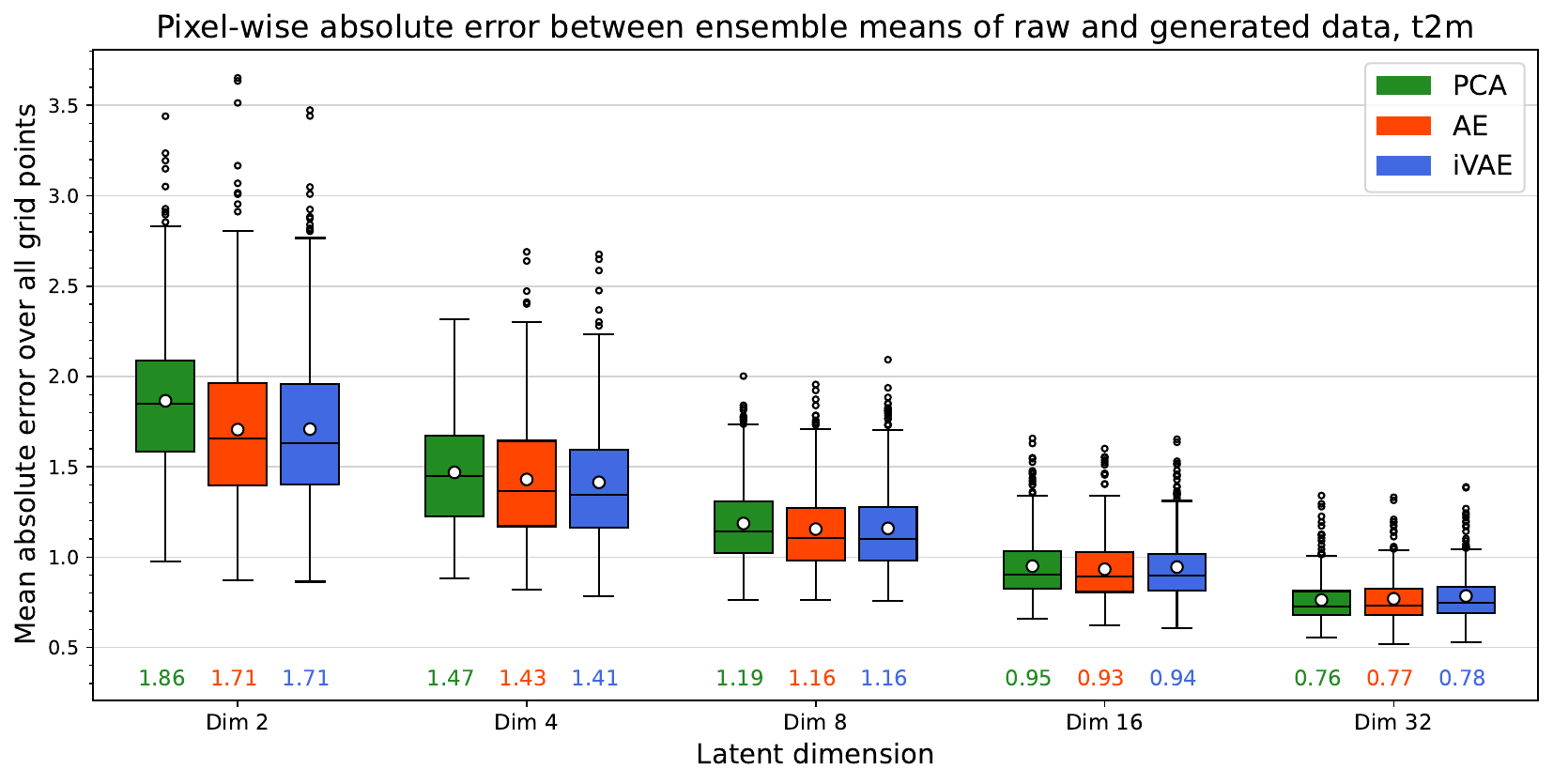}
    \includegraphics[width=0.96\textwidth]{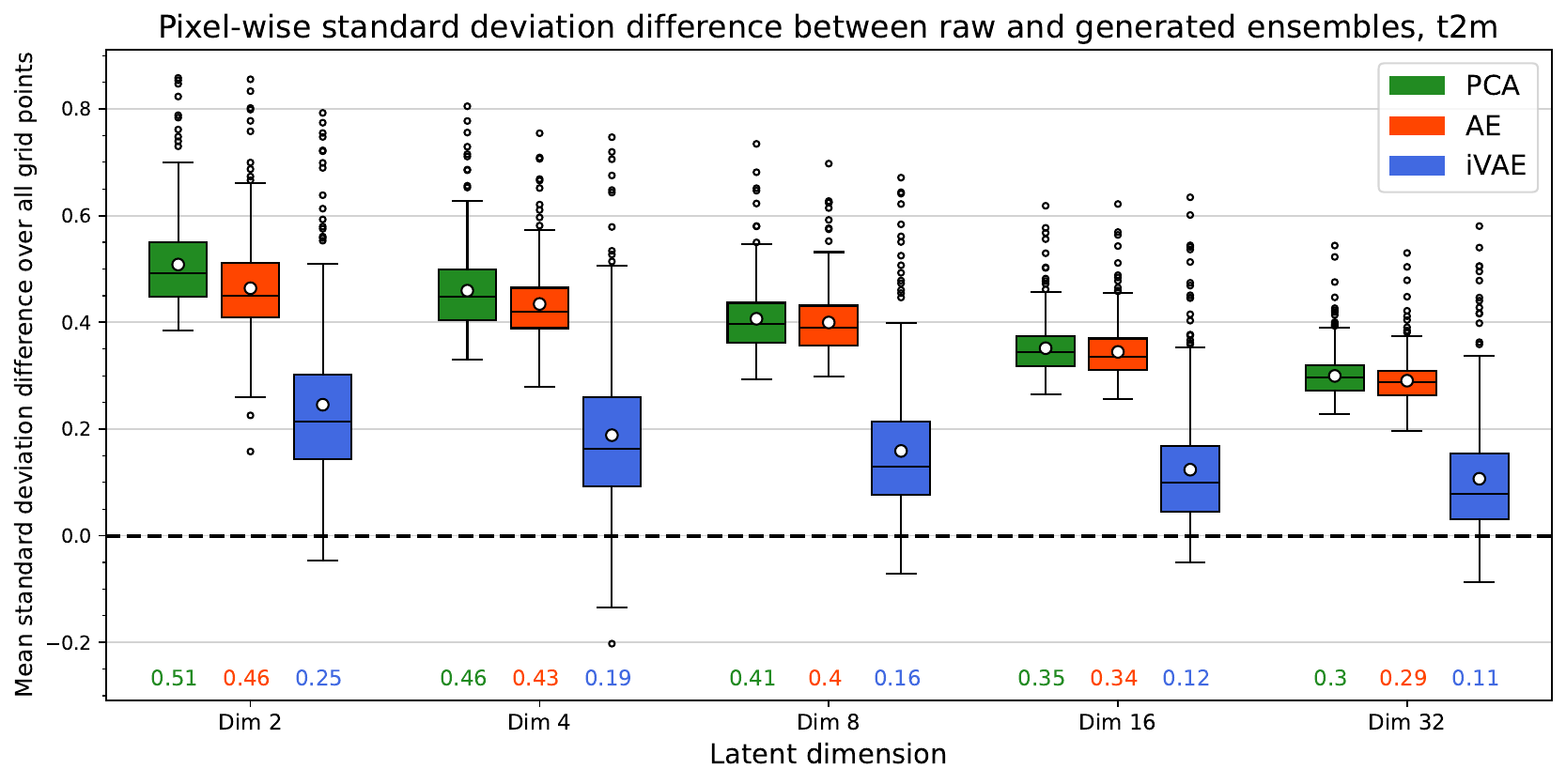}
    \caption{Boxplots of mean absolute differences between the mean values of input and reconstructed ensemble fields (top) and differences between the standard deviations of input and reconstructed ensemble fields (bottom) at each grid point. 
    Boxes show performance variability over 366 days in the test set of different methods for 2-m temperature data, considering 5 different dimensionalities of the latent representation.
    The mean values of the (absolute) differences are indicated below each box. The differences between the standard deviations are computed such that negative values indicate a larger variability of the reconstructed ensemble compared to the input ensemble.}
    \label{fig_mae_std_box_t}
\end{figure}

\begin{figure}
\centering
    \includegraphics[width=0.96\textwidth]{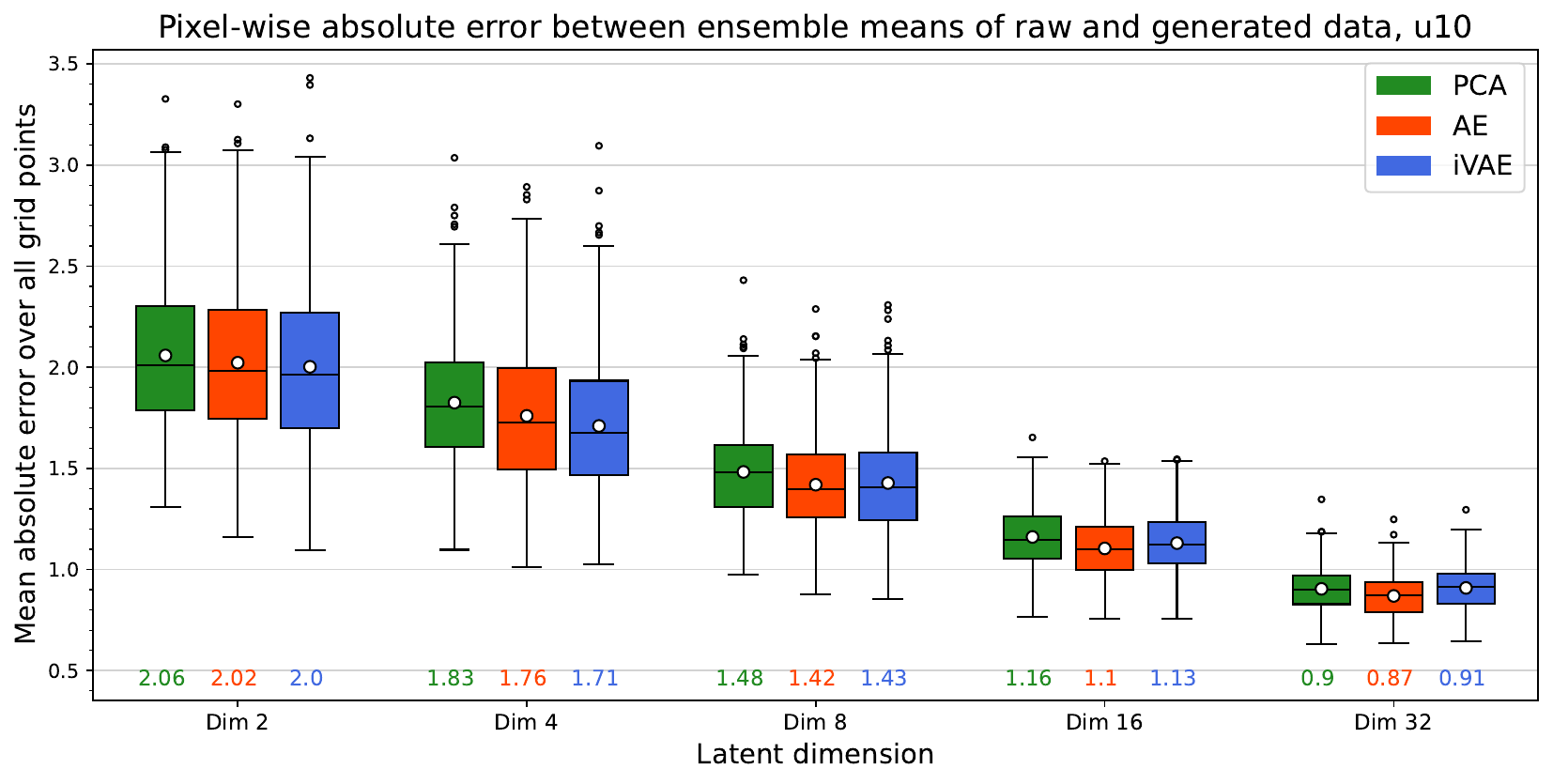}
    \includegraphics[width=0.96\textwidth]{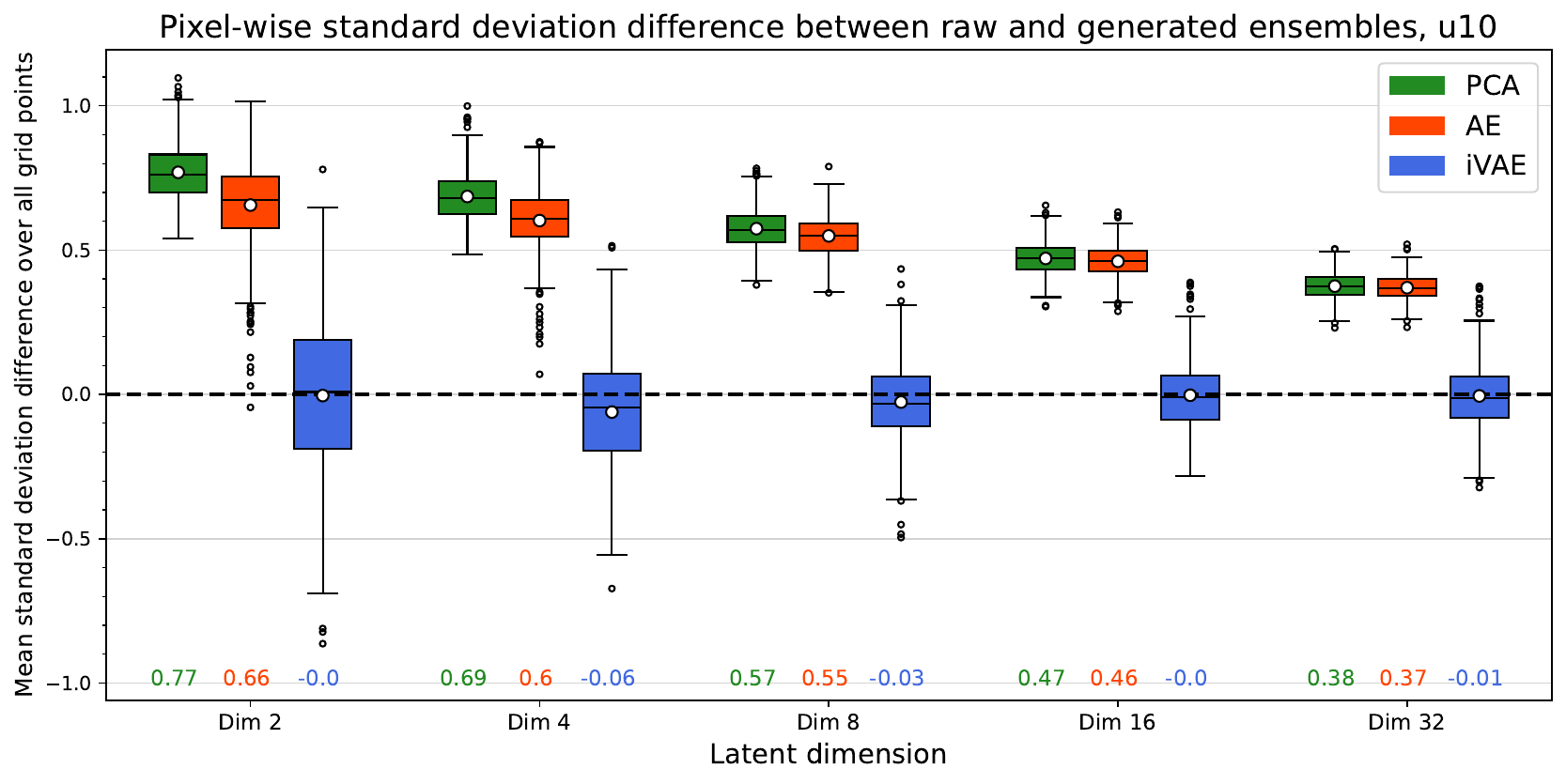}
    \caption{As Figure \ref{fig_mae_std_box_t}, but for the U component of 10-m wind speed.}
    \label{fig_mae_std_box_u}
\end{figure}

To assess the reconstruction quality in terms of general summary statistics of the input and output ensemble fields, Figure \ref{fig_mae_std_box_t} shows pixel-wise absolute differences of the corresponding ensemble mean and standard deviation values for temperature data.
For all methods, the differences between the summary statistics of the input and reconstructed ensemble fields decrease with increasing dimensionality of the latent representations.
In terms of the deviations of the ensemble mean, the two neural network-based methods show slightly better performance in lower-dimensional settings, while PCA shows minimally better performance for the largest dimensionality considered here.
However, these differences are very minor, in particular when compared to the variability within the boxplots.
More substantial differences can be observed for the standard deviation among the reconstructed ensemble members, where only the iVAE approach is able to produce variability among the members of a magnitude that matches the raw ensemble, whereas the reproduced ensembles from both two-step methods notably underestimate the variability of the input.
Qualitatively similar results are obtained for the wind speed data, shown in Figure \ref{fig_mae_std_box_u}, where the better performance of the iVAE approach at correctly reproducing the variability across ensemble members is even more apparent. 

\begin{figure}
\centering
    \includegraphics[width=0.94\textwidth]{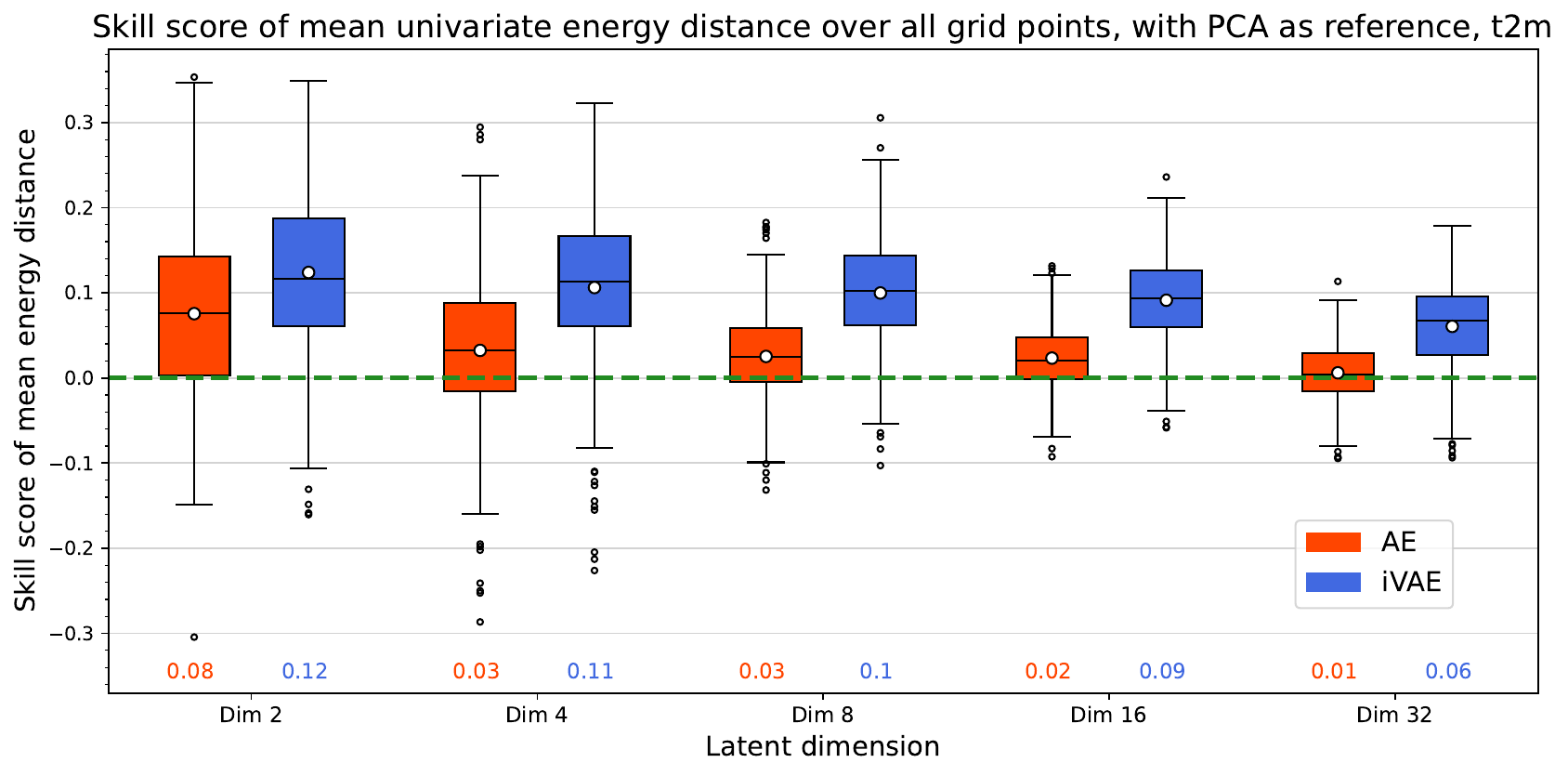}
    \includegraphics[width=0.94\textwidth]{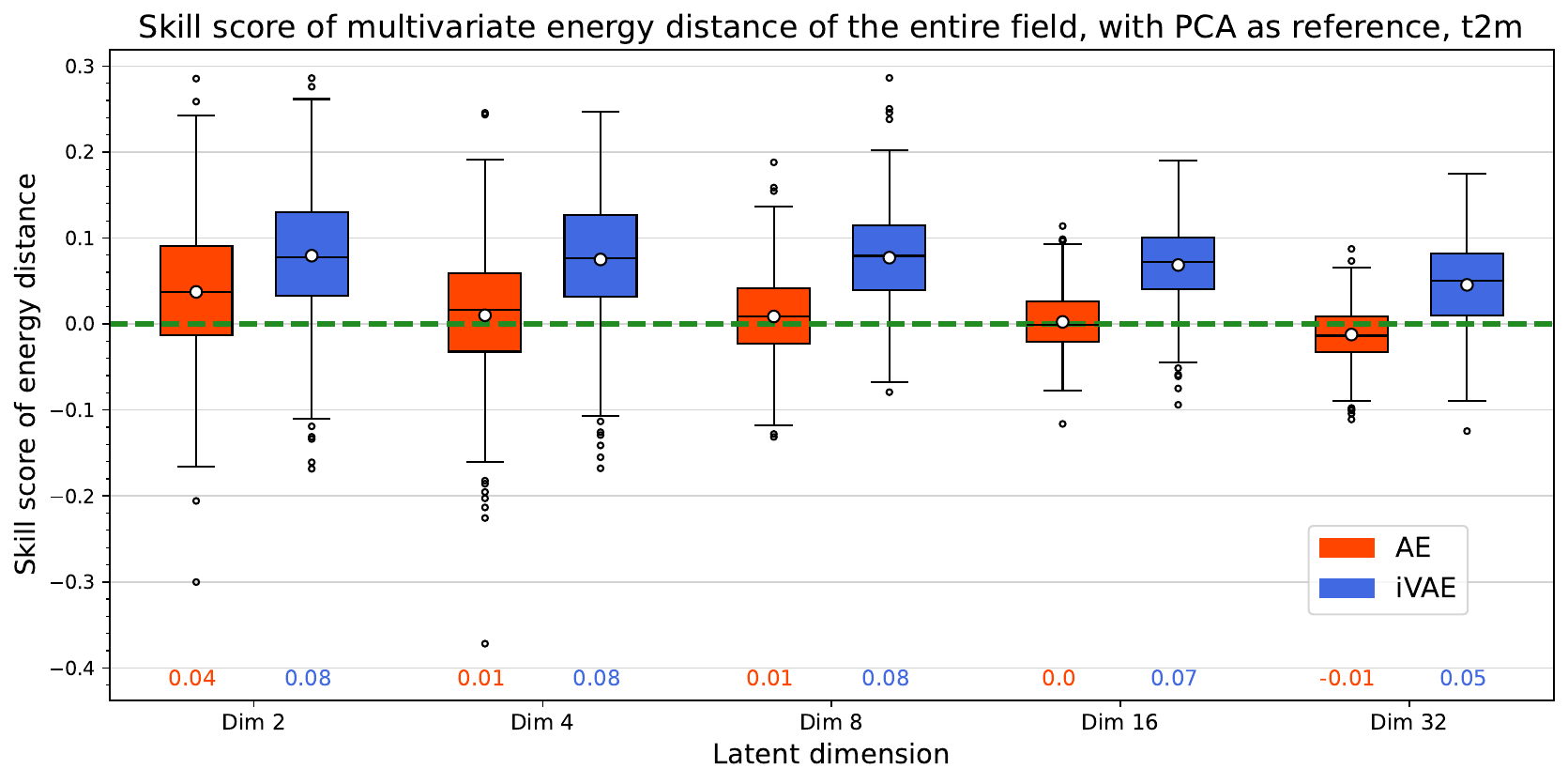}
    \caption{Boxplots of skill scores based on energy distances between the input and reconstructed ensemble fields over the 366 days in the test set for temperature data. The panels show mean univariate energy distances over all grid points (top) and multivariate energy distances computed for the entire fields (bottom). PCA-based approach shown in green dashed line is the reference method. The respective mean skill values are indicated below each box.}
    \label{fig_ed_box_t}
\end{figure}

\begin{figure}
\centering
    \includegraphics[width=0.94\textwidth]{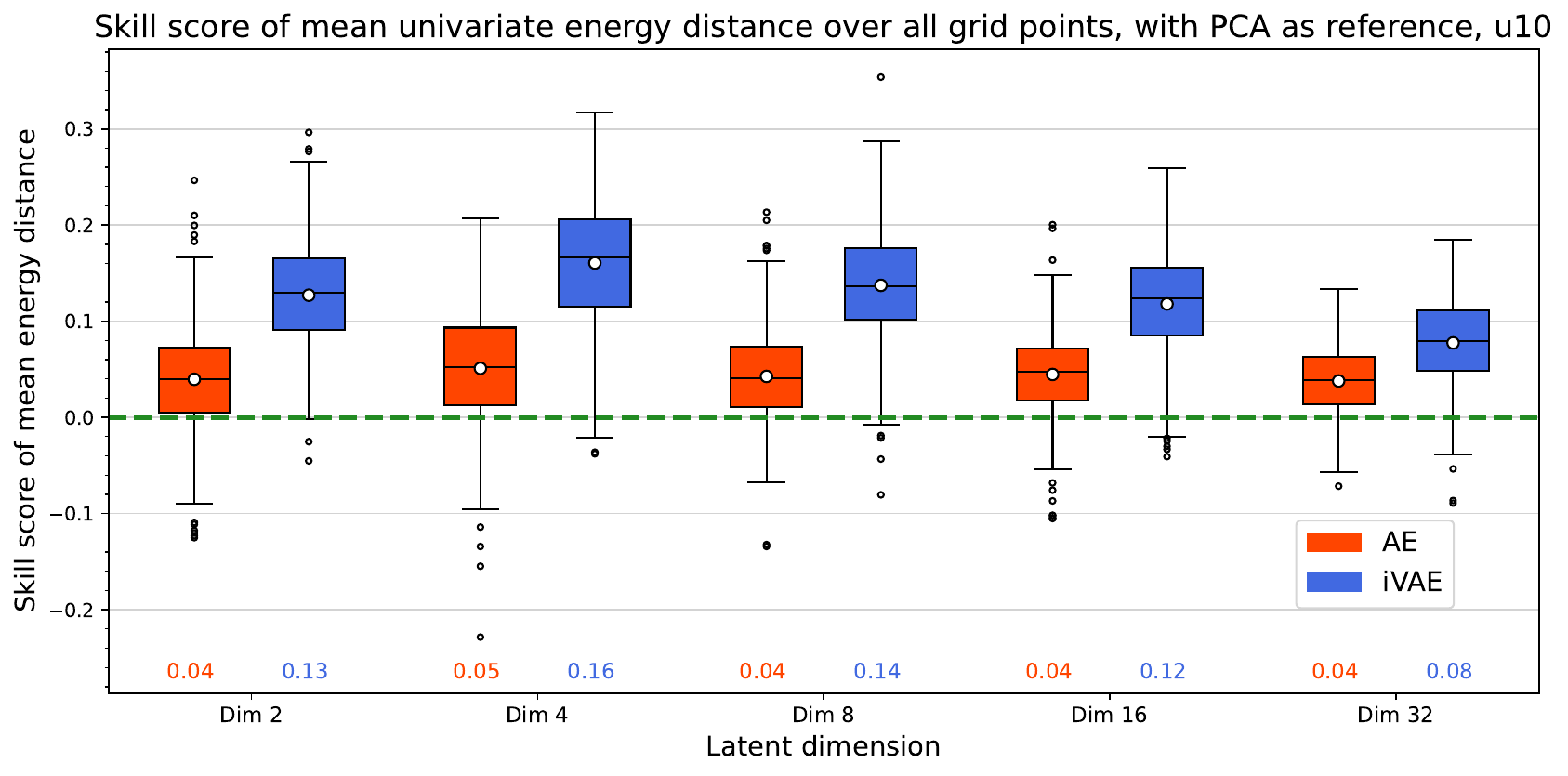}
    \includegraphics[width=0.94\textwidth]{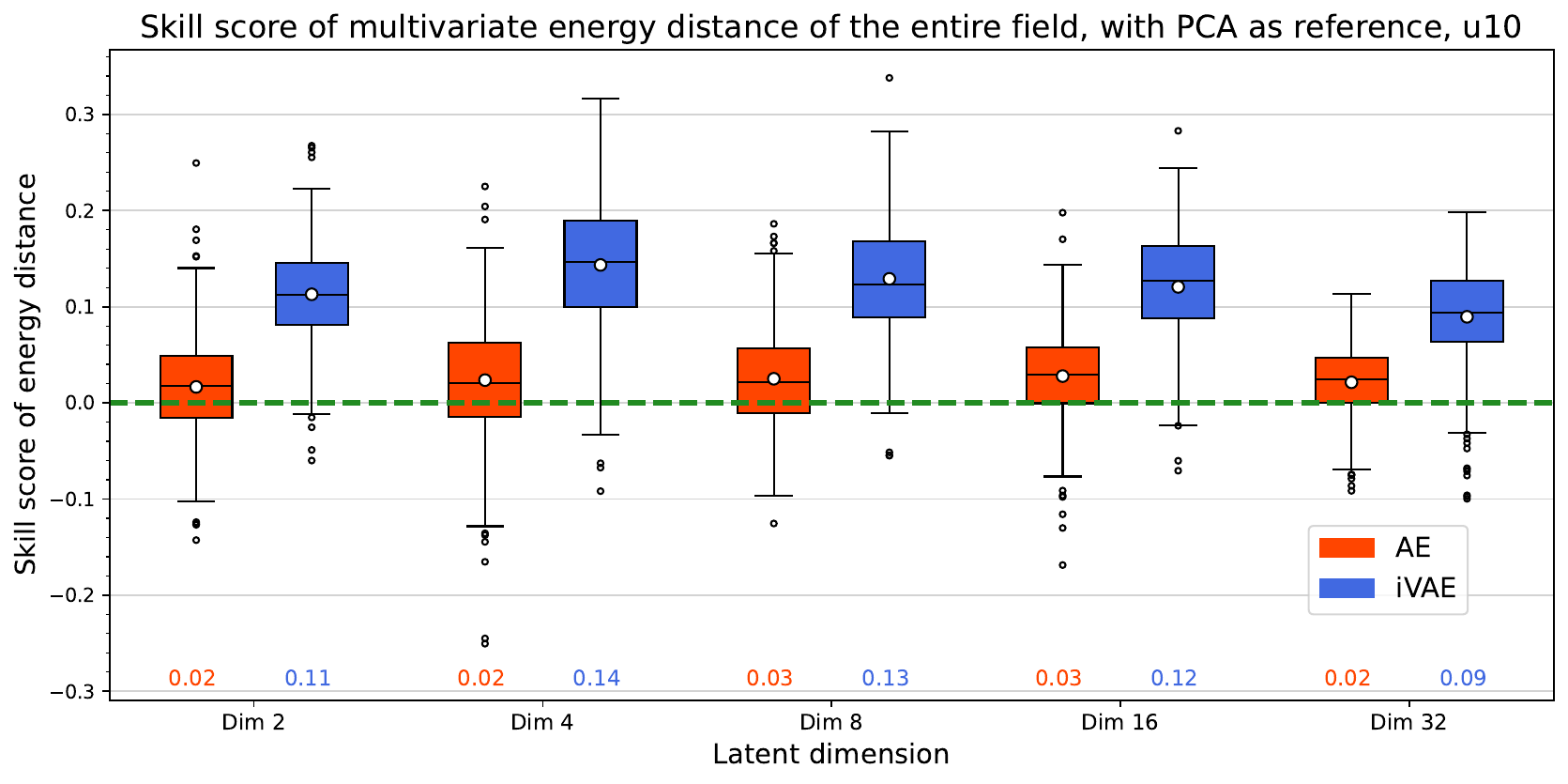}
    \caption{As Figure \ref{fig_ed_box_t}, but for the U component of 10-m wind speed.}
    \label{fig_ed_box_u}
\end{figure}

For a more fine-grained assessment of the probabilistic reconstruction quality, Figure \ref{fig_ed_box_t} shows skill scores based on energy distances between input and reconstructed temperature ensemble forecast fields for the AE-based and iVAE methods, with the PCA-based method as a baseline.
Positive values indicate an improvement in terms of the energy distance over the reference method. For example, a skill score of 0.1 corresponds to a 10\% lower energy distance compared to PCA.
The iVAE method consistently outperforms the other two approaches across all latent dimensions and in both univariate and multivariate evaluation.
These improvements are likely due to the variability of the reconstructed ensemble fields being close to the input ones as discussed in Figure \ref{fig_mae_std_box_t}, and potentially benefit from the inclusion of the multivariate energy distance as part of the loss function. 
The AE-based two-step approach generally outperforms the PCA-based approach, but the improvements are less pronounced than for the iVAE method, and poorer performance is observed for multivariate energy distances with a latent dimension greater than 8.
Both neural network-based approaches show less distinct improvements over the PCA-based method with increasing latent dimensions, possibly because PCA is sufficient to capture most of the variability information in the raw ensemble forecast fields when the dimension of representations is suitably large.
Similar conclusions can be drawn for the wind speed data, see Figure \ref{fig_ed_box_u}.
The most notable difference to the results for temperature data is that the AE-based method consistently outperforms the PCA-based method here, and the largest improvements from the iVAE method occur at a latent dimension of 4.

\subsection{Ablation studies on the weighted loss function of iVAE}\label{sec_ablation}

As discussed in Section \ref{sec_ivae}, the loss function of our iVAE model incorporates both the multivariate energy distance and the Sinkhorn distance to measure reconstruction error. Given that both distances are used to evaluate the performance of reconstructed ensemble forecast fields, we aim to investigate the impact of different weighting schemes on the evaluation results.

\begin{figure}
    \centering
    \includegraphics[width=0.63\linewidth]{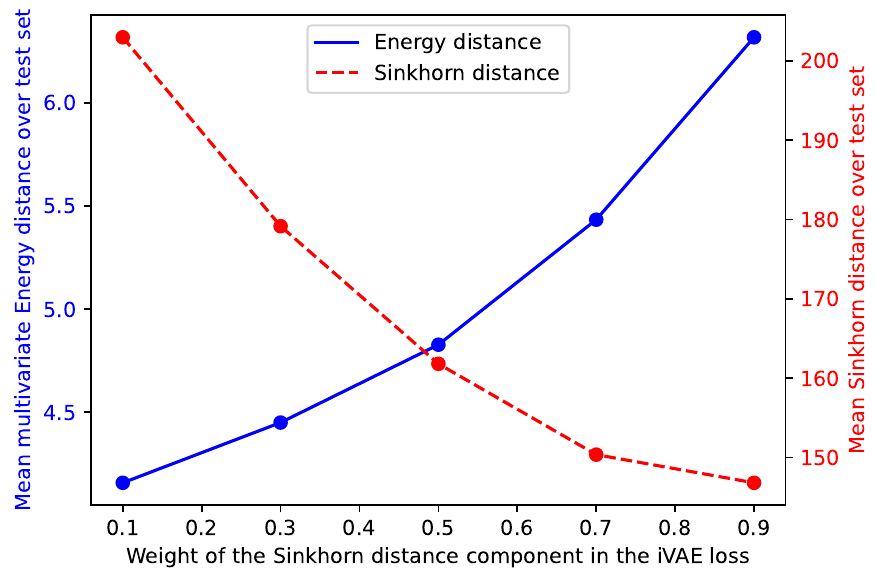}
    \caption{Mean multivariate energy distance (blue) and Sinkhorn distance (red) over the test set for reconstructed ensemble forecast fields of 2-m temperature generated by the iVAE method, shown as a function of the weight of the Sinkhorn distance component in the loss function.}
    \label{fig_loss_weight}
\end{figure}

To this end, we conduct ablation studies on several weighting scheme choices for temperature data. 
To ensure comparable scales of the weighted loss, we maintain $\omega_1 + \omega_2 = 1$ in the loss function (\ref{equ_loss}), and vary $\omega_2$ between 0 and 1. The averages of the corresponding evaluation distances in the test set are presented in Figure \ref{fig_loss_weight}.

As expected, our iVAE model demonstrates better performance in terms of Sinkhorn distance when a larger weight is assigned to the Sinkhorn distance component in the loss function during training, and vice versa. This results in a trade-off between improved energy distance performance and improved Sinkhorn distance performance. Our initial choice of an equal weighting scheme appears to be a suitable compromise.

\subsection{Exploratory analysis of the learned representations}

\begin{figure}
\centering
    \includegraphics[width=\textwidth]{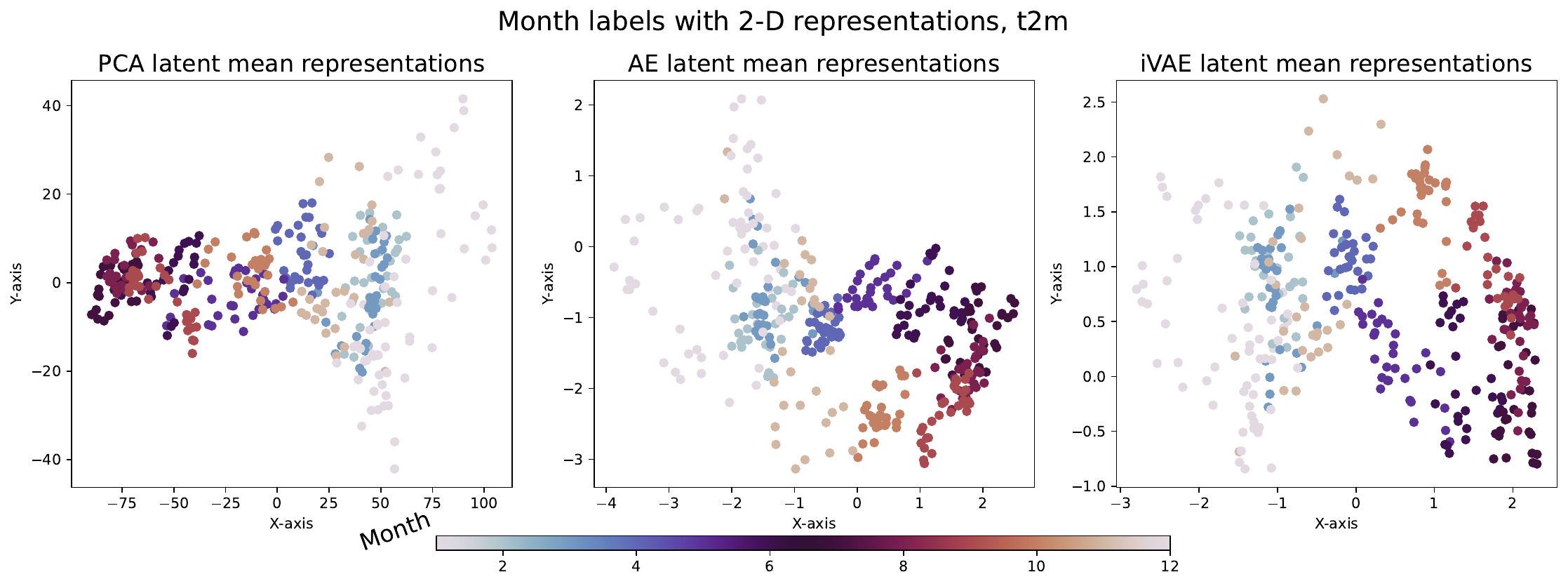}
    \caption{Scatterplots of the components of the mean vector of learned two-dimensional representations of temperature for the three different dimensionality reduction methods. The points are colored according to the month of the corresponding forecast date.}
    \label{fig_scatter_t2m}
\end{figure}

Here, we focus on another interesting question, which is whether the learned representations in the low-dimensional latent space carry any relevant meteorological information or can offer additional insights about the data at hand.
To that end, we focus on the temperature forecast data and try to detect seasonal patterns in the learned representations.
We restrict our attention to latent representations of dimension 2 to enable graphical visualization. 
Figure \ref{fig_scatter_t2m} shows scatterplots of the components of the mean vector of the learned latent distributions for the different approaches. 

The mean vectors of the latent representations from all three methods show clear patters corresponding to the seasonality of the 2-m temperature, with data points representing warmer and colder weather clustered together, respectively.
Notably, the two neural network-based approaches allow for a clearer distinction between different months, while the PCA-based representations tend to partially overlap for several months.

Similar scatterplots for the two components of 10-m wind speed are available in the Supplemental Material, however, there is a less clear seasonal pattern and a larger variability within individual months.
We further explored the connection of the learned representations of ensemble forecast fields of geopotential height at 500 hPa to quasi-stationary, recurrent, and persistent large-scale atmospheric circulation patterns, so-called weather regimes \citep{grams2017balancing}. 
While there are noticeable patterns and clusters which could be incorporated into post-processing models \citep{Mockert2024}, the analysis is more involved and a more detailed investigation is left for future work.
Some first results are available in the Supplemental Material.

\section{Discussion and conclusions}\label{conclusion}

We propose two types of approaches to learning probabilistic low-dimensional representations of ensemble forecast fields which aim to treat them as interchangeable samples from an underlying high-dimensional probability distribution. 
Two-step methods based on PCA or AE models first learn to assign a deterministic latent representation to each ensemble member, and proceed by fitting a multivariate probability distribution to the learned representations in the latent space. 
By contrast, the iVAE approach directly learns a probability distribution as latent representation and treats the ensemble members as invariant inputs.
Both approaches allow for efficiently reducing the dimensionality of ensemble forecast fields within a probabilistic setting, where the learned distributions in the latent space enable the generation of arbitrarily many reconstructed forecast fields.
Systematic comparisons of PCA- and AE-based approaches and the iVAE in case studies on temperature and wind speed forecasts over Europe indicate that the two NN-based approaches show promising results both in terms of the quality of reconstructed forecast fields and the informativeness of learned latent representations. 
While the results vary across all considered evaluation metrics, the iVAE model specifically performs best at preserving the variability information of the input ensemble forecasts.
When compared to PCA, the NN-based methods generally show better performance for lower latent dimensions, whereas PCA yields equally good or even better reconstructions when the latent dimension is large.
Overall, the computational costs of the iVAE are substantially higher than those of the AE-based and PCA-based approaches, although they remain manageable since the architectures are only comprised of dense layers.
On a multiple-node CPU cluster, it takes less than one hour to train the iVAE model, while it takes only about 10 minutes to train the AE- or PCA-based model.

Despite the promising results, there are limitations to both types of approaches.
All three methods assume a multivariate Gaussian distribution in the latent space, which might limit their applicability across different weather variables, specifically for variables such as precipitation, where the distribution should account for potential point masses at zero.
A particular challenge in the specific setting of our dimensionality reduction problem is the evaluation of different approaches.
In our experiments, we considered both deterministic and probabilistic metrics to assess the quality of reconstructed forecast fields.
However, there is a general lack of suitable probabilistic evaluation tools that take into account structural aspects of the (ensemble of) forecast fields, compared to perception-based metrics proposed in the computer vision literature such as the widely used structural similarity index \citep{wang2004image}.
Spatial evaluation approaches proposed in the meteorological literature might offer useful starting points, but are often heuristics-based, tailored to specific variables such as precipitation, and not straightforward to extend towards probabilistic settings, see, for example, \citet{gilleland2010verifying} and \citet{dorninger2018setup} for overviews. 

The proposed approaches provide several avenues for further generalization and analysis. 
Evidently, it would be of interest to investigate the scalability of the proposed methods towards larger grids and higher resolution forecast fields, as well as other variables.
In particular, spatial forecasts of precipitation have been a focal point of research interest, for example in spatial verification \citep{nigel2008scale}.
Further, while we applied the dimensionality reduction methods separately to ensemble forecast fields of different variables, it would be interesting to apply them jointly to forecasts of multiple variables, and potentially over multiple time steps, at the same time.
In addition, more advanced NN architectures such as transformers could be used as components of the AE-based or iVAE approaches.
While we did not observe any benefits from using vision transformer architectures in initial experiments in the context of our case study, the results may vary if more training data was available. 
Given the recent developments in modern AI-based weather forecasting, including the generation of AI-based ensemble forecasts \citep{NeuralGCM, GenCast, BuelteEtAl2024, MaheshEtAl2024, FuXiENS}, generating the corresponding ensemble forecast fields may become feasible and relevant for applications.

Finally, an important next step is to make progress towards integrating the proposed dimensionality reduction methods into downstream tasks.
As discussed in the introduction, a key motivation for learning low-dimensional latent representations was to use those representations either as input data in, for example, hydrological or energy forecasting models, or to augment the input for neural network-based post-processing models \citep{lerch2022convolutional}.
Further examples of potential applications include (sub)seasonal weather prediction, where PCA-based representations of ensemble forecast fields have been used as inputs to machine learning models \citep{Kiefer2023, kiefer2024random,ScheuererEtAl2024}, as well as analog forecasting, where compressed information from raw forecasts could be utilized for a more efficient generation of analogs \citep{analog2021grooms, machine2021yang}.
It is important to note that the quality requirements relevant to such applications differ from those in data reduction applications with a focus on data storage and management \citep[see, e.g.,][]{duben2019new,hoehlein2022evaluation}. While such approaches pursue accurate reconstruction as the central quality criterion, the achievable reconstruction quality may be decoupled from the information value of compressed representations in integrated modeling workflows.
For example, in the specific context of incorporating learned representations of spatial input fields into NN-based post-processing models, one has to carefully balance the added value of the spatial information and the increased number of input predictors when the availability of training data is limited. 
In the setting of \citet{lerch2022convolutional}, we did not find any improvements when using learned representations of the full ensemble of forecast fields instead of the mean forecast only. 
One explanation in line with other findings from related work \citep{hohlein_postprocessing_2024,FeikEtAl2024} might be that for post-processing, there seems to often be little value of including full information from an ensemble beyond simple summary statistics.

\section*{Acknowledgments}

The research leading to these results has been done within the Young Investigator Group ``Artificial Intelligence for Probabilistic Weather Forecasting'' funded by the Vector Stiftung. In addition, this project has received funding within project T4 ``Development of a deep learning prototype for operational probabilistic wind gust forecasting'' of the Transregional Collaborative Research Center SFB/TRR 165 “Waves to Weather” funded by the German Research Foundation (DFG). 
We thank Fabian Mockert and Christian Grams for providing weather regime data and helpful discussions, and thank Nina Horat, Benedikt Schulz, Uwe Ehret and Tilmann Gneiting for helpful comments and discussions.
This work is supported by the Helmholtz Association Initiative and Networking Fund on the HAICORE@KIT partition.
The authors acknowledge support by the state of Baden-Württemberg through bwHPC.

\bibliographystyle{myims2}
\bibliography{reference}

\begin{thebibliography}{76}
\expandafter\ifx\csname natexlab\endcsname\relax\def\natexlab#1{#1}\fi
\expandafter\ifx\csname url\endcsname\relax
  \def\url#1{\texttt{#1}}\fi
\expandafter\ifx\csname urlprefix\endcsname\relax\def\urlprefix{URL }\fi
\providecommand{\eprint}[2][]{\url{#2}}

\bibitem[{Allen et~al.(2021)Allen, Evans, Buchanan and
  Kwasniok}]{AllenEtAl2021}
{Allen, S.}, {Evans, G.~R.}, {Buchanan, P.} and {Kwasniok, F.} (2021).
\newblock Incorporating the north atlantic oscillation into the post-processing
  of mogreps-g wind speed forecasts.
\newblock \textit{Quarterly Journal of the Royal Meteorological Society},
  {147}, 1403--1418.

\bibitem[{An and Cho(2015)}]{an2015variational}
{An, J.} and {Cho, S.} (2015).
\newblock Variational autoencoder based anomaly detection using reconstruction
  probability.
\newblock \textit{Special lecture on IE}, {2}, 1--18.

\bibitem[{Arjovsky et~al.(2017)Arjovsky, Chintala and
  Bottou}]{arjovsky2017wasserstein}
{Arjovsky, M.}, {Chintala, S.} and {Bottou, L.} (2017).
\newblock {W}asserstein generative adversarial networks.
\newblock In \textit{Proceedings of the 34th International Conference on
  Machine Learning} (D.~Precup and Y.~W. Teh, eds.), vol.~70 of
  \textit{Proceedings of Machine Learning Research}. PMLR, 214--223.

\bibitem[{Baccouche et~al.(2012)Baccouche, Mamalet, Wolf, Garcia and
  Baskurt}]{baccouche2012spatio}
{Baccouche, M.}, {Mamalet, F.}, {Wolf, C.}, {Garcia, C.} and {Baskurt, A.}
  (2012).
\newblock Spatio-temporal convolutional sparse auto-encoder for sequence
  classification.
\newblock In \textit{BMVC}. 1--12.

\bibitem[{Bauer et~al.(2015)Bauer, Thorpe and Brunet}]{bauer2015the}
{Bauer, P.}, {Thorpe, A.} and {Brunet, G.} (2015).
\newblock The quiet revolution of numerical weather prediction.
\newblock \textit{Nature}, {525}, 47--55.

\bibitem[{Bourlard and Kamp(1988)}]{bourlard1988auto}
{Bourlard, H.} and {Kamp, Y.} (1988).
\newblock Auto-association by multilayer perceptrons and singular value
  decomposition.
\newblock \textit{Biological Cybernetics}, {59}, 291--294.

\bibitem[{Burda et~al.(2016)Burda, Grosse and
  Salakhutdinov}]{burda2015importance}
{Burda, Y.}, {Grosse, R.} and {Salakhutdinov, R.} (2016).
\newblock Importance weighted autoencoders.
\newblock \eprint{1509.00519},
  \urlprefix\url{https://arxiv.org/abs/1509.00519}.

\bibitem[{Bülte et~al.(2024)Bülte, Horat, Quinting and
  Lerch}]{BuelteEtAl2024}
{Bülte, C.}, {Horat, N.}, {Quinting, J.} and {Lerch, S.} (2024).
\newblock Uncertainty quantification for data-driven weather models.
\newblock \eprint{2403.13458},
  \urlprefix\url{https://arxiv.org/abs/2403.13458}.

\bibitem[{Chapman et~al.(2022)Chapman, Monache, Alessandrini, Subramanian,
  Ralph, Xie, Lerch and Hayatbini}]{Chapman2022}
{Chapman, W.~E.}, {Monache, L.~D.}, {Alessandrini, S.}, {Subramanian, A.~C.},
  {Ralph, F.~M.}, {Xie, S.-P.}, {Lerch, S.} and {Hayatbini, N.} (2022).
\newblock Probabilistic predictions from deterministic atmospheric river
  forecasts with deep learning.
\newblock \textit{Monthly Weather Review}, {150}, 215 -- 234.

\bibitem[{Chen et~al.(2024)Chen, Janke, Steinke and Lerch}]{chen2024generative}
{Chen, J.}, {Janke, T.}, {Steinke, F.} and {Lerch, S.} (2024).
\newblock Generative machine learning methods for multivariate ensemble
  postprocessing.
\newblock \textit{The Annals of Applied Statistics}, {18}, 159--183.

\bibitem[{Courty et~al.(2016)Courty, Flamary, Tuia and
  Rakotomamonjy}]{courty2016optimal}
{Courty, N.}, {Flamary, R.}, {Tuia, D.} and {Rakotomamonjy, A.} (2016).
\newblock Optimal transport for domain adaptation.
\newblock \textit{IEEE transactions on pattern analysis and machine
  intelligence}, {39}, 1853--1865.

\bibitem[{Cuturi(2013)}]{cuturi2013sinkhorn}
{Cuturi, M.} (2013).
\newblock Sinkhorn distances: Lightspeed computation of optimal transport.
\newblock In \textit{Advances in Neural Information Processing Systems}
  (C.~Burges, L.~Bottou, M.~Welling, Z.~Ghahramani and K.~Weinberger, eds.),
  vol.~26. Curran Associates, Inc.

\bibitem[{Dorninger et~al.(2018)Dorninger, Gilleland, Casati, Mittermaier,
  Ebert, Brown and Wilson}]{dorninger2018setup}
{Dorninger, M.}, {Gilleland, E.}, {Casati, B.}, {Mittermaier, M.~P.}, {Ebert,
  E.~E.}, {Brown, B.~G.} and {Wilson, L.~J.} (2018).
\newblock The setup of the mesovict project.
\newblock \textit{Bulletin of the American Meteorological Society}, {99}, 1887
  -- 1906.

\bibitem[{Dosovitskiy et~al.(2021)Dosovitskiy, Beyer, Kolesnikov, Weissenborn,
  Zhai, Unterthiner, Dehghani, Minderer, Heigold, Gelly, Uszkoreit and
  Houlsby}]{dosovitskiy2021image}
{Dosovitskiy, A.}, {Beyer, L.}, {Kolesnikov, A.}, {Weissenborn, D.}, {Zhai,
  X.}, {Unterthiner, T.}, {Dehghani, M.}, {Minderer, M.}, {Heigold, G.},
  {Gelly, S.}, {Uszkoreit, J.} and {Houlsby, N.} (2021).
\newblock An image is worth 16x16 words: Transformers for image recognition at
  scale.
\newblock \eprint{2010.11929},
  \urlprefix\url{https://arxiv.org/abs/2010.11929}.

\bibitem[{Düben et~al.(2019)Düben, Leutbecher and Bauer}]{duben2019new}
{Düben, P.~D.}, {Leutbecher, M.} and {Bauer, P.} (2019).
\newblock New methods for data storage of model output from ensemble
  simulations.
\newblock \textit{Monthly Weather Review}, {147}, 677 -- 689.

\bibitem[{Eisenberger et~al.(2022)Eisenberger, Toker, Leal-Taix\'e, Bernard and
  Cremers}]{eisenberger2022unified}
{Eisenberger, M.}, {Toker, A.}, {Leal-Taix\'e, L.}, {Bernard, F.} and {Cremers,
  D.} (2022).
\newblock A unified framework for implicit sinkhorn differentiation.
\newblock In \textit{Proceedings of the IEEE/CVF Conference on Computer Vision
  and Pattern Recognition (CVPR)}. 509--518.

\bibitem[{Feik et~al.(2024)Feik, Lerch and Stühmer}]{FeikEtAl2024}
{Feik, M.}, {Lerch, S.} and {Stühmer, J.} (2024).
\newblock Graph neural networks and spatial information learning for
  post-processing ensemble weather forecasts.
\newblock International Conference on Machine Learning 2024 - Machine Learning
  for Earth System Modeling Workshop., \eprint{2407.11050},
  \urlprefix\url{https://arxiv.org/abs/2407.11050}.

\bibitem[{Frogner et~al.(2015)Frogner, Zhang, Mobahi, Araya-Polo and
  Poggio}]{frogner2015learning}
{Frogner, C.}, {Zhang, C.}, {Mobahi, H.}, {Araya-Polo, M.} and {Poggio, T.}
  (2015).
\newblock Learning with a wasserstein loss.
\newblock In \textit{Proceedings of the 29th International Conference on Neural
  Information Processing Systems - Volume 2}. NIPS'15, MIT Press, Cambridge,
  MA, USA, 2053–2061.

\bibitem[{Gilleland et~al.(2010)Gilleland, Ahijevych, Brown and
  Ebert}]{gilleland2010verifying}
{Gilleland, E.}, {Ahijevych, D.~A.}, {Brown, B.~G.} and {Ebert, E.~E.} (2010).
\newblock Verifying forecasts spatially.
\newblock \textit{Bulletin of the American Meteorological Society}, {91}, 1365
  -- 1376.

\bibitem[{Gondara(2016)}]{gondara2016medical}
{Gondara, L.} (2016).
\newblock Medical image denoising using convolutional denoising autoencoders.
\newblock In \textit{2016 IEEE 16th International Conference on Data Mining
  Workshops (ICDMW)}. 241--246.

\bibitem[{Grams et~al.(2017)Grams, Beerli, Pfenninger, Staffell and
  Wernli}]{grams2017balancing}
{Grams, C.~M.}, {Beerli, R.}, {Pfenninger, S.}, {Staffell, I.} and {Wernli, H.}
  (2017).
\newblock Balancing europe’s wind-power output through spatial deployment
  informed by weather regimes.
\newblock \textit{Nature Climate Change}, {7}, 557--562.

\bibitem[{Grooms(2021)}]{analog2021grooms}
{Grooms, I.} (2021).
\newblock Analog ensemble data assimilation and a method for constructing
  analogs with variational autoencoders.
\newblock \textit{Quarterly Journal of the Royal Meteorological Society},
  {147}, 139--149.

\bibitem[{Grönquist et~al.(2021)Grönquist, Yao, Ben-Nun, Dryden, Dueben, Li
  and Hoefler}]{gronquist2021}
{Grönquist, P.}, {Yao, C.}, {Ben-Nun, T.}, {Dryden, N.}, {Dueben, P.}, {Li,
  S.} and {Hoefler, T.} (2021).
\newblock Deep learning for post-processing ensemble weather forecasts.
\newblock \textit{Philosophical Transactions of the Royal Society A:
  Mathematical, Physical and Engineering Sciences}, {379}, 20200092.

\bibitem[{He et~al.(2016)He, Zhang, Ren and Sun}]{he2016deep}
{He, K.}, {Zhang, X.}, {Ren, S.} and {Sun, J.} (2016).
\newblock Deep residual learning for image recognition.
\newblock In \textit{Proceedings of the IEEE Conference on Computer Vision and
  Pattern Recognition (CVPR)}.

\bibitem[{Hinton and Salakhutdinov(2006)}]{hinton2006reducing}
{Hinton, G.~E.} and {Salakhutdinov, R.~R.} (2006).
\newblock Reducing the dimensionality of data with neural networks.
\newblock \textit{Science}, {313}, 504--507.
\newblock Publisher: American Association for the Advancement of Science.

\bibitem[{Hinton and Zemel(1993)}]{hinton1993autoencoders}
{Hinton, G.~E.} and {Zemel, R.} (1993).
\newblock Autoencoders, minimum description length and helmholtz free energy.
\newblock In \textit{Advances in Neural Information Processing Systems}
  (J.~Cowan, G.~Tesauro and J.~Alspector, eds.), vol.~6. Morgan-Kaufmann.

\bibitem[{Horat and Lerch(2024)}]{HoratLerch2024}
{Horat, N.} and {Lerch, S.} (2024).
\newblock Deep learning for postprocessing global probabilistic forecasts on
  subseasonal time scales.
\newblock \textit{Monthly Weather Review}, {152}, 667--687.

\bibitem[{Höhlein et~al.(2024)Höhlein, Schulz, Westermann and
  Lerch}]{hohlein_postprocessing_2024}
{Höhlein, K.}, {Schulz, B.}, {Westermann, R.} and {Lerch, S.} (2024).
\newblock Postprocessing of ensemble weather forecasts using
  permutation-invariant neural networks.
\newblock \textit{Artificial Intelligence for the Earth Systems}, {3}, e230070.

\bibitem[{Höhlein et~al.(2022)Höhlein, Weiss, Necker, Weissmann, Miyoshi and
  Westermann}]{hoehlein2022evaluation}
{Höhlein, K.}, {Weiss, S.}, {Necker, T.}, {Weissmann, M.}, {Miyoshi, T.} and
  {Westermann, R.} (2022).
\newblock {Evaluation of Volume Representation Networks for Meteorological
  Ensemble Compression}.
\newblock In \textit{Vision, Modeling, and Visualization} (J.~Bender, M.~Botsch
  and D.~A. Keim, eds.). The Eurographics Association.

\bibitem[{Jolliffe and Cadima(2016)}]{jolliffe2016principal}
{Jolliffe, I.~T.} and {Cadima, J.} (2016).
\newblock Principal component analysis: a review and recent developments.
\newblock \textit{Philosophical Transactions of the Royal Society A:
  Mathematical, Physical and Engineering Sciences}, {374}, 20150202.

\bibitem[{Kantorovich(1960)}]{kantorovich1960mathematical}
{Kantorovich, L.~V.} (1960).
\newblock Mathematical methods of organizing and planning production.
\newblock \textit{Management science}, {6}, 366--422.

\bibitem[{Kiefer et~al.(2023)Kiefer, Lerch, Ludwig and Pinto}]{Kiefer2023}
{Kiefer, S.~M.}, {Lerch, S.}, {Ludwig, P.} and {Pinto, J.~G.} (2023).
\newblock Can machine learning models be a suitable tool for predicting central
  european cold winter weather on subseasonal to seasonal time scales?
\newblock \textit{Artificial Intelligence for the Earth Systems}, {2}, e230020.

\bibitem[{Kiefer et~al.(2024)Kiefer, Lerch, Ludwig and
  Pinto}]{kiefer2024random}
{Kiefer, S.~M.}, {Lerch, S.}, {Ludwig, P.} and {Pinto, J.~G.} (2024).
\newblock Random forests’ postprocessing capability of enhancing predictive
  skill on subseasonal time scales — a flow-dependent view on central
  european winter weather.
\newblock \textit{Artificial Intelligence for the Earth Systems}, {3}, e240014.

\bibitem[{Kingma(2013)}]{kingma2022autoencoding}
{Kingma, D.~P.} (2013).
\newblock Auto-encoding variational bayes.
\newblock \eprint{1312.6114}, \urlprefix\url{https://arxiv.org/abs/1312.6114}.

\bibitem[{Kingma and Welling(2019)}]{kingma2019foundations}
{Kingma, D.~P.} and {Welling, M.} (2019).
\newblock An introduction to variational autoencoders.
\newblock \textit{Found. Trends Mach. Learn.}, {12}, 307–392.

\bibitem[{Kochkov et~al.(2024)Kochkov, Yuval, Langmore, Norgaard, Smith,
  Mooers, Klöwer, Lottes, Rasp, Düben, Hatfield, Battaglia, Sanchez-Gonzalez,
  Willson, Brenner and Hoyer}]{NeuralGCM}
{Kochkov, D.}, {Yuval, J.}, {Langmore, I.}, {Norgaard, P.}, {Smith, J.},
  {Mooers, G.}, {Klöwer, M.}, {Lottes, J.}, {Rasp, S.}, {Düben, P.},
  {Hatfield, S.}, {Battaglia, P.}, {Sanchez-Gonzalez, A.}, {Willson, M.},
  {Brenner, M.~P.} and {Hoyer, S.} (2024).
\newblock Neural general circulation models for weather and climate.
\newblock \textit{Nature}, {632}, 1060--1066.

\bibitem[{Kolouri et~al.(2017)Kolouri, Park, Thorpe, Slepcev and
  Rohde}]{kolouri2017optimal}
{Kolouri, S.}, {Park, S.~R.}, {Thorpe, M.}, {Slepcev, D.} and {Rohde, G.~K.}
  (2017).
\newblock Optimal mass transport: Signal processing and machine-learning
  applications.
\newblock \textit{IEEE signal processing magazine}, {34}, 43--59.

\bibitem[{Kramer(1991)}]{kramer1991nonlinear}
{Kramer, M.~A.} (1991).
\newblock Nonlinear principal component analysis using autoassociative neural
  networks.
\newblock \textit{AIChE Journal}, {37}, 233--243.

\bibitem[{Kwok and Tsang(2004)}]{kwok2004preimage}
{Kwok, J.-Y.} and {Tsang, I.-H.} (2004).
\newblock The pre-image problem in kernel methods.
\newblock \textit{IEEE Transactions on Neural Networks}, {15}, 1517--1525.

\bibitem[{Larsen et~al.(2016)Larsen, Sønderby, Larochelle and
  Winther}]{larsen2016autoencoding}
{Larsen, A. B.~L.}, {Sønderby, S.~K.}, {Larochelle, H.} and {Winther, O.}
  (2016).
\newblock Autoencoding beyond pixels using a learned similarity metric.
\newblock In \textit{Proceedings of The 33rd International Conference on
  Machine Learning} (M.~F. Balcan and K.~Q. Weinberger, eds.), vol.~48 of
  \textit{Proceedings of Machine Learning Research}. PMLR, New York, New York,
  USA, 1558--1566.

\bibitem[{Lerch and Polsterer(2022)}]{lerch2022convolutional}
{Lerch, S.} and {Polsterer, K.~L.} (2022).
\newblock Convolutional autoencoders for spatially-informed ensemble
  post-processing.
\newblock International Conference on Learning Representations (ICLR) 2022 - AI
  for Earth and Space Science Workshop., \eprint{2204.05102},
  \urlprefix\url{https://arxiv.org/abs/2204.05102}.

\bibitem[{Li et~al.(2018)Li, Jamieson, DeSalvo, Rostamizadeh and
  Talwalkar}]{li2018hyperband}
{Li, L.}, {Jamieson, K.}, {DeSalvo, G.}, {Rostamizadeh, A.} and {Talwalkar, A.}
  (2018).
\newblock Hyperband: A novel bandit-based approach to hyperparameter
  optimization.
\newblock \textit{Journal of Machine Learning Research}, {18}, 1--52.

\bibitem[{Liaw et~al.(2018)Liaw, Liang, Nishihara, Moritz, Gonzalez and
  Stoica}]{liaw2018tune}
{Liaw, R.}, {Liang, E.}, {Nishihara, R.}, {Moritz, P.}, {Gonzalez, J.~E.} and
  {Stoica, I.} (2018).
\newblock Tune: A research platform for distributed model selection and
  training.
\newblock \eprint{1807.05118},
  \urlprefix\url{https://arxiv.org/abs/1807.05118}.

\bibitem[{Loshchilov and Hutter(2019)}]{loshchilov2019decoupled}
{Loshchilov, I.} and {Hutter, F.} (2019).
\newblock Decoupled weight decay regularization.
\newblock \eprint{1711.05101},
  \urlprefix\url{https://arxiv.org/abs/1711.05101}.

\bibitem[{Lucas et~al.(2019)Lucas, Tucker, Grosse and
  Norouzi}]{lucas2019understanding}
{Lucas, J.}, {Tucker, G.}, {Grosse, R.} and {Norouzi, M.} (2019).
\newblock Understanding posterior collapse in generative latent variable
  models.
\newblock \urlprefix\url{https://openreview.net/forum?id=r1xaVLUYuE}.

\bibitem[{Mahesh et~al.(2024)Mahesh, Collins, Bonev, Brenowitz, Cohen, Elms,
  Harrington, Kashinath, Kurth, North, OBrien, Pritchard, Pruitt, Risser,
  Subramanian and Willard}]{MaheshEtAl2024}
{Mahesh, A.}, {Collins, W.}, {Bonev, B.}, {Brenowitz, N.}, {Cohen, Y.}, {Elms,
  J.}, {Harrington, P.}, {Kashinath, K.}, {Kurth, T.}, {North, J.}, {OBrien,
  T.}, {Pritchard, M.}, {Pruitt, D.}, {Risser, M.}, {Subramanian, S.} and
  {Willard, J.} (2024).
\newblock Huge ensembles part i: Design of ensemble weather forecasts using
  spherical fourier neural operators.
\newblock \eprint{2408.03100},
  \urlprefix\url{https://arxiv.org/abs/2408.03100}.

\bibitem[{Mika et~al.(1998)Mika, Sch\"{o}lkopf, Smola, M\"{u}ller, Scholz and
  R\"{a}tsch}]{mika1998kernel}
{Mika, S.}, {Sch\"{o}lkopf, B.}, {Smola, A.}, {M\"{u}ller, K.-R.}, {Scholz, M.}
  and {R\"{a}tsch, G.} (1998).
\newblock Kernel pca and de-noising in feature spaces.
\newblock In \textit{Advances in Neural Information Processing Systems}
  (M.~Kearns, S.~Solla and D.~Cohn, eds.), vol.~11. MIT Press.

\bibitem[{Mockert et~al.(2024)Mockert, Grams, Lerch, Osman and
  Quinting}]{Mockert2024}
{Mockert, F.}, {Grams, C.~M.}, {Lerch, S.}, {Osman, M.} and {Quinting, J.}
  (2024).
\newblock Multivariate post-processing of probabilistic sub-seasonal weather
  regime forecasts.
\newblock \textit{Quarterly Journal of the Royal Meteorological Society},
  {150}, 4771--4787.

\bibitem[{Patrini et~al.(2020)Patrini, van~den Berg, Forr{\'{e}}, Carioni,
  Bhargav, Welling, Genewein and Nielsen}]{patrini2020sinkhorn}
{Patrini, G.}, {van~den Berg, R.}, {Forr{\'{e}}, P.}, {Carioni, M.}, {Bhargav,
  S.}, {Welling, M.}, {Genewein, T.} and {Nielsen, F.} (2020).
\newblock Sinkhorn autoencoders.
\newblock In \textit{Proceedings of The 35th Uncertainty in Artificial
  Intelligence Conference} (R.~P. Adams and V.~Gogate, eds.), vol. 115 of
  \textit{Proceedings of Machine Learning Research}. PMLR, 733--743.

\bibitem[{Pearson(1901)}]{pearson1901pca}
{Pearson, K.} (1901).
\newblock Liii. on lines and planes of closest fit to systems of points in
  space.
\newblock \textit{The London, Edinburgh, and Dublin Philosophical Magazine and
  Journal of Science}, {2}, 559--572.

\bibitem[{Pedregosa et~al.(2011)Pedregosa, Varoquaux, Gramfort, Michel,
  Thirion, Grisel, Blondel, Prettenhofer, Weiss, Dubourg, Vanderplas, Passos,
  Cournapeau, Brucher, Perrot and Duchesnay}]{scikitlearn}
{Pedregosa, F.}, {Varoquaux, G.}, {Gramfort, A.}, {Michel, V.}, {Thirion, B.},
  {Grisel, O.}, {Blondel, M.}, {Prettenhofer, P.}, {Weiss, R.}, {Dubourg, V.},
  {Vanderplas, J.}, {Passos, A.}, {Cournapeau, D.}, {Brucher, M.}, {Perrot, M.}
  and {Duchesnay, E.} (2011).
\newblock Scikit-learn: Machine learning in {P}ython.
\newblock \textit{Journal of Machine Learning Research}, {12}, 2825--2830.

\bibitem[{Price et~al.(2024)Price, Sanchez-Gonzalez, Alet, Andersson, El-Kadi,
  Masters, Ewalds, Stott, Mohamed, Battaglia, Lam and Willson}]{GenCast}
{Price, I.}, {Sanchez-Gonzalez, A.}, {Alet, F.}, {Andersson, T.~R.}, {El-Kadi,
  A.}, {Masters, D.}, {Ewalds, T.}, {Stott, J.}, {Mohamed, S.}, {Battaglia,
  P.}, {Lam, R.} and {Willson, M.} (2024).
\newblock Gencast: Diffusion-based ensemble forecasting for medium-range
  weather.
\newblock \eprint{2312.15796},
  \urlprefix\url{https://arxiv.org/abs/2312.15796}.

\bibitem[{Pu et~al.(2016)Pu, Gan, Henao, Yuan, Li, Stevens and
  Carin}]{pu2016variational}
{Pu, Y.}, {Gan, Z.}, {Henao, R.}, {Yuan, X.}, {Li, C.}, {Stevens, A.} and
  {Carin, L.} (2016).
\newblock Variational autoencoder for deep learning of images, labels and
  captions.
\newblock In \textit{Advances in Neural Information Processing Systems}
  (D.~Lee, M.~Sugiyama, U.~Luxburg, I.~Guyon and R.~Garnett, eds.), vol.~29.
  Curran Associates, Inc.

\bibitem[{Rasp and Lerch(2018)}]{rasp2018neural}
{Rasp, S.} and {Lerch, S.} (2018).
\newblock Neural networks for postprocessing ensemble weather forecasts.
\newblock \textit{Monthly Weather Review}, {146}, 3885--3900.

\bibitem[{Rizzo and Székely(2016)}]{rizzo2016energy}
{Rizzo, M.~L.} and {Székely, G.~J.} (2016).
\newblock Energy distance.
\newblock \textit{WIREs Computational Statistics}, {8}, 27--38.

\bibitem[{Roberts and Lean(2008)}]{nigel2008scale}
{Roberts, N.~M.} and {Lean, H.~W.} (2008).
\newblock Scale-selective verification of rainfall accumulations from
  high-resolution forecasts of convective events.
\newblock \textit{Monthly Weather Review}, {136}, 78 -- 97.

\bibitem[{Rodwell et~al.(2018)Rodwell, Richardson, Parsons and
  Wernli}]{rodwell2018flow}
{Rodwell, M.~J.}, {Richardson, D.~S.}, {Parsons, D.~B.} and {Wernli, H.}
  (2018).
\newblock Flow-dependent reliability: A path to more skillful ensemble
  forecasts.
\newblock \textit{Bulletin of the American Meteorological Society}, {99}, 1015
  -- 1026.

\bibitem[{Roweis and Saul(2000)}]{roweis2000nonlinear}
{Roweis, S.~T.} and {Saul, L.~K.} (2000).
\newblock Nonlinear dimensionality reduction by locally linear embedding.
\newblock \textit{Science}, {290}, 2323--2326.

\bibitem[{Sakurada and Yairi(2014)}]{sakurada2014anomaly}
{Sakurada, M.} and {Yairi, T.} (2014).
\newblock Anomaly detection using autoencoders with nonlinear dimensionality
  reduction.
\newblock In \textit{Proceedings of the MLSDA 2014 2nd Workshop on Machine
  Learning for Sensory Data Analysis}. MLSDA'14, Association for Computing
  Machinery, New York, NY, USA, 4–11.

\bibitem[{Scheuerer et~al.(2024)Scheuerer, Heinrich-Mertsching, Bahaga,
  Gudoshava and Thorarinsdottir}]{ScheuererEtAl2024}
{Scheuerer, M.}, {Heinrich-Mertsching, C.}, {Bahaga, T.~K.}, {Gudoshava, M.}
  and {Thorarinsdottir, T.~L.} (2024).
\newblock Applications of machine learning to predict seasonal precipitation
  for east africa.
\newblock \eprint{2409.06238},
  \urlprefix\url{https://arxiv.org/abs/2409.06238}.

\bibitem[{Scheuerer et~al.(2020)Scheuerer, Switanek, Worsnop and
  Hamill}]{Scheuerer2020}
{Scheuerer, M.}, {Switanek, M.~B.}, {Worsnop, R.~P.} and {Hamill, T.~M.}
  (2020).
\newblock Using artificial neural networks for generating probabilistic
  subseasonal precipitation forecasts over {California}.
\newblock \textit{Monthly Weather Review}, {148}, 3489--3506.

\bibitem[{Sch{\"o}lkopf et~al.(1997)Sch{\"o}lkopf, Smola and
  M{\"u}ller}]{schoelkopf1997kernel}
{Sch{\"o}lkopf, B.}, {Smola, A.} and {M{\"u}ller, K.-R.} (1997).
\newblock Kernel principal component analysis.
\newblock In \textit{Artificial Neural Networks --- ICANN'97} (W.~Gerstner,
  A.~Germond, M.~Hasler and J.-D. Nicoud, eds.). Springer Berlin Heidelberg,
  Berlin, Heidelberg, 583--588.

\bibitem[{Schulz and Lerch(2022)}]{schulz2022machine}
{Schulz, B.} and {Lerch, S.} (2022).
\newblock Machine learning methods for postprocessing ensemble forecasts of
  wind gusts: A systematic comparison.
\newblock \textit{Monthly Weather Review}, {150}, 235 -- 257.

\bibitem[{Székely and Rizzo(2013)}]{szekely2013energy}
{Székely, G.~J.} and {Rizzo, M.~L.} (2013).
\newblock Energy statistics: A class of statistics based on distances.
\newblock \textit{Journal of Statistical Planning and Inference}, {143},
  1249--1272.

\bibitem[{Tenenbaum et~al.(2000)Tenenbaum, de~Silva and
  Langford}]{joshua2000global}
{Tenenbaum, J.~B.}, {de~Silva, V.} and {Langford, J.~C.} (2000).
\newblock A global geometric framework for nonlinear dimensionality reduction.
\newblock \textit{Science}, {290}, 2319--2323.

\bibitem[{Thorarinsdottir et~al.(2013)Thorarinsdottir, Gneiting and
  Gissibl}]{thorarinsdottir2013using}
{Thorarinsdottir, T.~L.}, {Gneiting, T.} and {Gissibl, N.} (2013).
\newblock Using proper divergence functions to evaluate climate models.
\newblock \textit{SIAM/ASA Journal on Uncertainty Quantification}, {1},
  522--534.

\bibitem[{Tolstikhin et~al.(2019)Tolstikhin, Bousquet, Gelly and
  Schoelkopf}]{tolstikhin2017wasserstein}
{Tolstikhin, I.}, {Bousquet, O.}, {Gelly, S.} and {Schoelkopf, B.} (2019).
\newblock Wasserstein auto-encoders.
\newblock \eprint{1711.01558},
  \urlprefix\url{https://arxiv.org/abs/1711.01558}.

\bibitem[{Vannitsem et~al.(2021)Vannitsem, Bremnes, Demaeyer, Evans, Flowerdew,
  Hemri, Lerch, Roberts, Theis, Atencia, Bouallègue, Bhend, Dabernig, Cruz,
  Hieta, Mestre, Moret, Plenković, Schmeits, Taillardat, den Bergh,
  Schaeybroeck, Whan and Ylhaisi}]{vannitsem2021statistical}
{Vannitsem, S.}, {Bremnes, J.~B.}, {Demaeyer, J.}, {Evans, G.~R.}, {Flowerdew,
  J.}, {Hemri, S.}, {Lerch, S.}, {Roberts, N.}, {Theis, S.}, {Atencia, A.},
  {Bouallègue, Z.~B.}, {Bhend, J.}, {Dabernig, M.}, {Cruz, L.~D.}, {Hieta,
  L.}, {Mestre, O.}, {Moret, L.}, {Plenković, I.~O.}, {Schmeits, M.},
  {Taillardat, M.}, {den Bergh, J.~V.}, {Schaeybroeck, B.~V.}, {Whan, K.} and
  {Ylhaisi, J.} (2021).
\newblock Statistical postprocessing for weather forecasts: Review, challenges,
  and avenues in a big data world.
\newblock \textit{Bulletin of the American Meteorological Society}, {102},
  E681--E699.

\bibitem[{Veldkamp et~al.(2021)Veldkamp, Whan, Dirksen and
  Schmeits}]{Veldkamp2021wind}
{Veldkamp, S.}, {Whan, K.}, {Dirksen, S.} and {Schmeits, M.} (2021).
\newblock Statistical postprocessing of wind speed forecasts using
  convolutional neural networks.
\newblock \textit{Monthly Weather Review}, {149}, 1141--1152.

\bibitem[{Wang et~al.(2014)Wang, Huang, Wang and Wang}]{Wang2014generalized}
{Wang, W.}, {Huang, Y.}, {Wang, Y.} and {Wang, L.} (2014).
\newblock Generalized autoencoder: A neural network framework for
  dimensionality reduction.
\newblock In \textit{Proceedings of the IEEE Conference on Computer Vision and
  Pattern Recognition (CVPR) Workshops}.

\bibitem[{Wang et~al.(2016)Wang, Yao and Zhao}]{wang2016autoencoder}
{Wang, Y.}, {Yao, H.} and {Zhao, S.} (2016).
\newblock Auto-encoder based dimensionality reduction.
\newblock \textit{Neurocomputing}, {184}, 232--242.

\bibitem[{Wang et~al.(2004)Wang, Bovik, Sheikh and Simoncelli}]{wang2004image}
{Wang, Z.}, {Bovik, A.}, {Sheikh, H.} and {Simoncelli, E.} (2004).
\newblock Image quality assessment: from error visibility to structural
  similarity.
\newblock \textit{IEEE Transactions on Image Processing}, {13}, 600--612.

\bibitem[{Yang and Grooms(2021)}]{machine2021yang}
{Yang, L.~M.} and {Grooms, I.} (2021).
\newblock Machine learning techniques to construct patched analog ensembles for
  data assimilation.
\newblock \textit{Journal of Computational Physics}, {443}, 110532.

\bibitem[{Zaheer et~al.(2017)Zaheer, Kottur, Ravanbakhsh, Poczos, Salakhutdinov
  and Smola}]{zaheer2017deep}
{Zaheer, M.}, {Kottur, S.}, {Ravanbakhsh, S.}, {Poczos, B.}, {Salakhutdinov,
  R.~R.} and {Smola, A.~J.} (2017).
\newblock Deep sets.
\newblock In \textit{Advances in Neural Information Processing Systems}
  (I.~Guyon, U.~V. Luxburg, S.~Bengio, H.~Wallach, R.~Fergus, S.~Vishwanathan
  and R.~Garnett, eds.), vol.~30. Curran Associates, Inc.

\bibitem[{Zhong et~al.(2024)Zhong, Chen, Li, Liu, Fan, Feng, Dai, Luo, Wu and
  Lu}]{FuXiENS}
{Zhong, X.}, {Chen, L.}, {Li, H.}, {Liu, J.}, {Fan, X.}, {Feng, J.}, {Dai, K.},
  {Luo, J.-J.}, {Wu, J.} and {Lu, B.} (2024).
\newblock Fuxi-ens: A machine learning model for medium-range ensemble weather
  forecasting.
\newblock \eprint{2405.05925},
  \urlprefix\url{https://arxiv.org/abs/2405.05925}.

\bibitem[{Zhou and Paffenroth(2017)}]{zhou2017anomaly}
{Zhou, C.} and {Paffenroth, R.~C.} (2017).
\newblock Anomaly detection with robust deep autoencoders.
\newblock In \textit{Proceedings of the 23rd ACM SIGKDD International
  Conference on Knowledge Discovery and Data Mining}. KDD '17, Association for
  Computing Machinery, New York, NY, USA, 665–674.

\end{thebibliography}

\newpage

\setcounter{figure}{0}
\renewcommand{\figurename}{Fig.}
\renewcommand{\thefigure}{S\arabic{figure}}

\section*{Supplement to ``Low-dimensional representation of ensemble forecast fields using autoencoder-based methods"}

This Supplemental Material provides additional evaluation results for the three dimensionality reduction approaches presented in the main paper.

\begin{figure}[h]
	\centering
	\includegraphics[width = 0.65\textwidth]{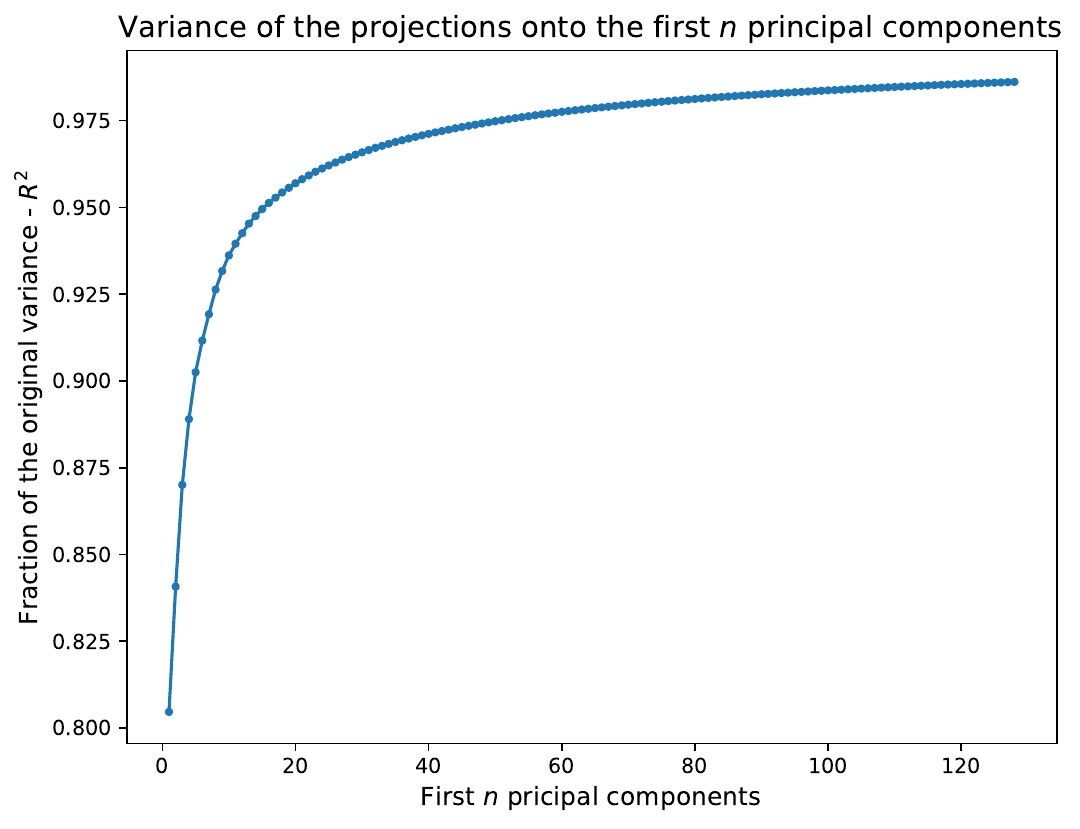}
	\caption{Proportion of variance explained depending on the number of retained principal components in PCA applied to temperature data.}
	\label{fig_pca_r2}
\end{figure}

Figure \ref{fig_pca_r2} illustrates the proportion of variance explained by PCA with different choices of retained principal components. The first 5 principal components account for more than 90\% of the variance in the original data, while the first 16 principal components account for more than 95\%. This observation helps to explain the strong performance of the PCA-based approach when the latent dimension is large, as discussed in Section 4.2 of the main paper.

\begin{figure}
	\centering
	\includegraphics[width=\textwidth]{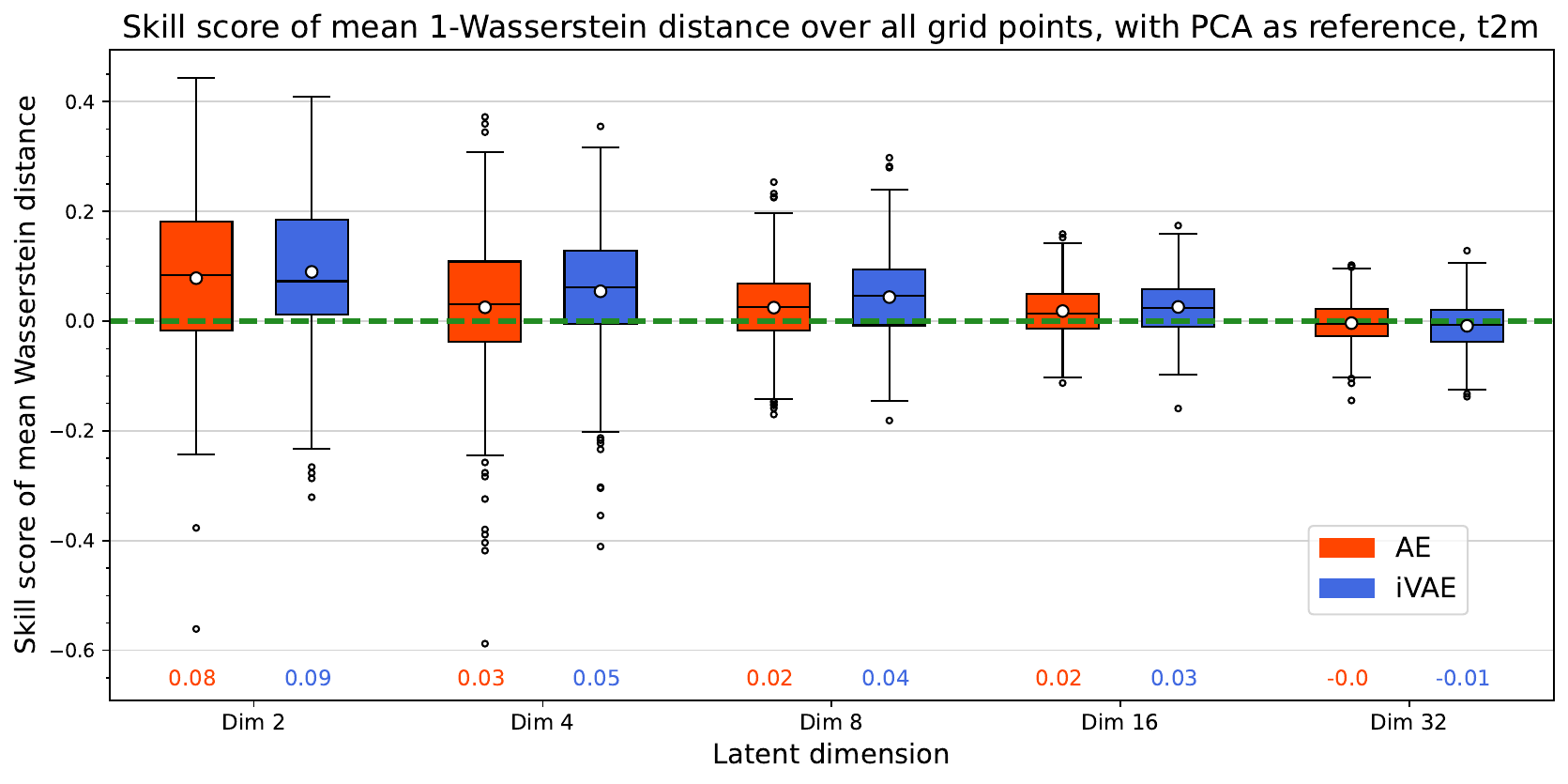}
	\includegraphics[width=\textwidth]{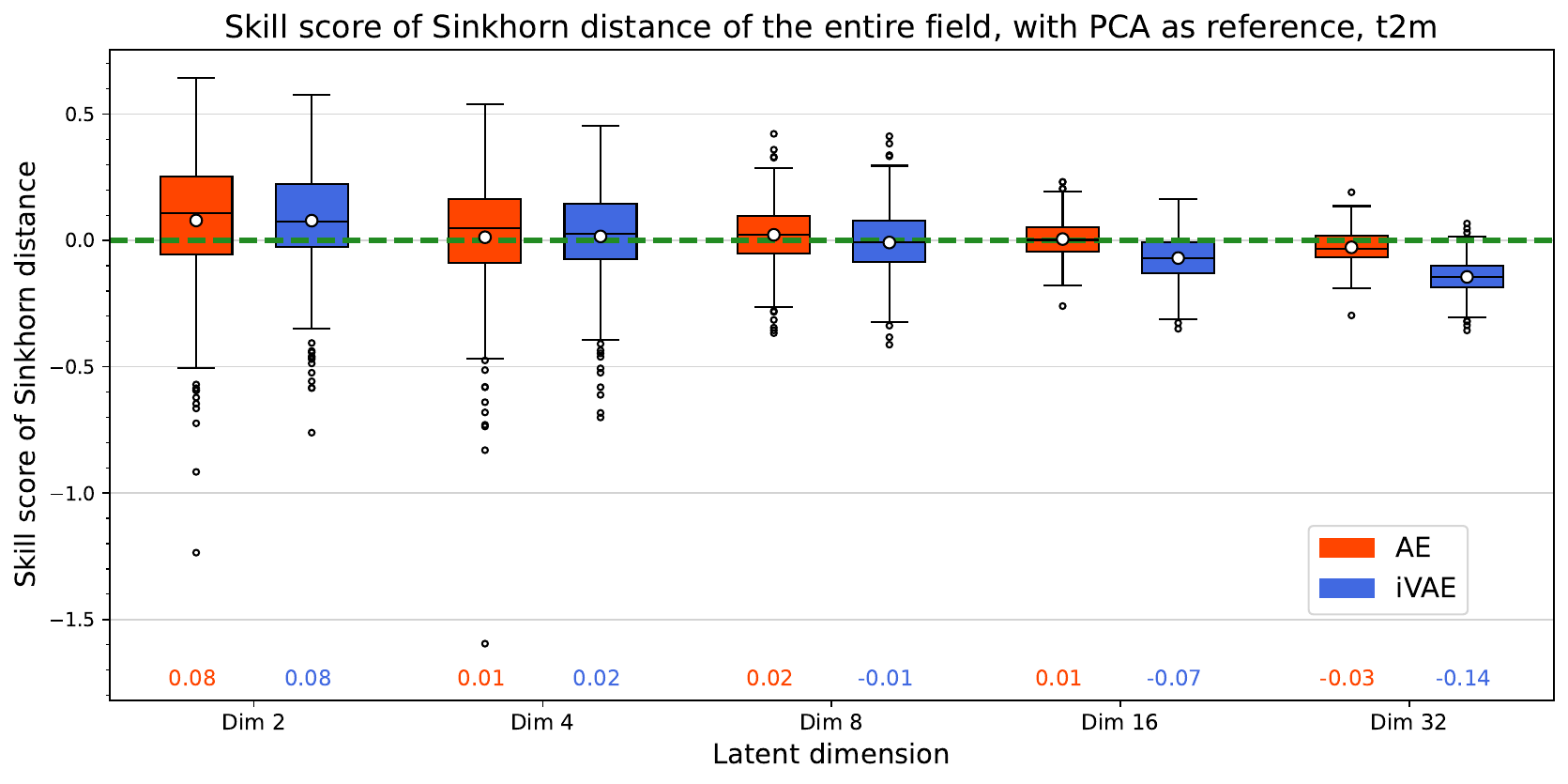}
	\caption{Boxplots of skill scores based on optimal transportation distances between the input and reconstructed ensemble fields over the 366 days in the test set for temperature data. The panels show mean univariate 1-Wasserstein distances over all grid points (top) and multivariate Sinkhorn distances computed for the entire fields (bottom). PCA-based approach shown in green dashed line is the reference method. The respective mean skill values are indicated below each box.}
	\label{fig_wd_box_t}
\end{figure}

\begin{figure}
	\centering
	\includegraphics[width=\textwidth]{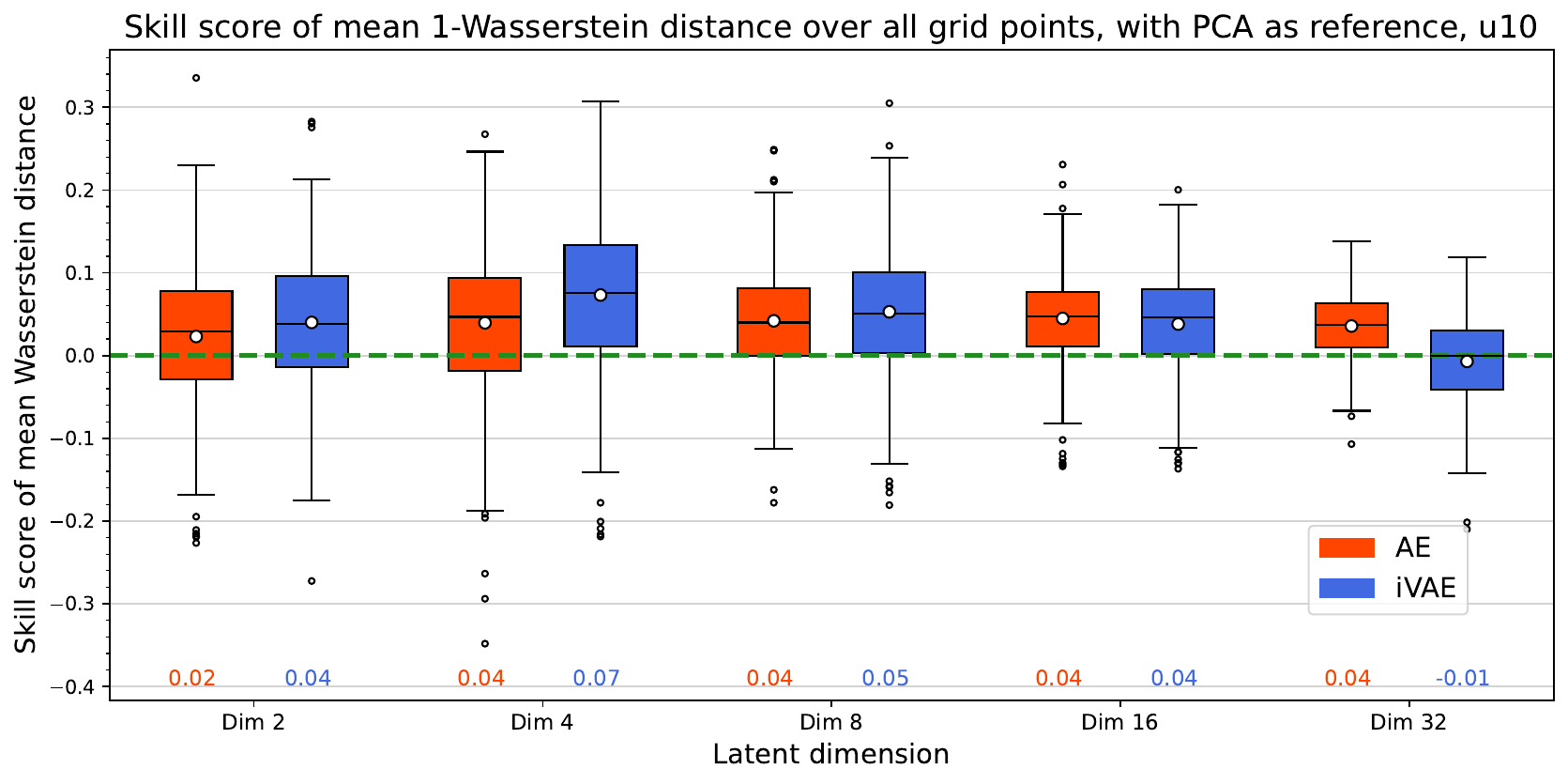}
	\includegraphics[width=\textwidth]{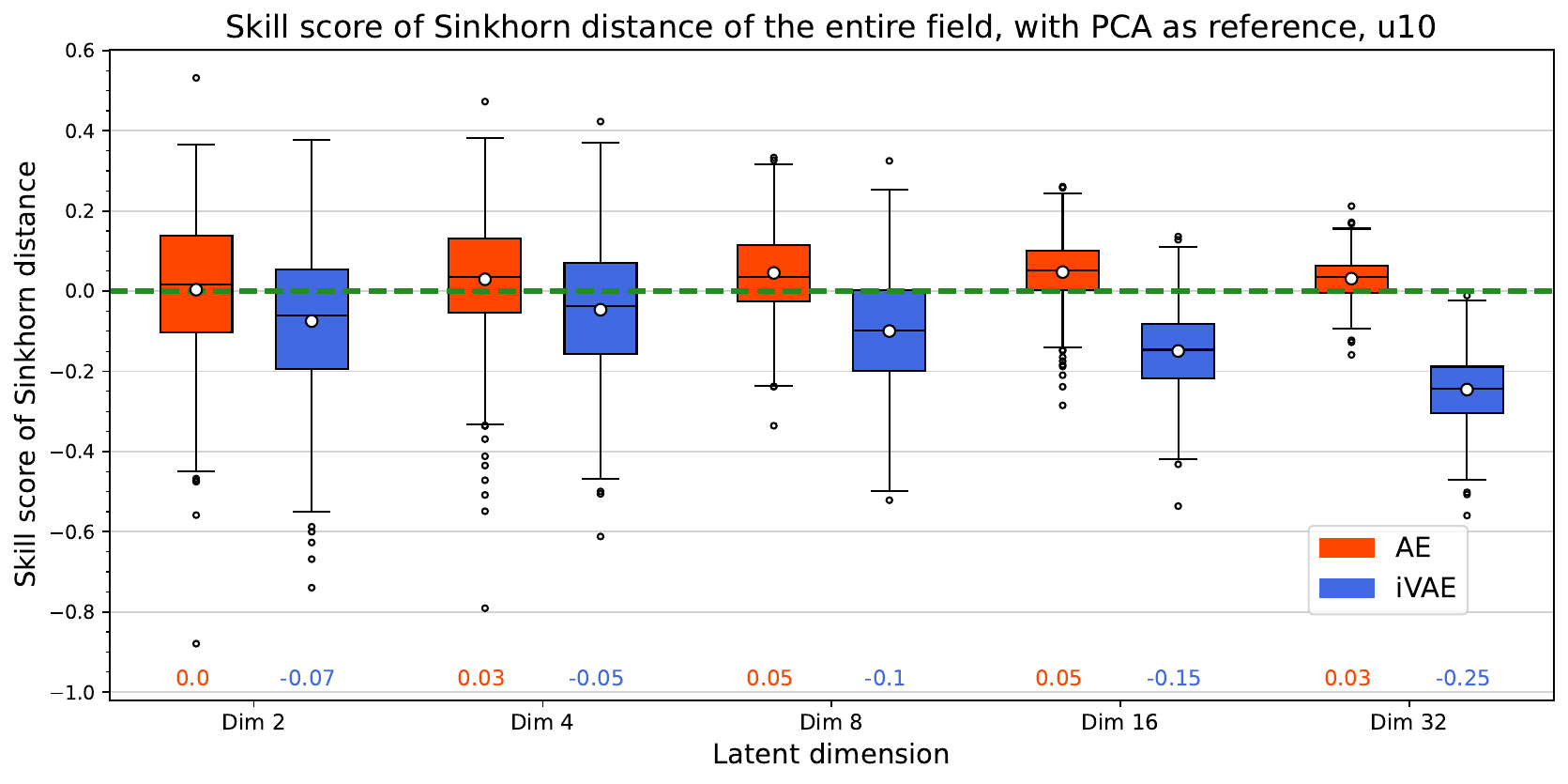}
	\caption{As Figure \ref{fig_wd_box_t}, but for the U component of 10-m wind speed.}
	\label{fig_wd_box_u}
\end{figure}

The evaluation results for optimal transport distances for temperature and the U component of wind speed data are presented in Figures \ref{fig_wd_box_t} and \ref{fig_wd_box_u}.
We observe slightly worse performance from the two neural network-based approaches for temperature data, where both methods fail to outperform the PCA-based method with a high latent dimension of 32. Notably, the iVAE method performs the worst in terms of Sinkhorn distance, despite incorporating the Sinkhorn distance component in its loss function.
Conversely, both neural network-based approaches demonstrate significantly better performance for wind speed data, consistent with our findings on energy distances in the main paper.

\begin{figure}
	\centering
	\includegraphics[width=\textwidth]{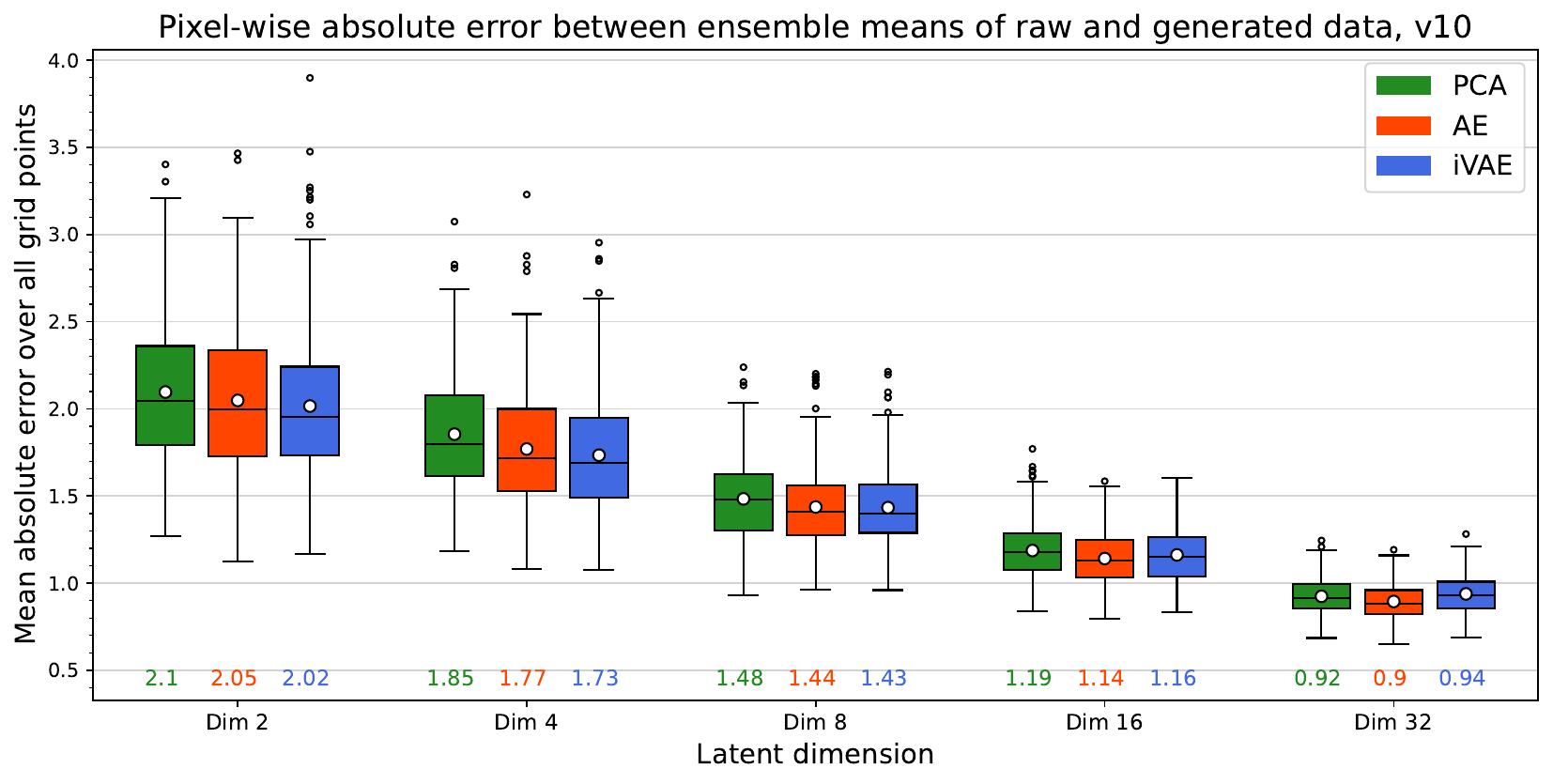}
	\includegraphics[width=\textwidth]{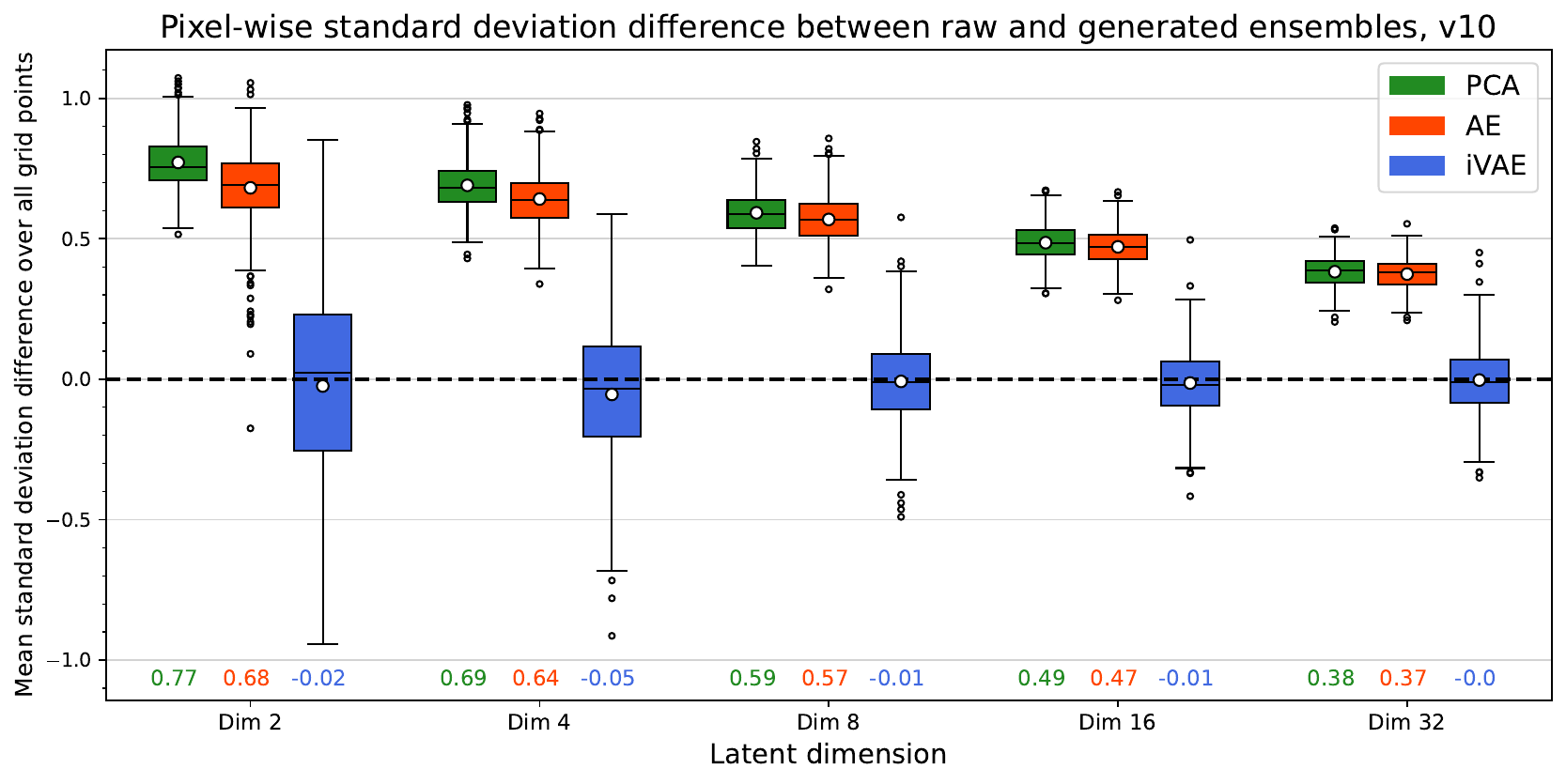}
	\caption{Boxplots of absolute differences between the mean values of input and reconstructed ensemble fields (top) and differences between the standard deviations of input and reconstructed ensemble fields (bottom) at each grid point. 
		Boxes show the variability over 366 days in the test set of different methods for the V component of 10-m wind speed data, considering 5 different dimensionalities of the latent representation.
		The mean values of the (absolute) differences are indicated below each box. The differences between the standard deviations are computed such that negative values indicate a larger variability of the reconstructed ensemble compared to the input ensemble.}
	\label{fig_mae_std_box_v}
\end{figure}

\begin{figure}
	\centering
	\includegraphics[width=\textwidth]{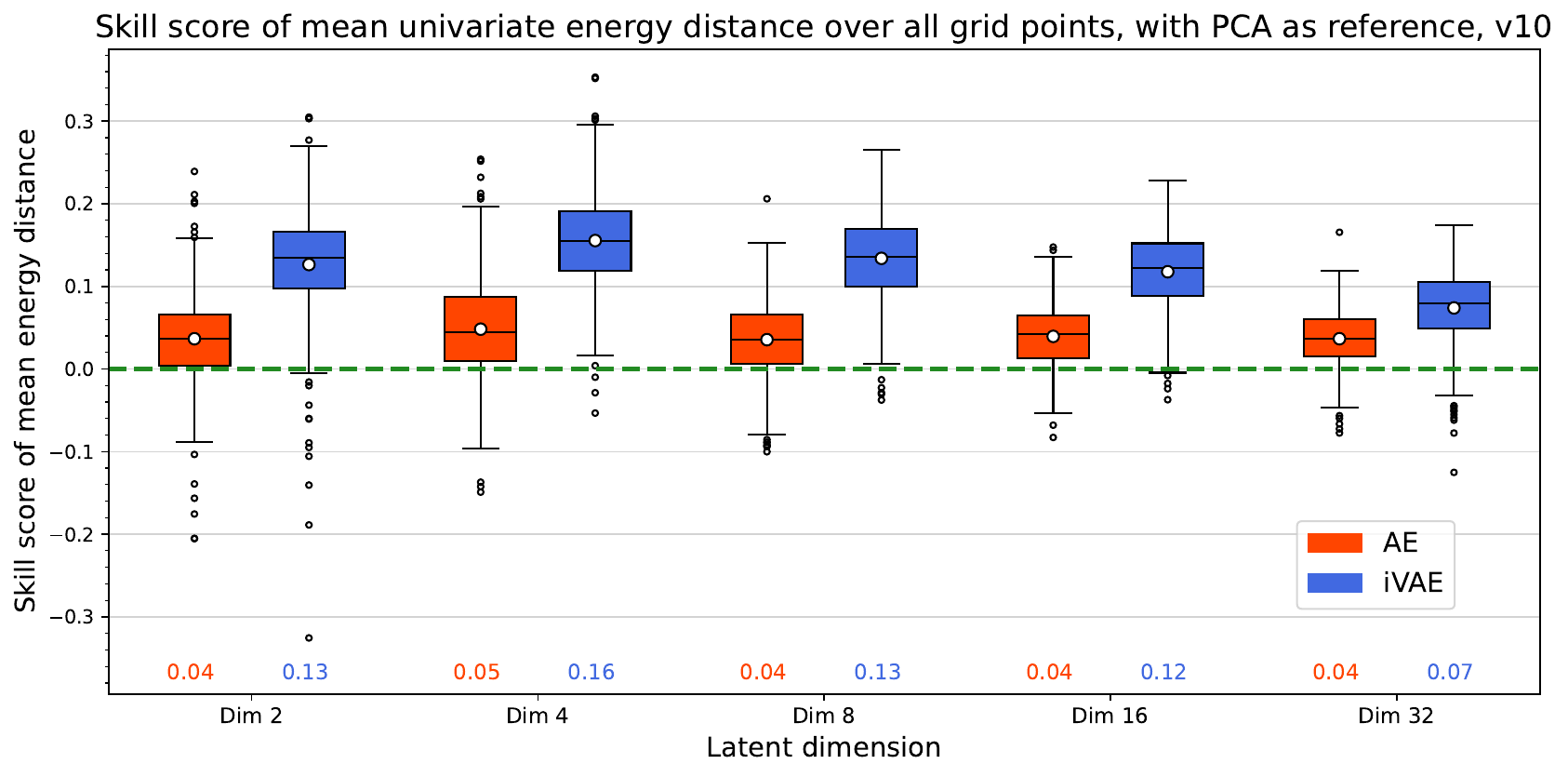}
	\includegraphics[width=\textwidth]{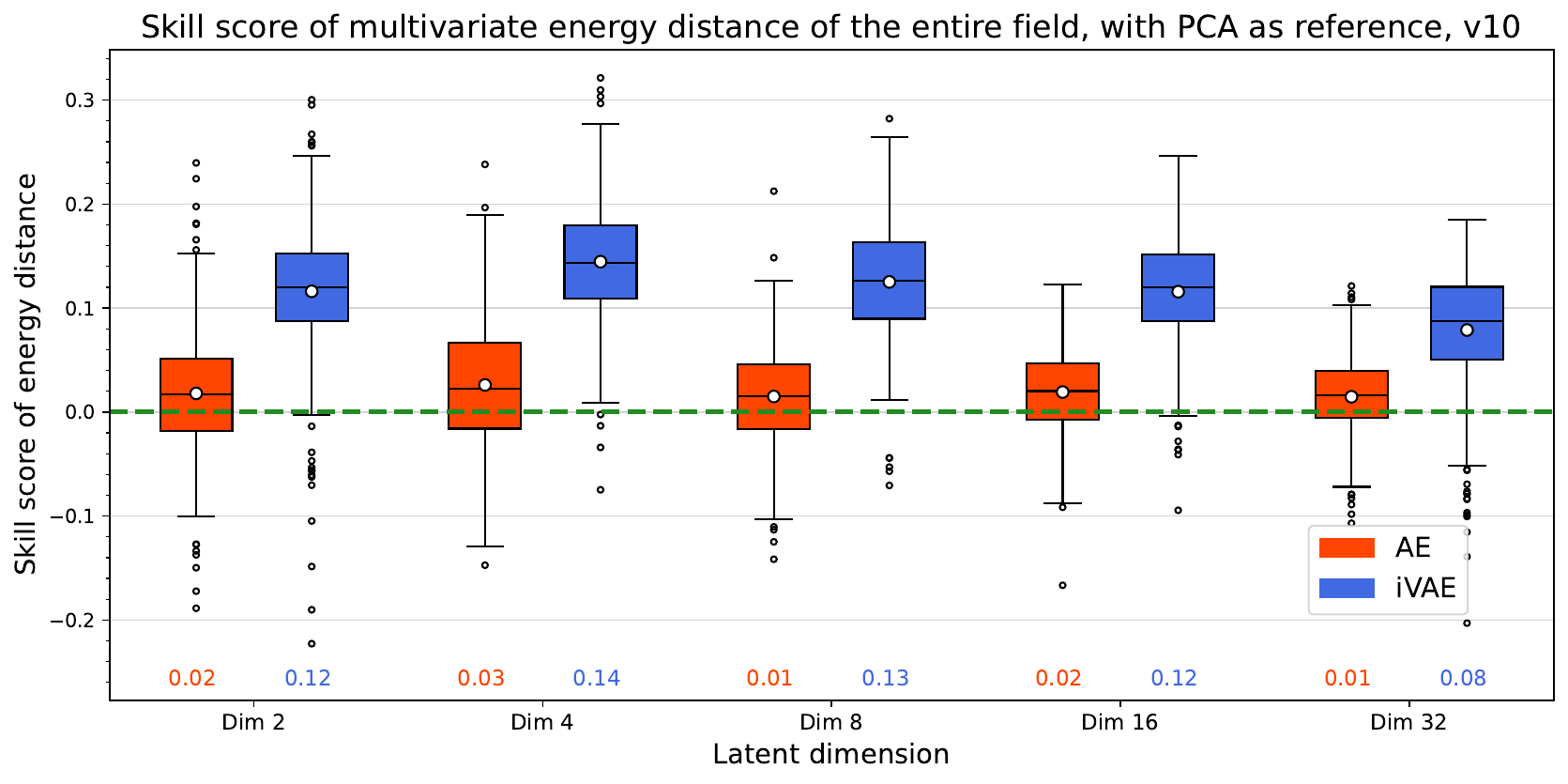}
	\caption{Boxplots of skill scores based on energy distances between the input and reconstructed ensemble fields over the 366 days in the test set for the V component of wind speed. The panels show mean univariate energy distances over all grid points (top) and multivariate energy distances computed for the entire fields (bottom). The PCA-based approach shown in green dashed line is the reference method. The respective mean skill values are indicated below each box.}
	\label{fig_ed_box_v}
\end{figure}

\begin{figure}
	\centering
	\includegraphics[width=\textwidth]{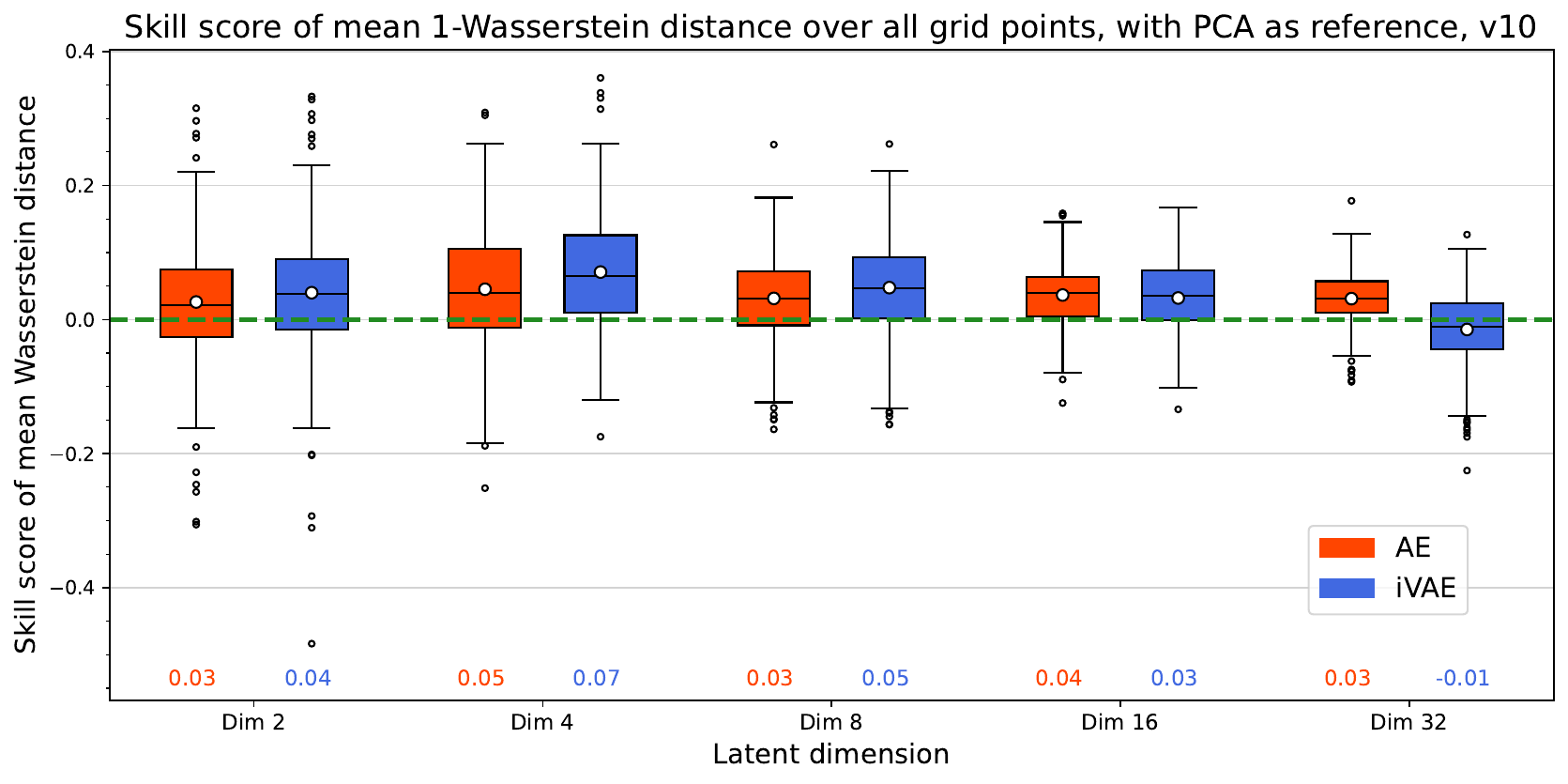}
	\includegraphics[width=\textwidth]{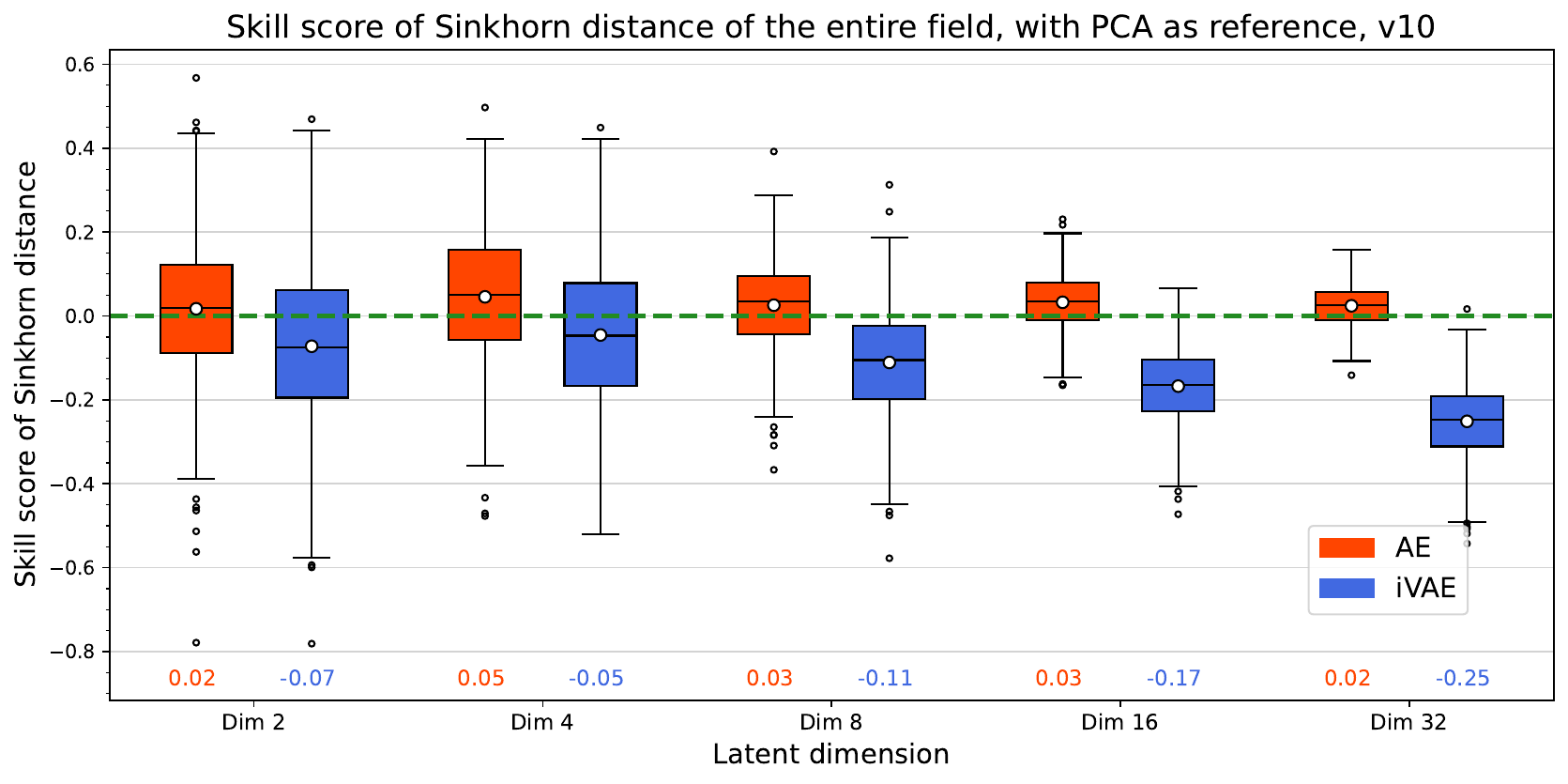}
	\caption{As Figure \ref{fig_wd_box_t}, but for the V component of 10-m wind speed.}
	\label{fig_wd_box_v}
\end{figure}

\begin{figure}
	\centering
	\includegraphics[width=\textwidth]{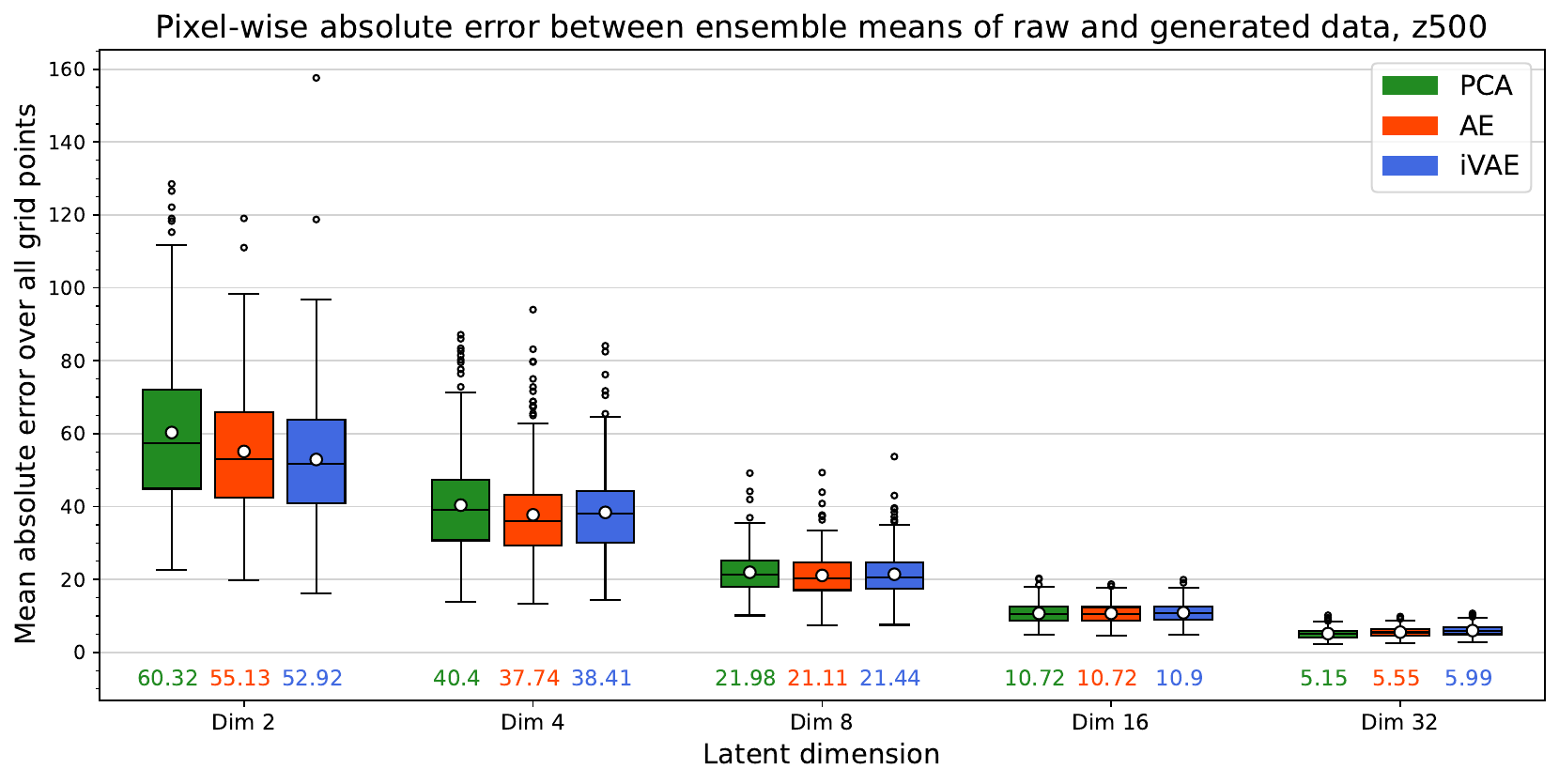}
	\includegraphics[width=\textwidth]{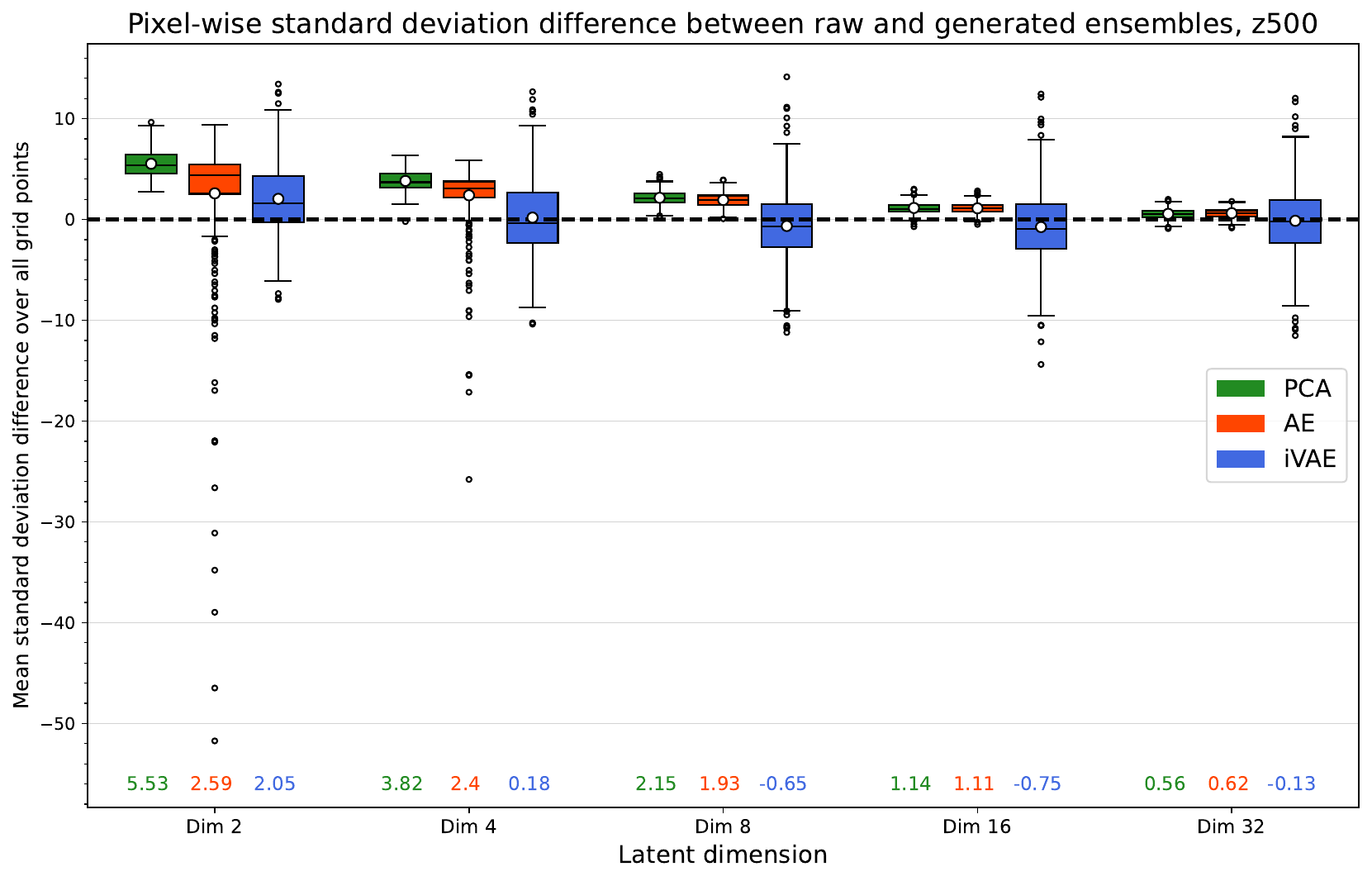}
	\caption{As Figure \ref{fig_mae_std_box_v}, but for geopotential height at 500 hPa.}
	\label{fig_mae_std_box_z}
\end{figure}

\begin{figure}
	\centering
	\includegraphics[width=\textwidth]{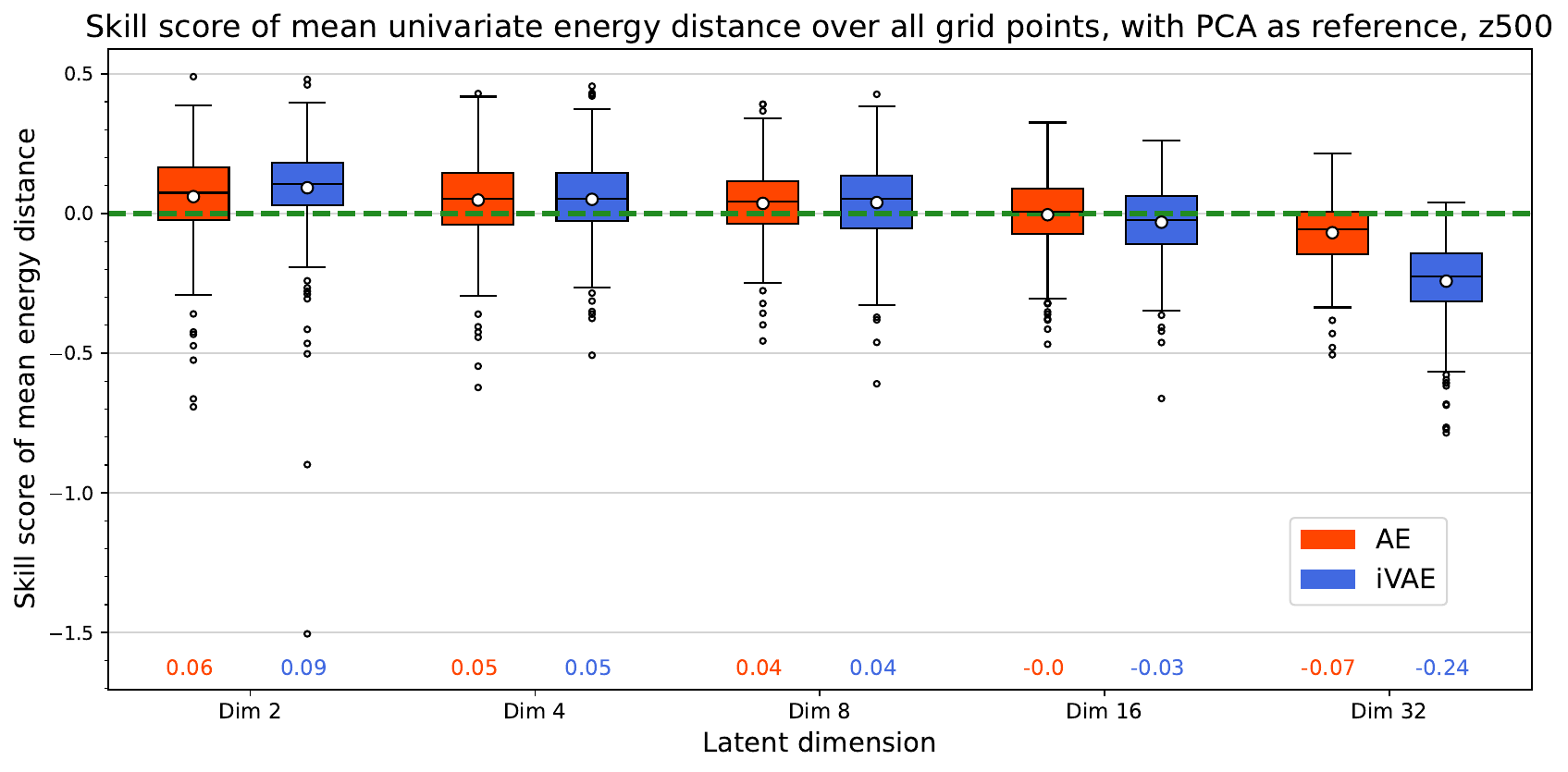}
	\includegraphics[width=\textwidth]{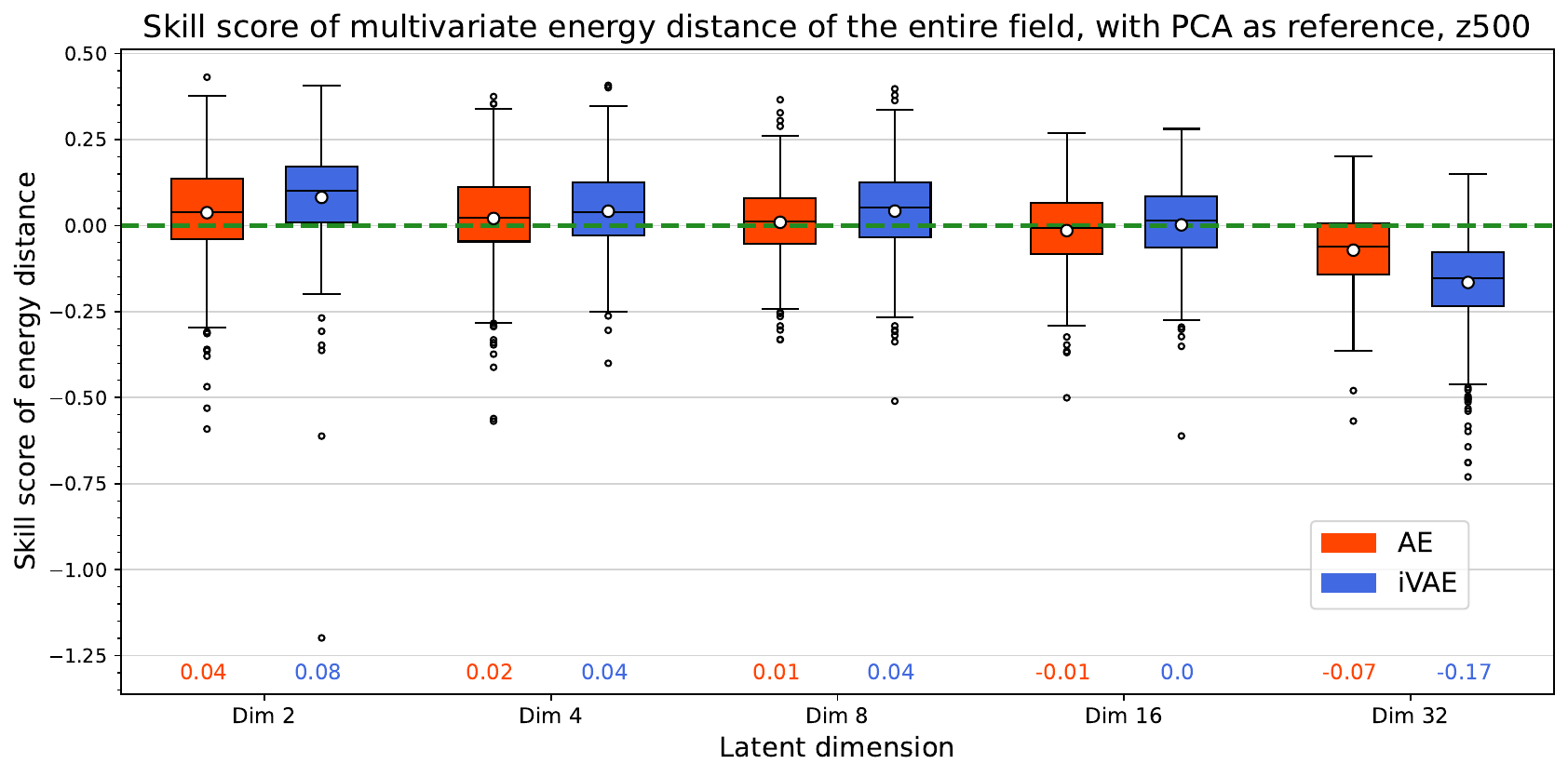}
	\caption{As Figure \ref{fig_ed_box_v}, but for geopotential height at 500 hPa.}
	\label{fig_ed_box_z}
\end{figure}

\begin{figure}
	\centering
	\includegraphics[width=\textwidth]{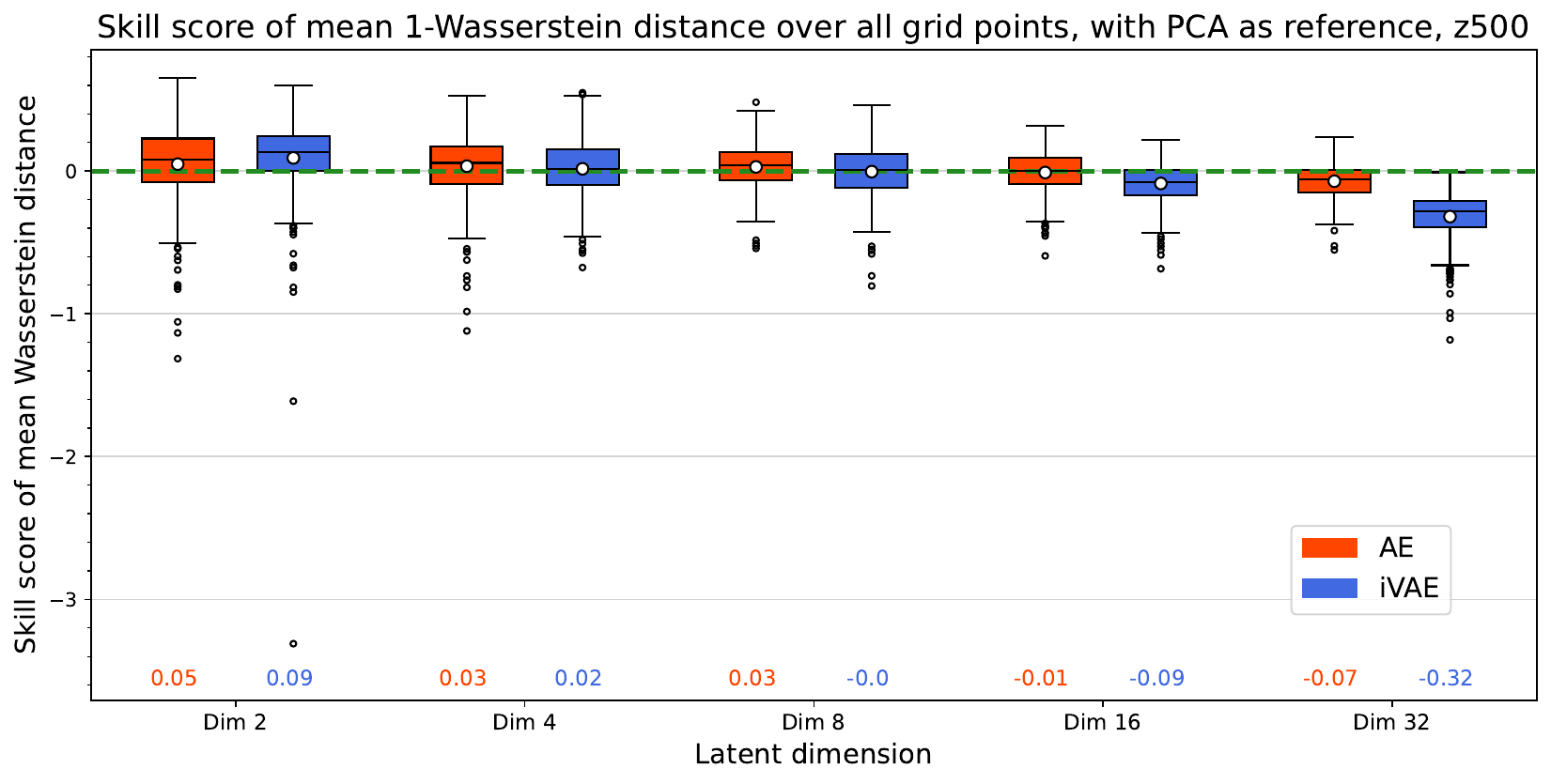}
	\includegraphics[width=\textwidth]{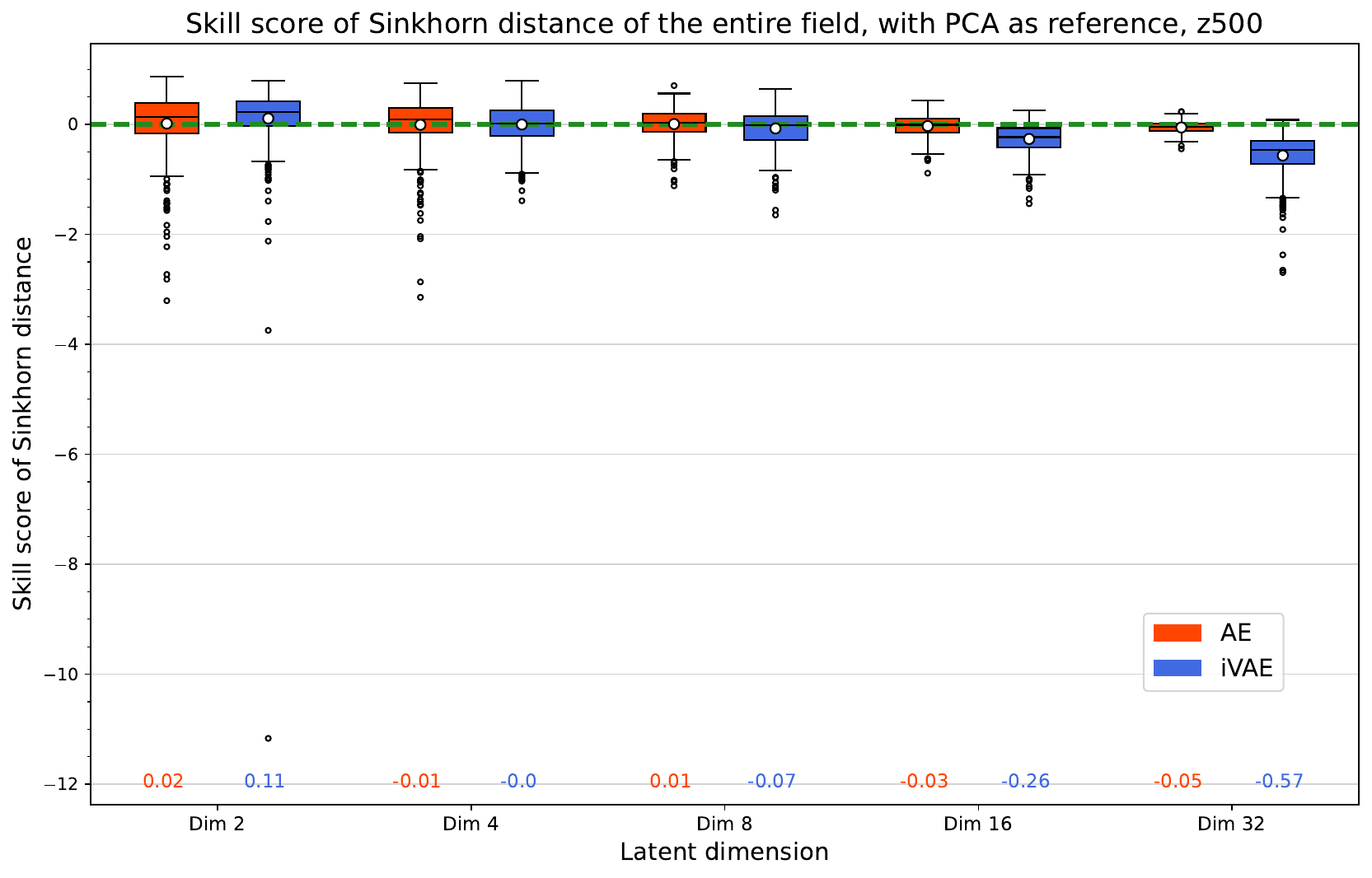}
	\caption{As Figure \ref{fig_wd_box_t}, but for geopotential height at 500 hPa.}
	\label{fig_wd_box_z}
\end{figure}

\begin{figure}
	\centering
	\includegraphics[width=\textwidth]{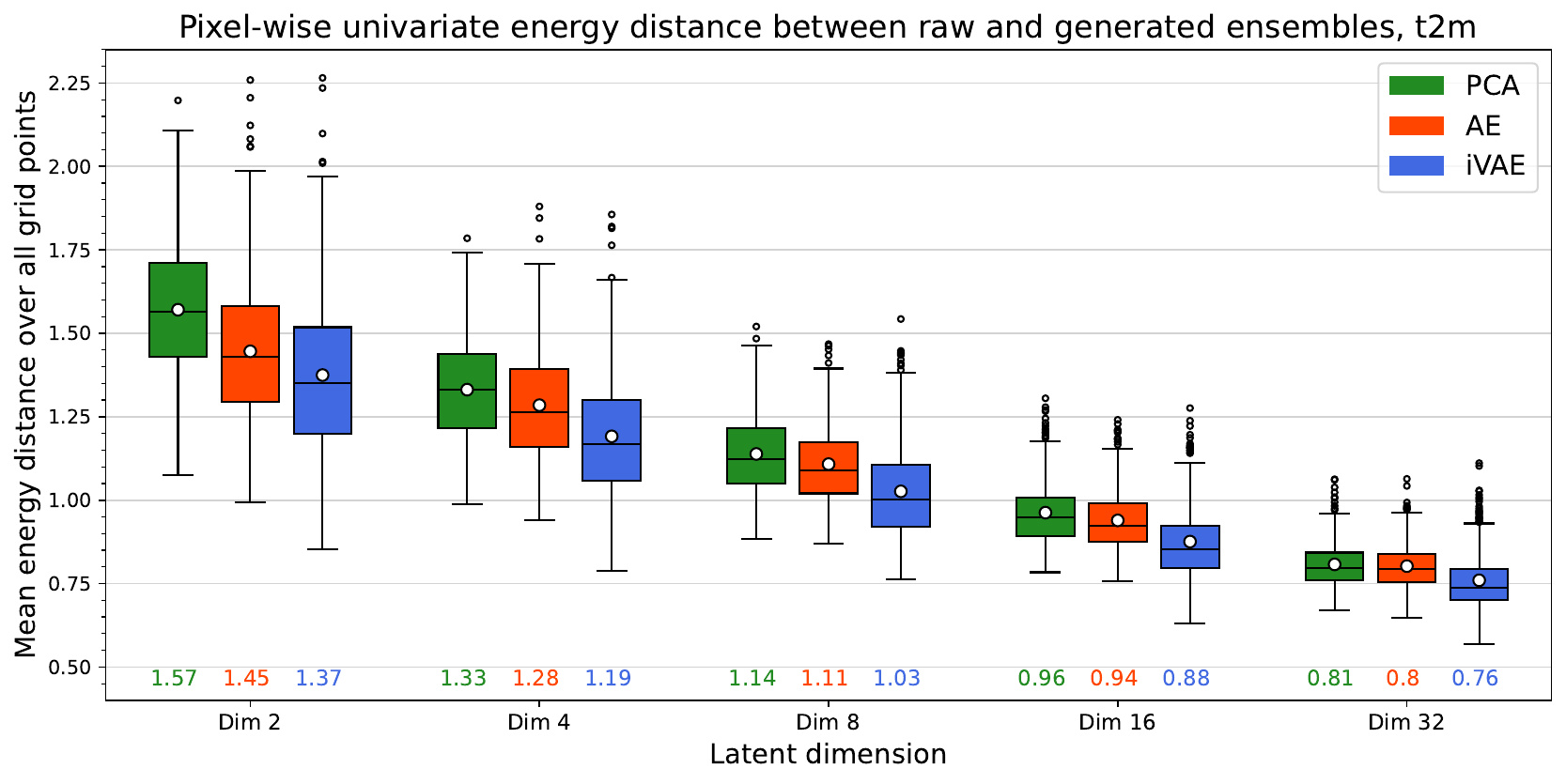}
	\includegraphics[width=\textwidth]{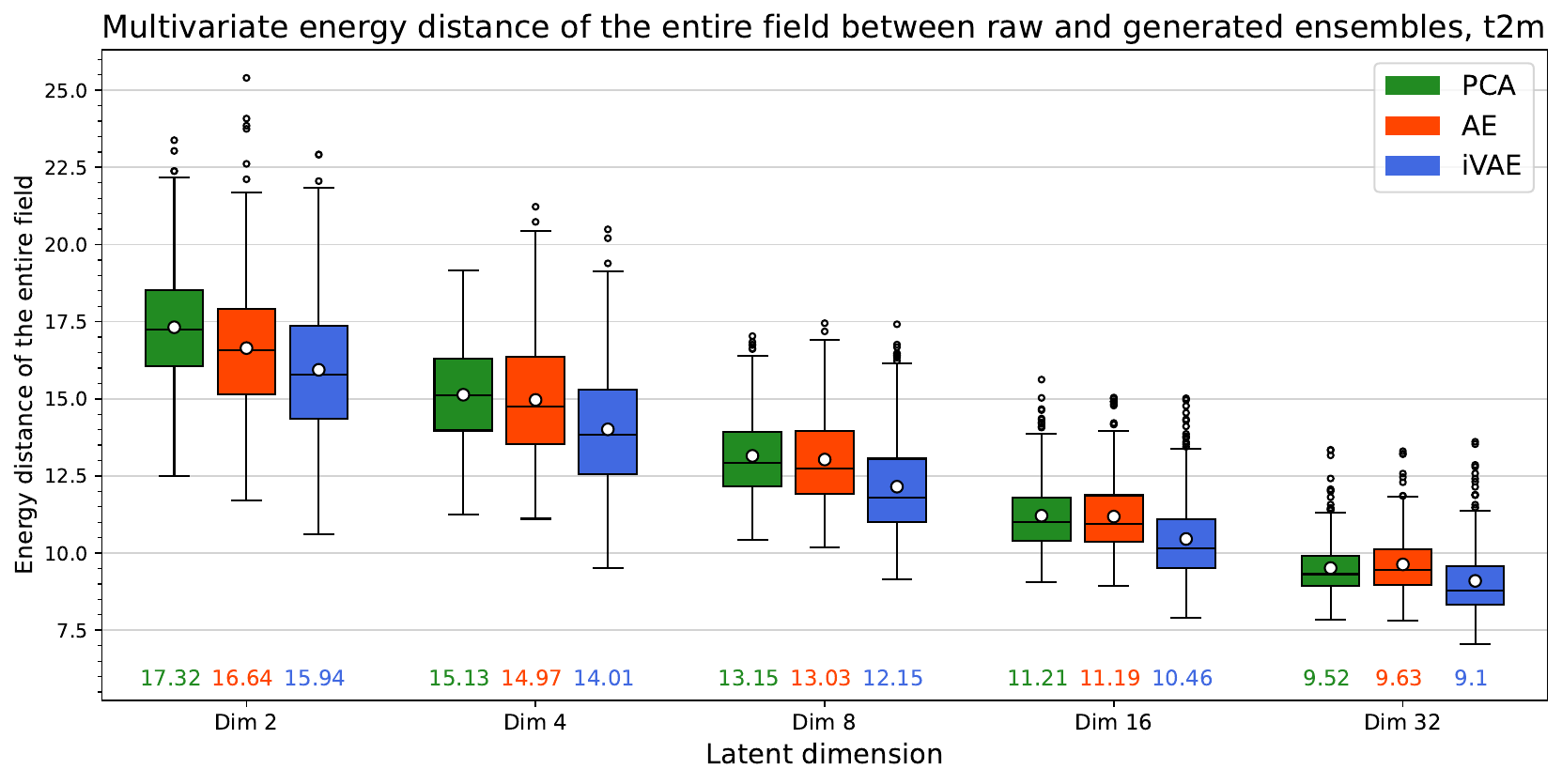}
	\caption{Boxplots of energy distances between the input and reconstructed ensemble fields over the 366 days in the test set for temperature data. The panels show mean univariate energy distances over all grid points (top) and multivariate energy distances computed for the entire fields (bottom). The respective mean values are indicated below each box.}
	\label{fig_ed_t}
\end{figure}

\begin{figure}
	\centering
	\includegraphics[width=\textwidth]{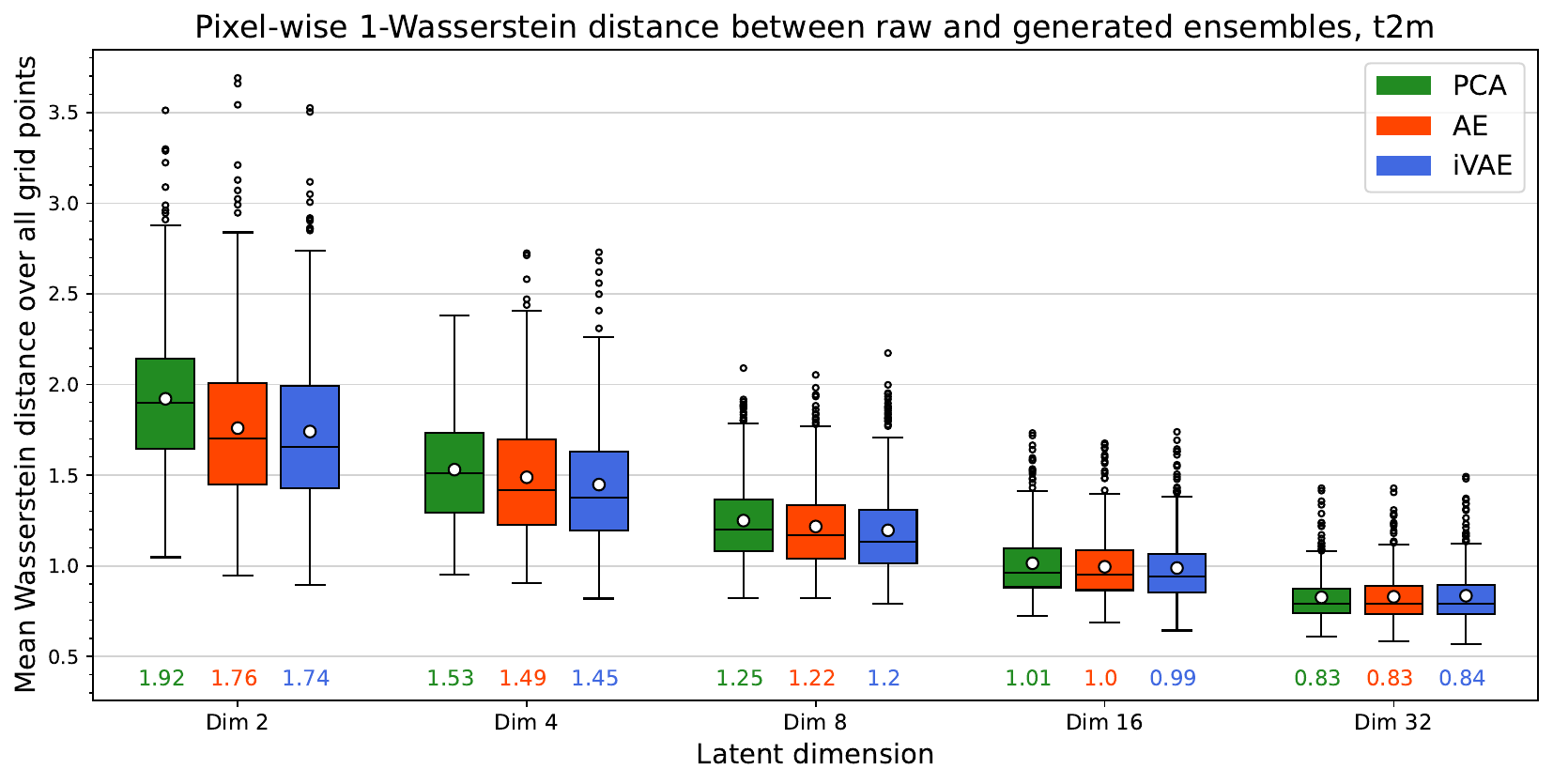}
	\includegraphics[width=\textwidth]{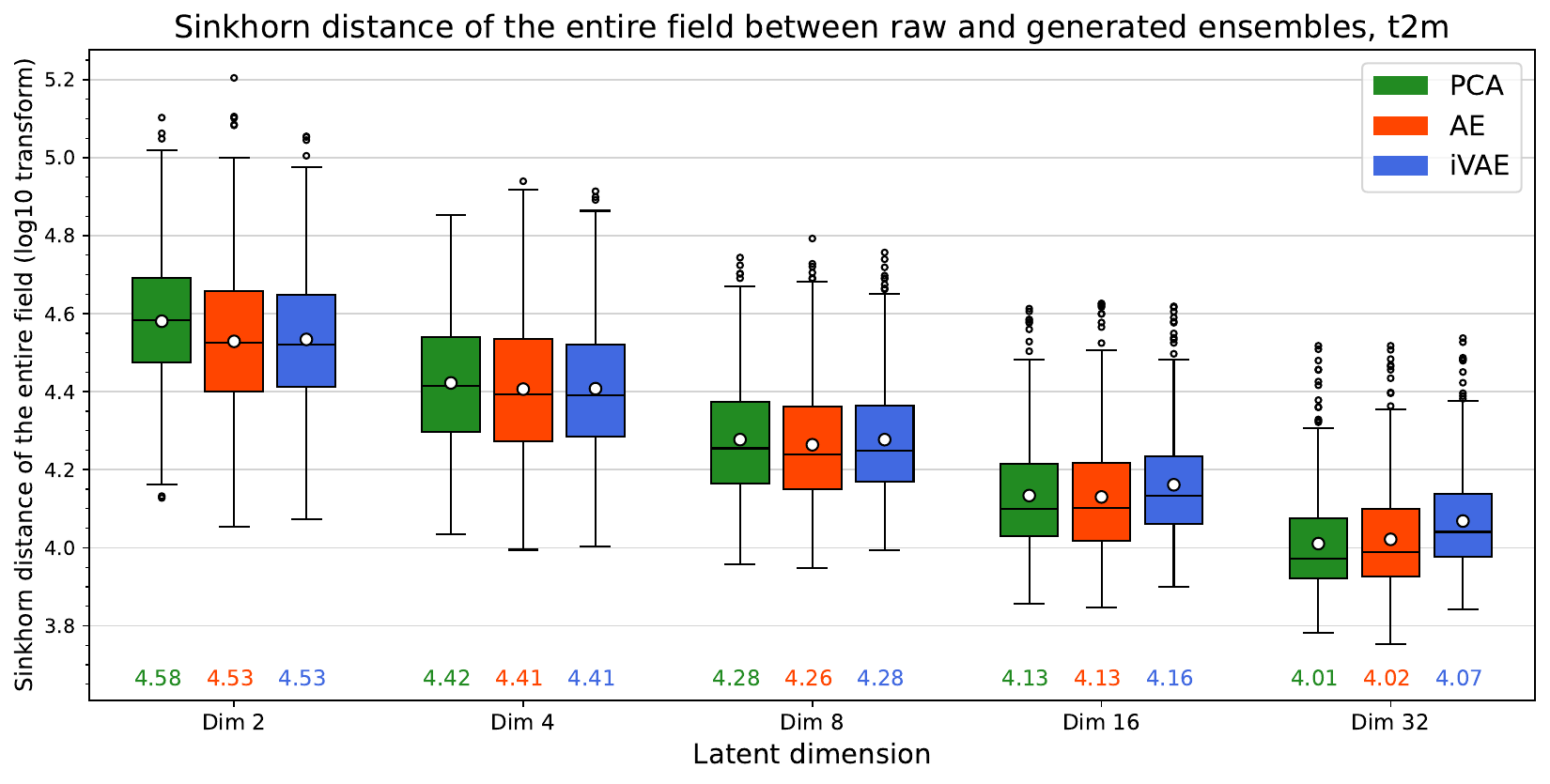}
	\caption{Boxplots of optimal transportation distances between the input and reconstructed ensemble fields over the 366 days in the test set for temperature data. The panels show mean univariate 1-Wasserstein distances over all grid points (top) and multivariate Sinkhorn distances computed for the entire fields (bottom). The respective mean values are indicated below each box.}
	\label{fig_wd_t}
\end{figure}

\begin{figure}
	\centering
	\includegraphics[width=\textwidth]{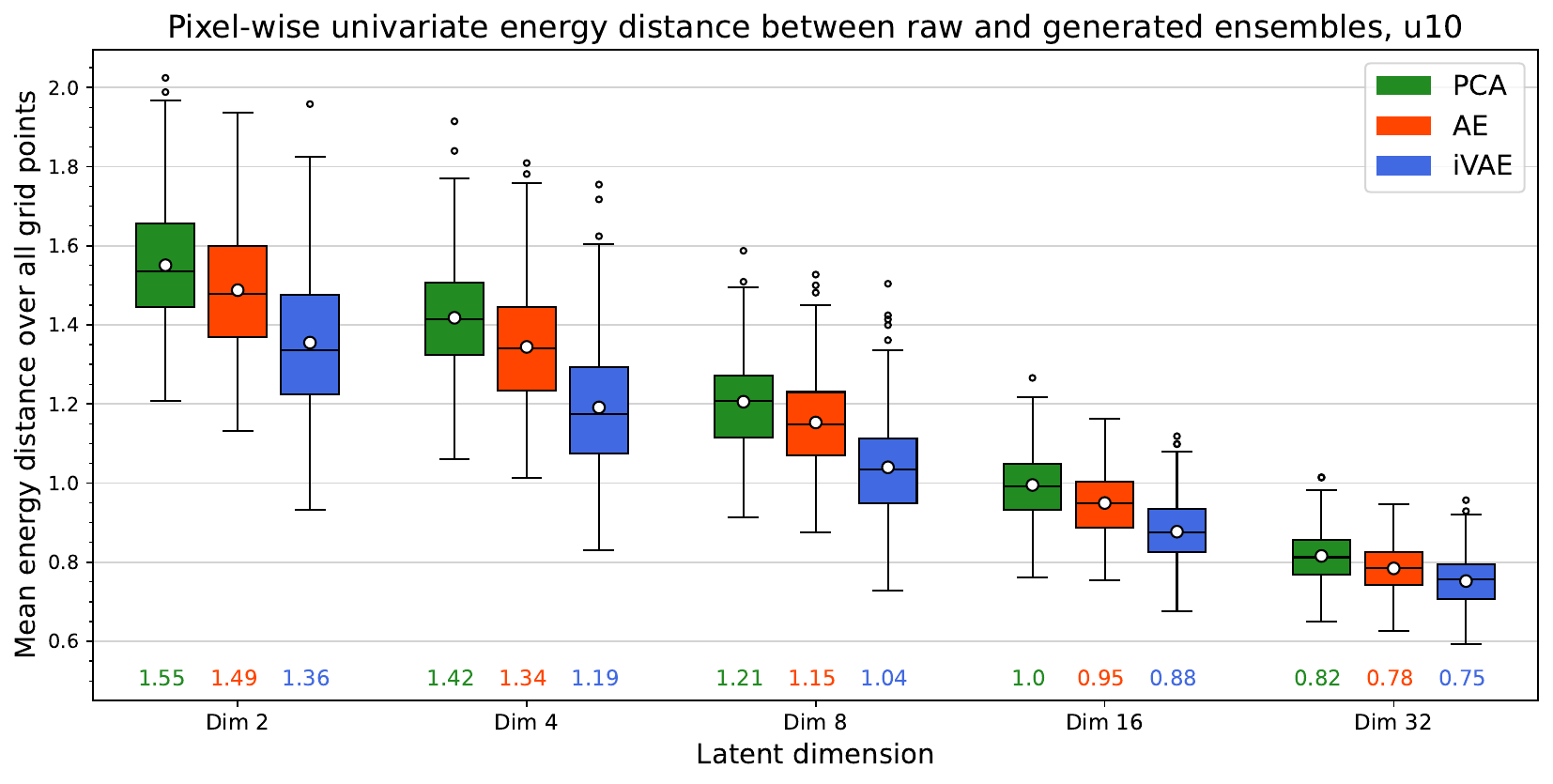}
	\includegraphics[width=\textwidth]{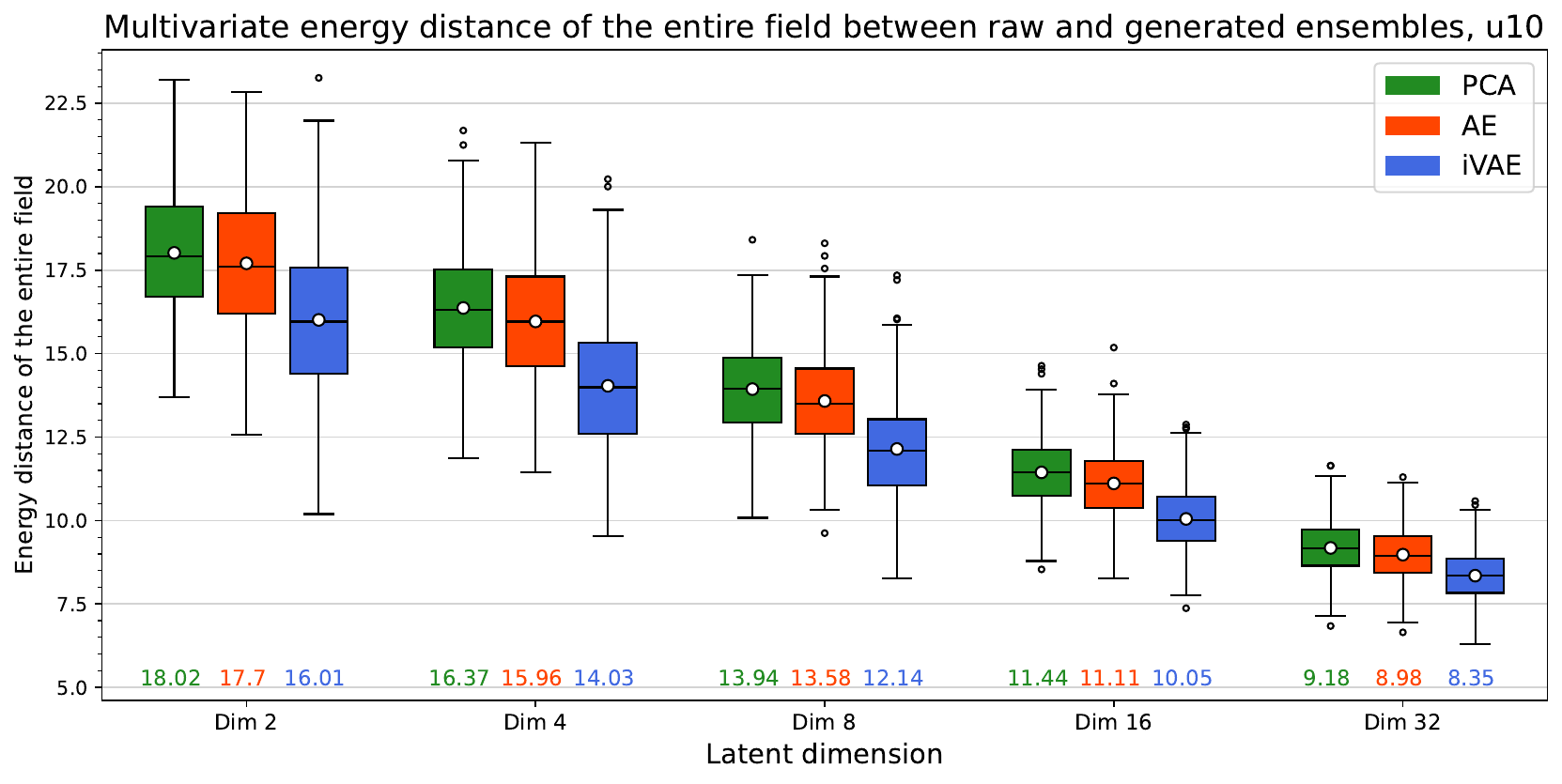}
	\caption{As Figure \ref{fig_ed_t}, but for the U component of wind speed.}
	\label{fig_ed_u}
\end{figure}

\begin{figure}
	\centering
	\includegraphics[width=\textwidth]{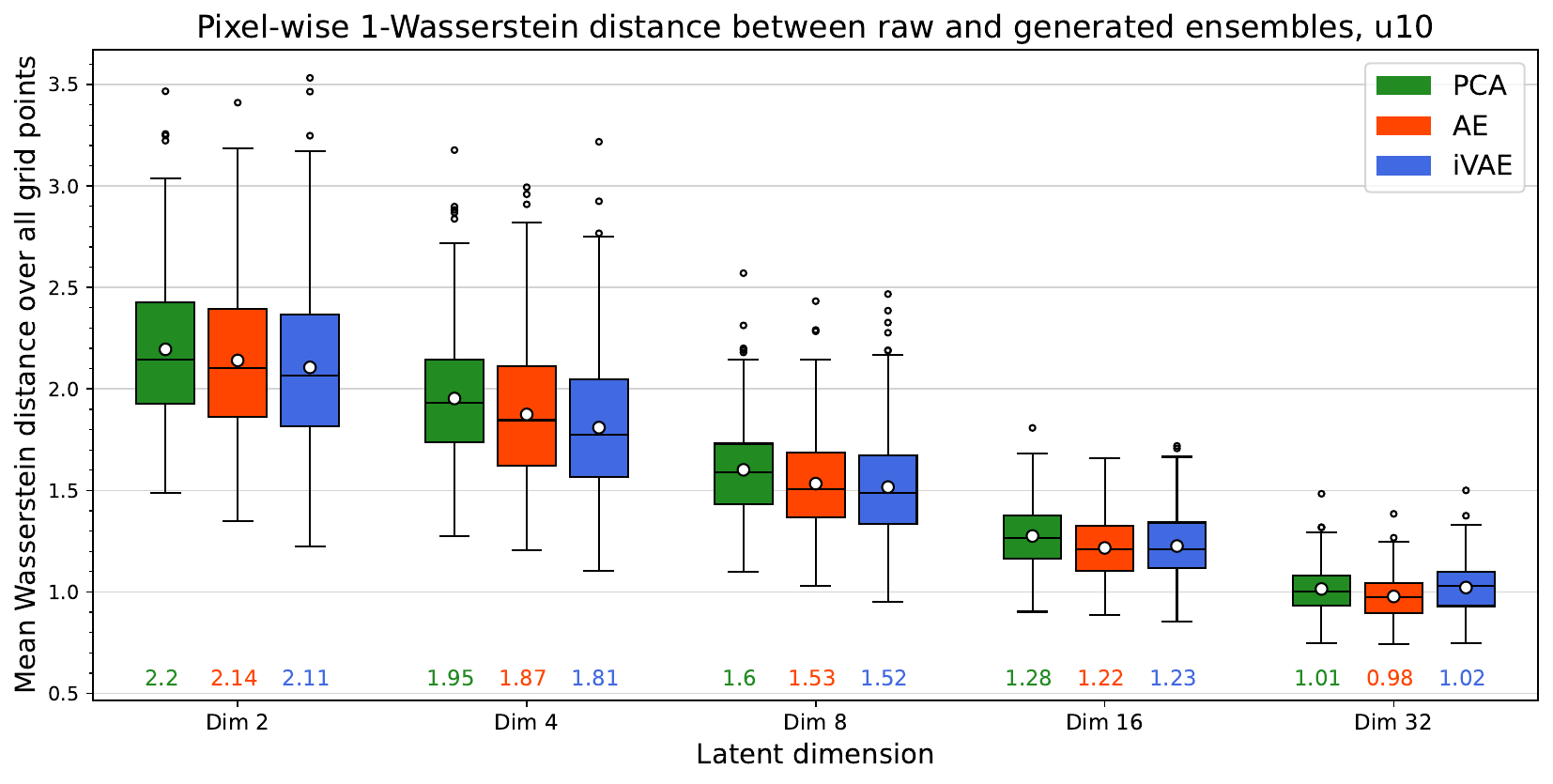}
	\includegraphics[width=\textwidth]{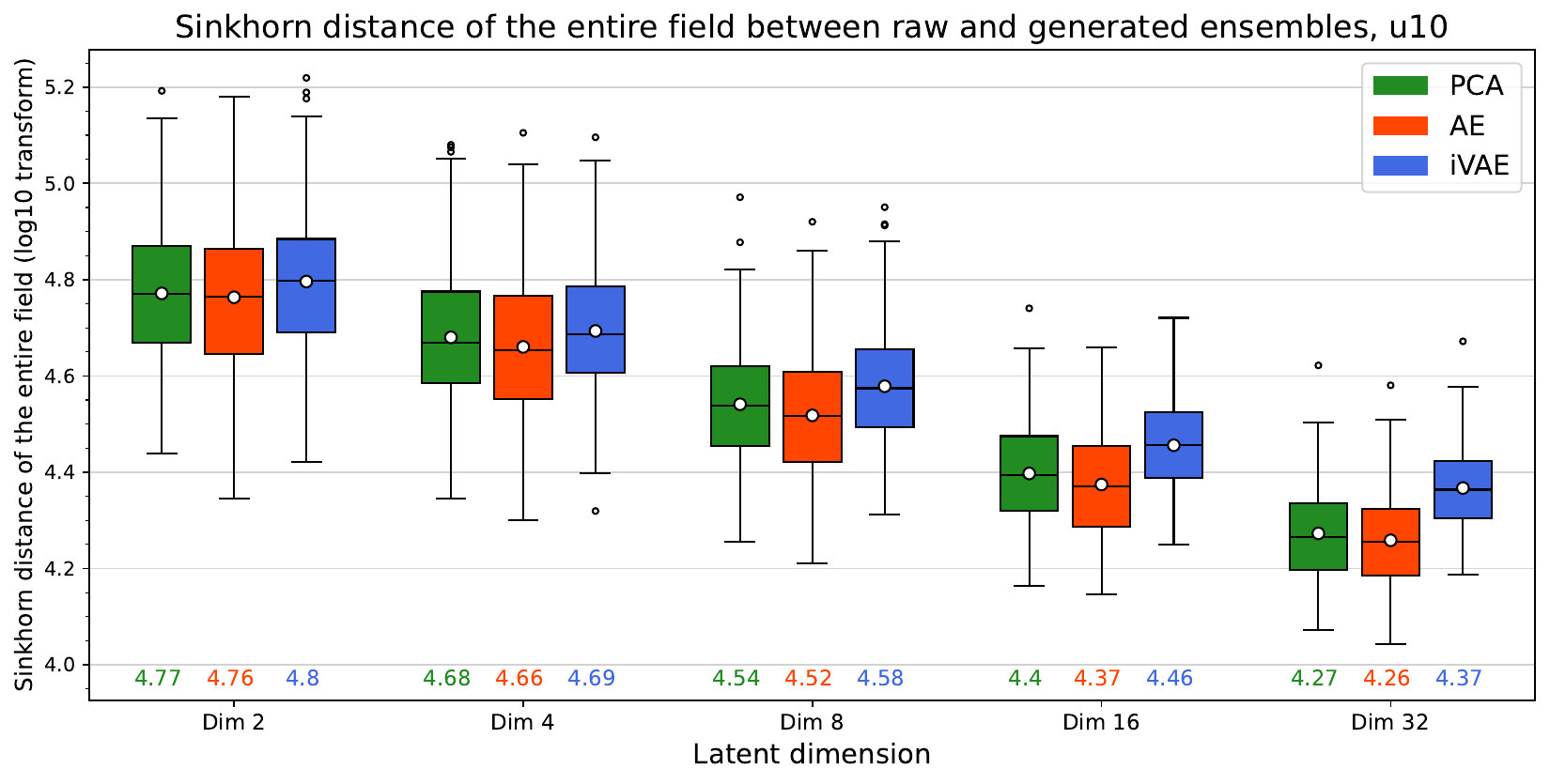}
	\caption{As Figure \ref{fig_wd_t}, but for the U component of wind speed.}
	\label{fig_wd_u}
\end{figure}

\begin{figure}
	\centering
	\includegraphics[width=\textwidth]{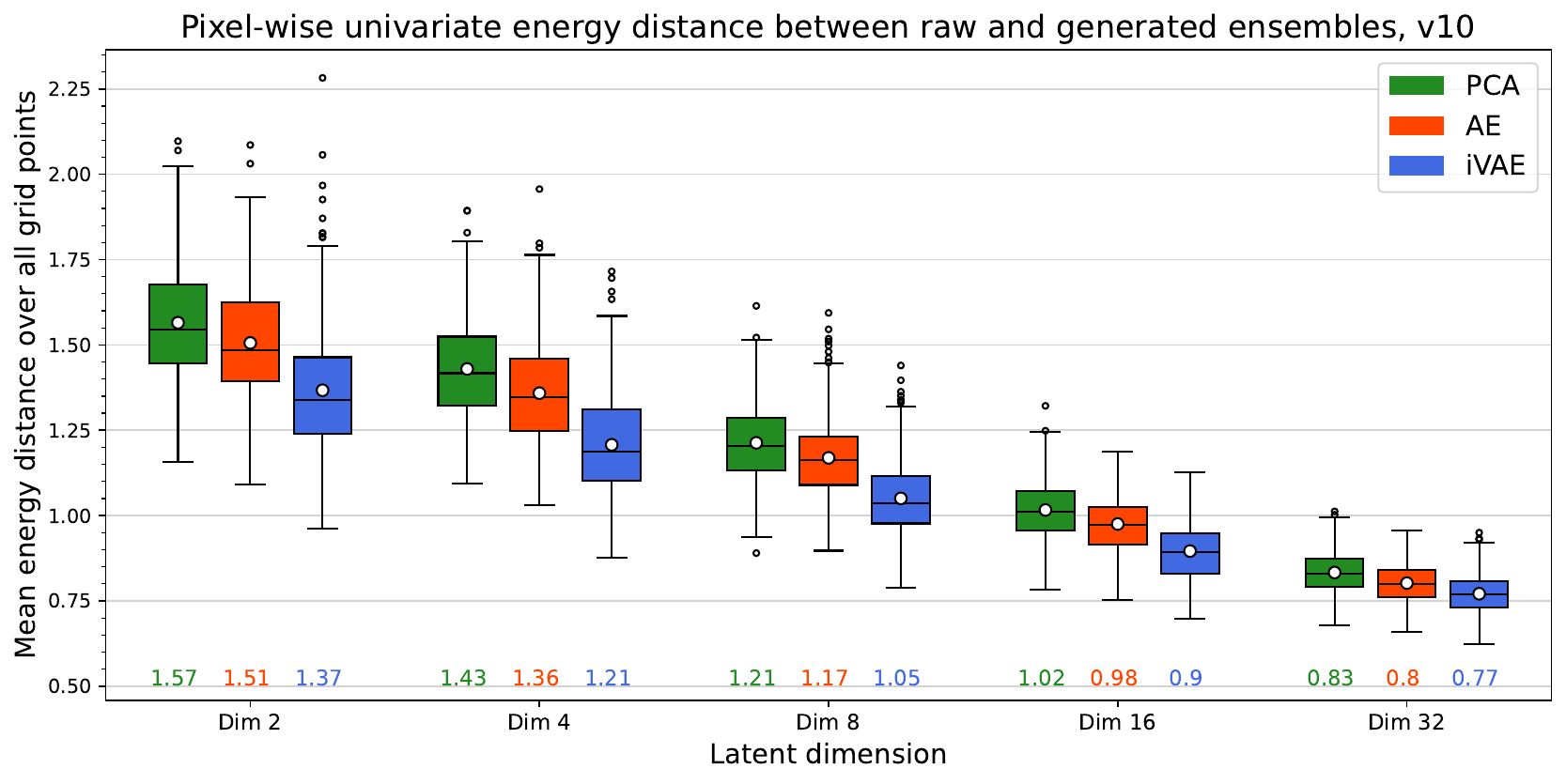}
	\includegraphics[width=\textwidth]{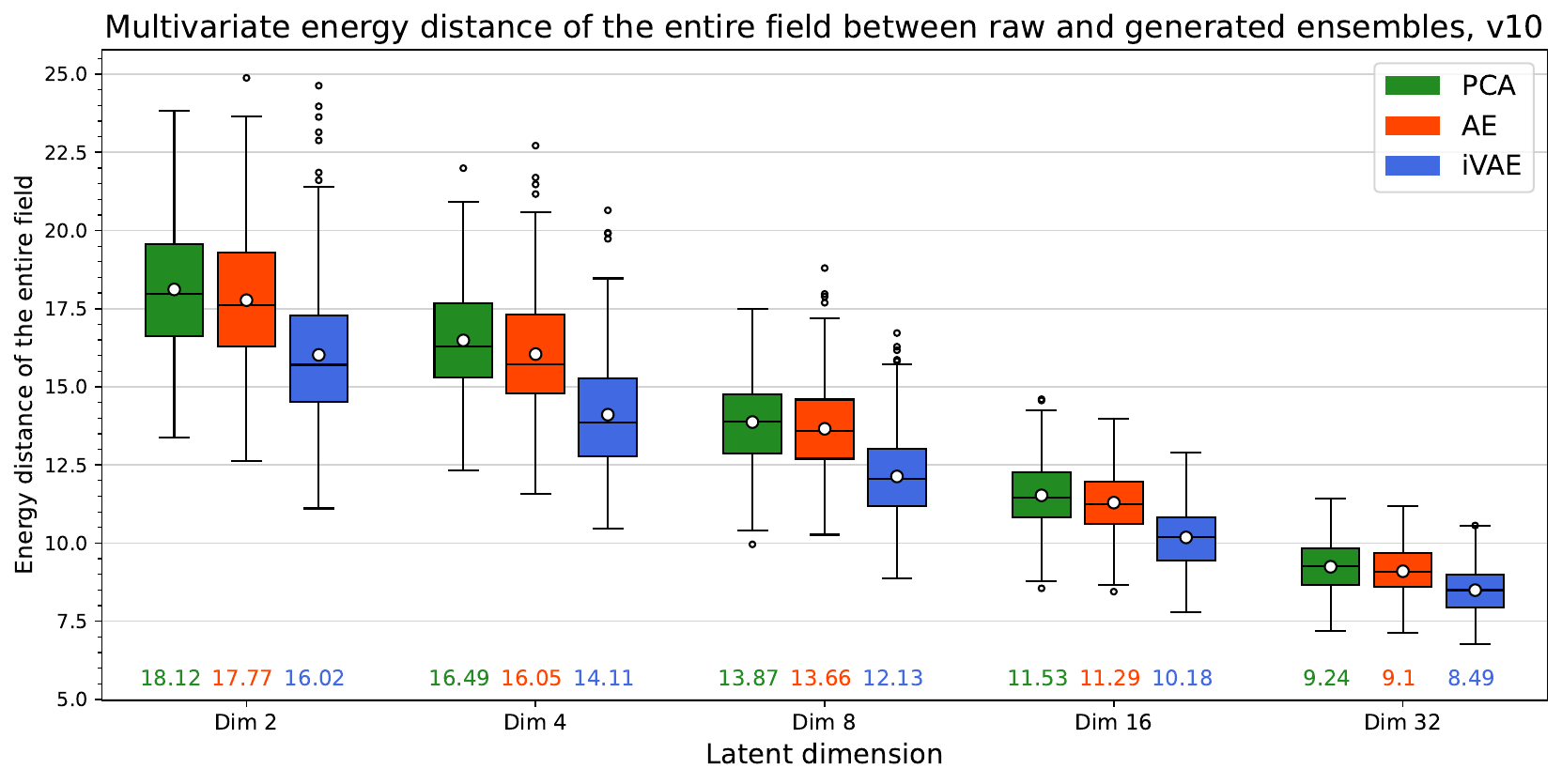}
	\caption{As Figure \ref{fig_ed_t}, but for the V component of wind speed.}
	\label{fig_ed_v}
\end{figure}

\begin{figure}
	\centering
	\includegraphics[width=\textwidth]{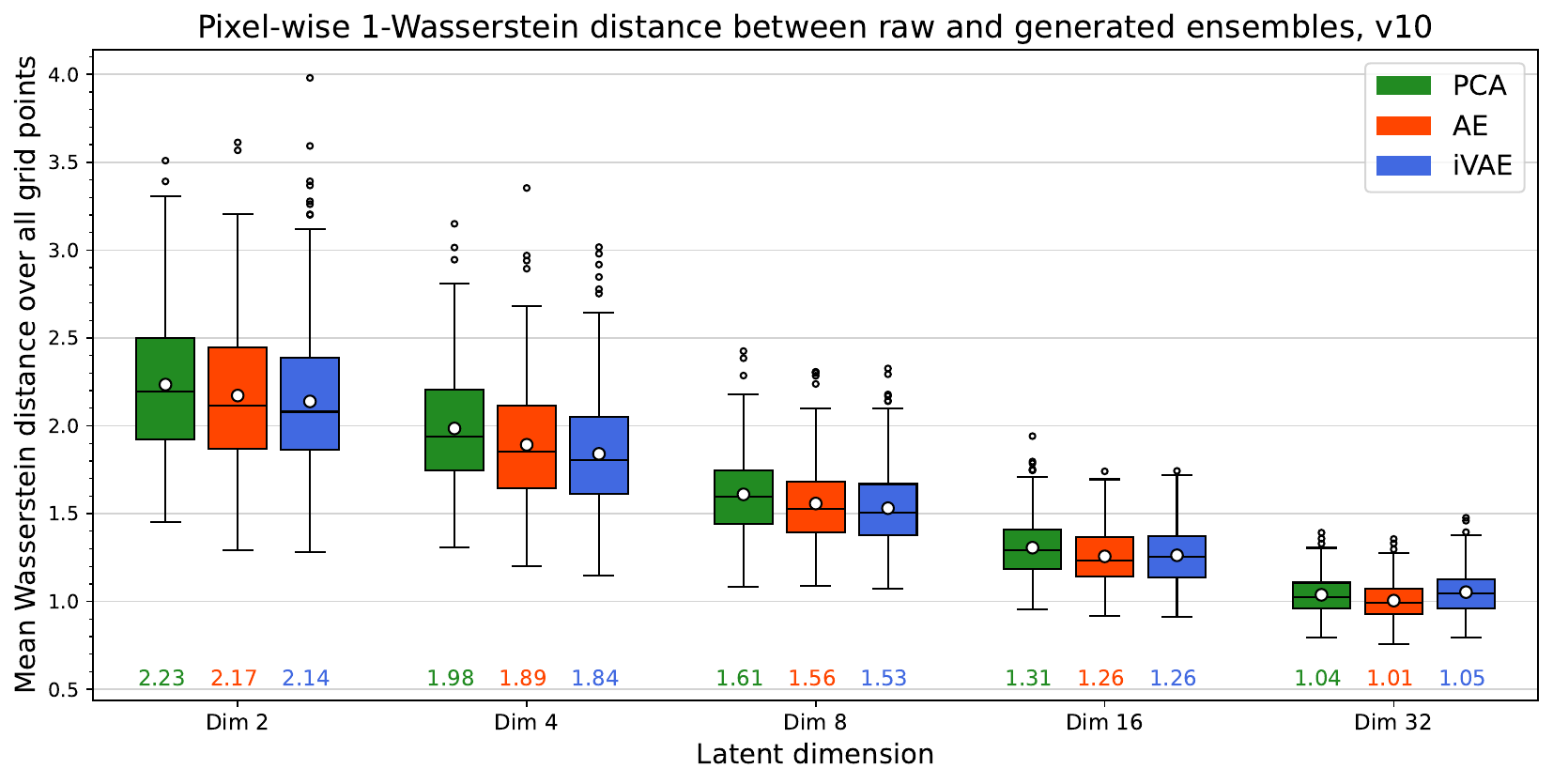}
	\includegraphics[width=\textwidth]{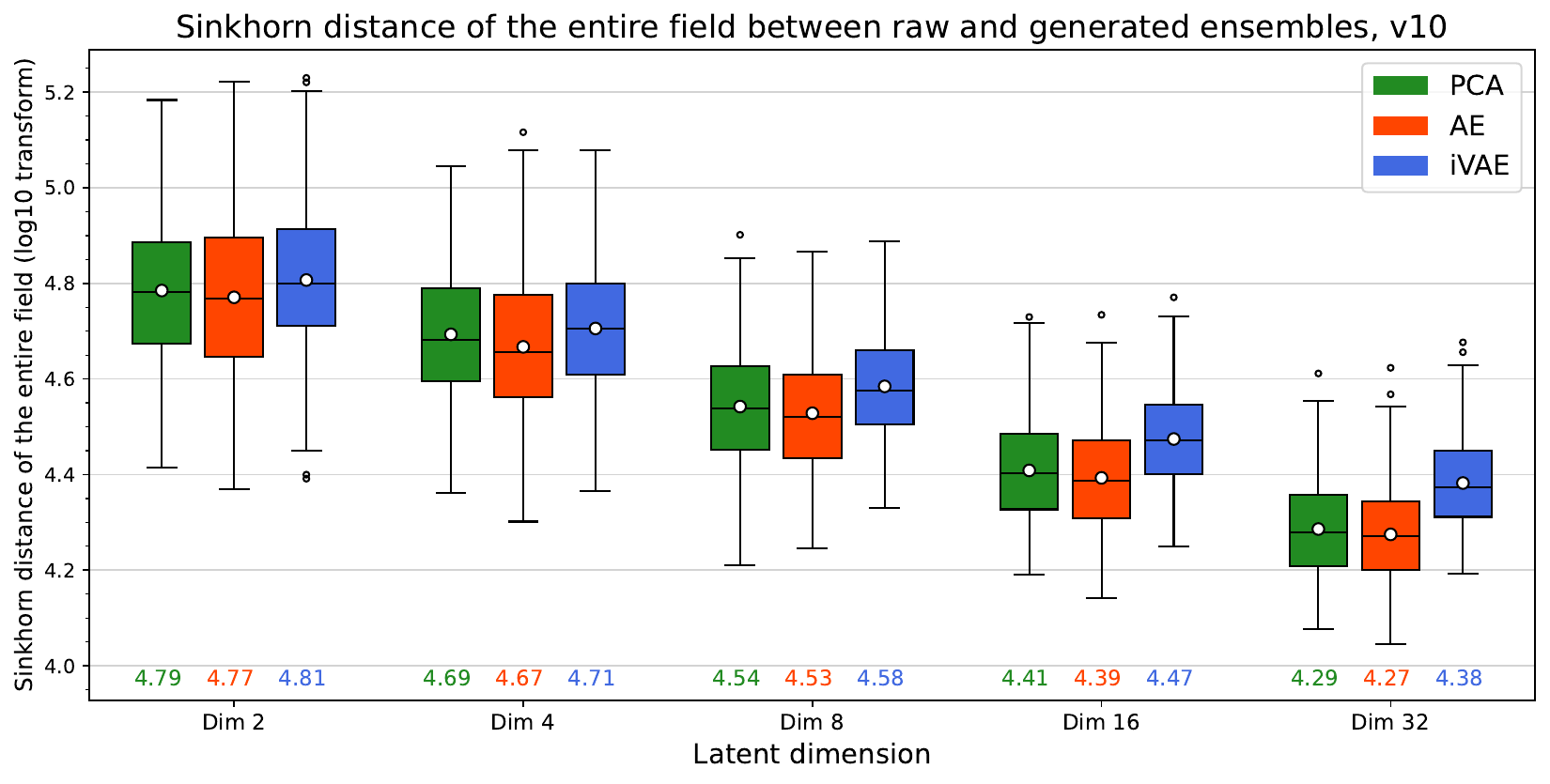}
	\caption{As Figure \ref{fig_wd_t}, but for the V component of wind speed.}
	\label{fig_wd_v}
\end{figure}

\begin{figure}
	\centering
	\includegraphics[width=\textwidth]{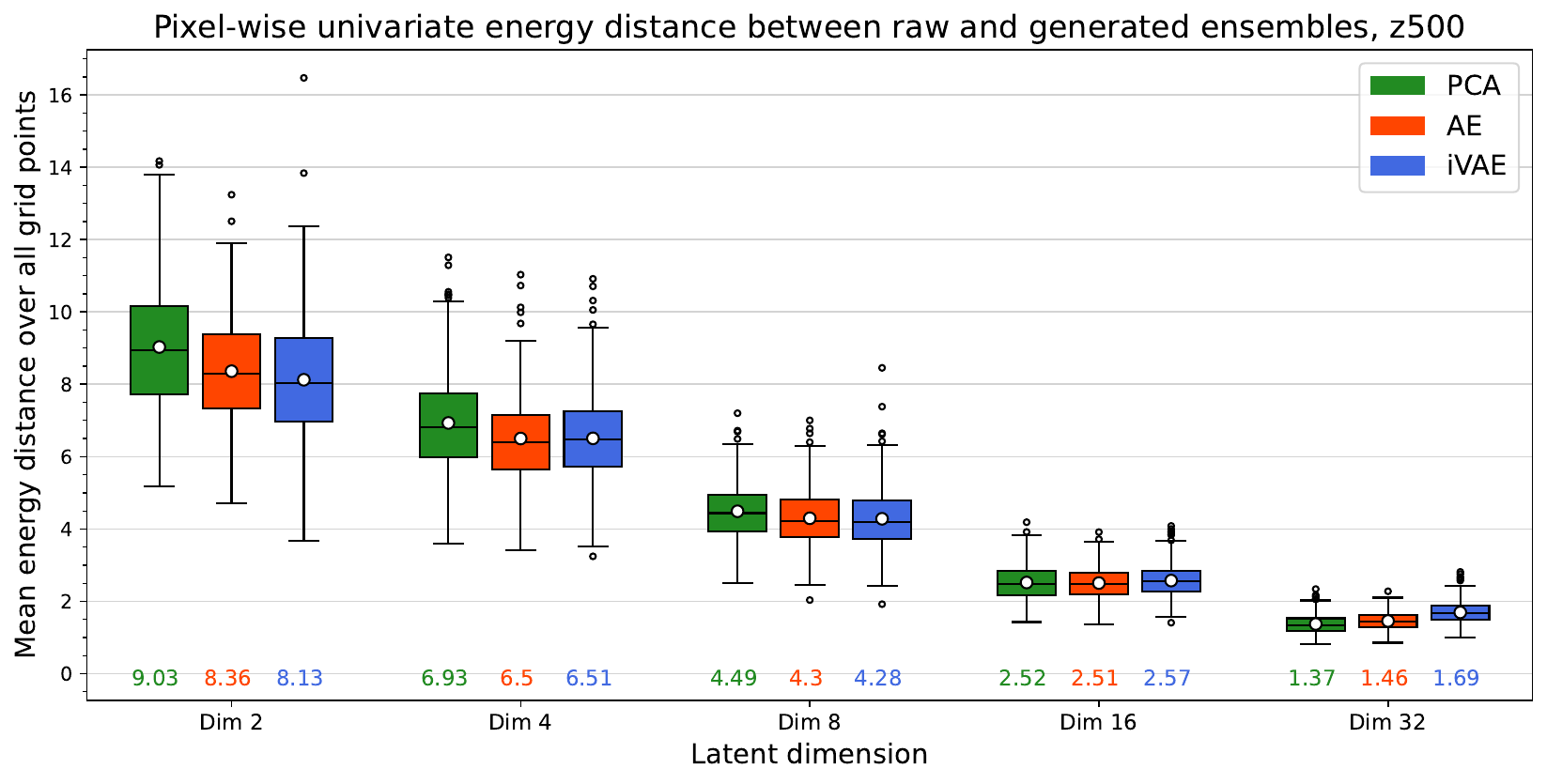}
	\includegraphics[width=\textwidth]{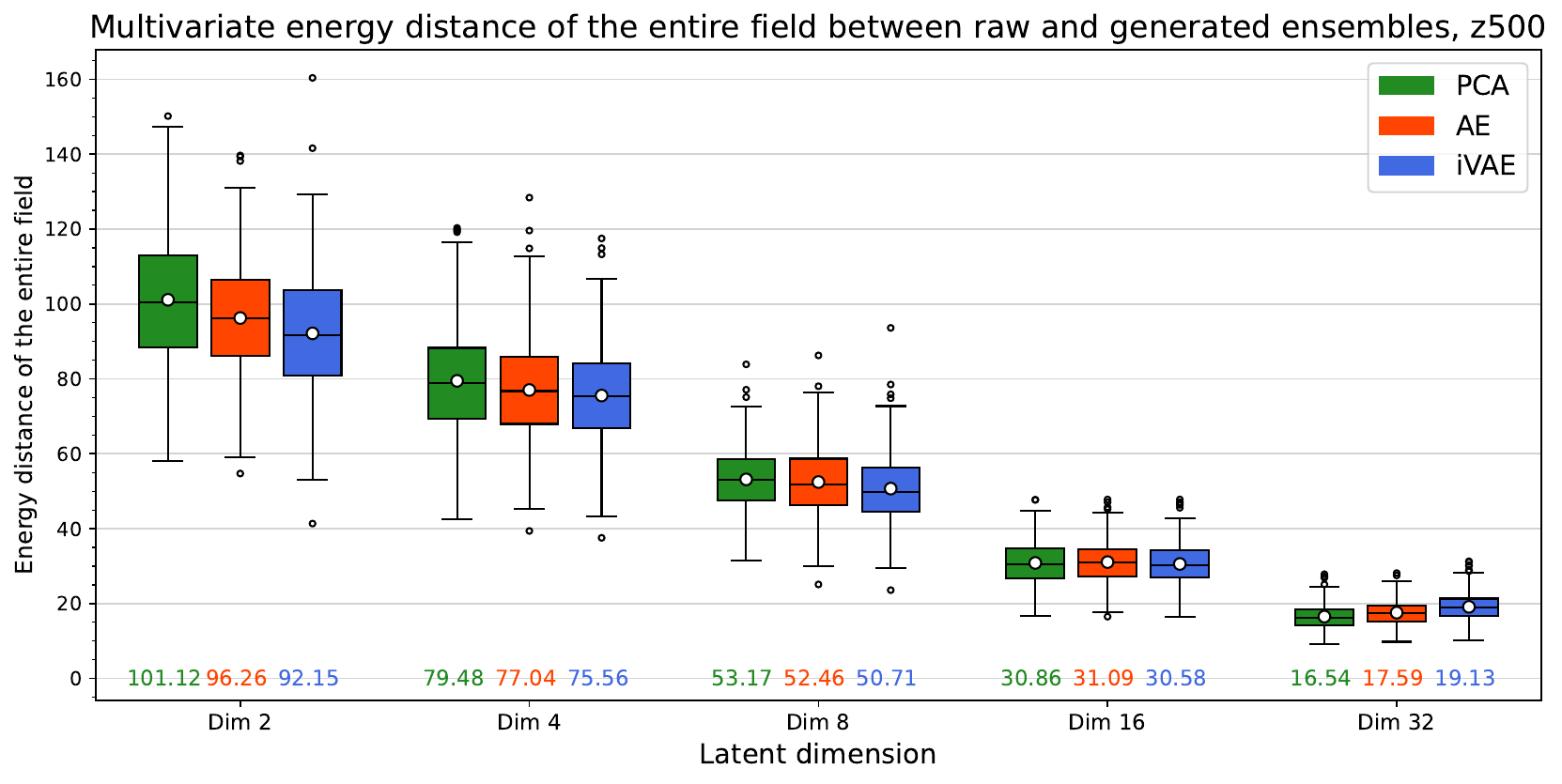}
	\caption{As Figure \ref{fig_ed_t}, but for the geopotential height.}
	\label{fig_ed_z}
\end{figure}

\begin{figure}
	\centering
	\includegraphics[width=\textwidth]{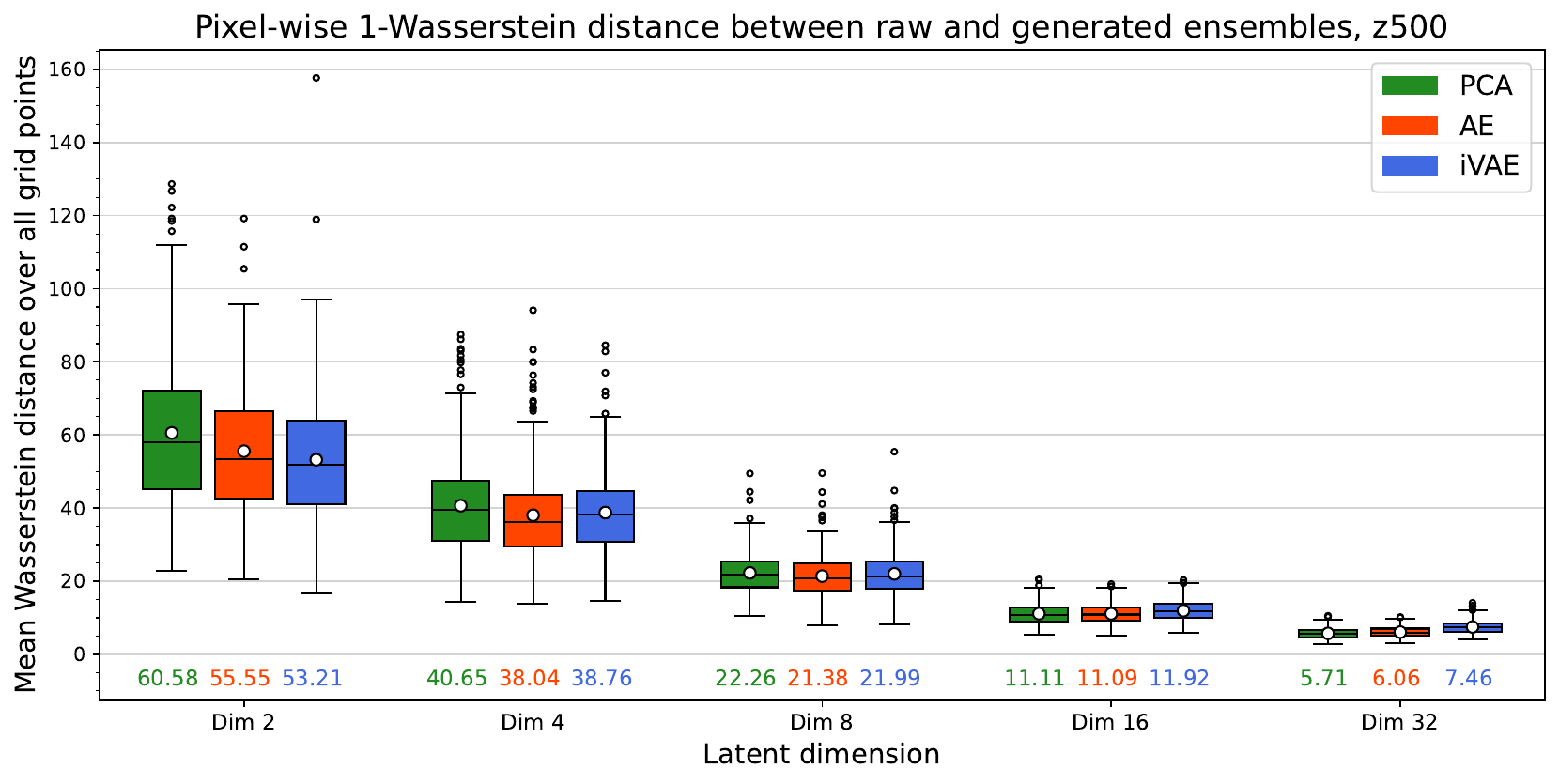}
	\includegraphics[width=\textwidth]{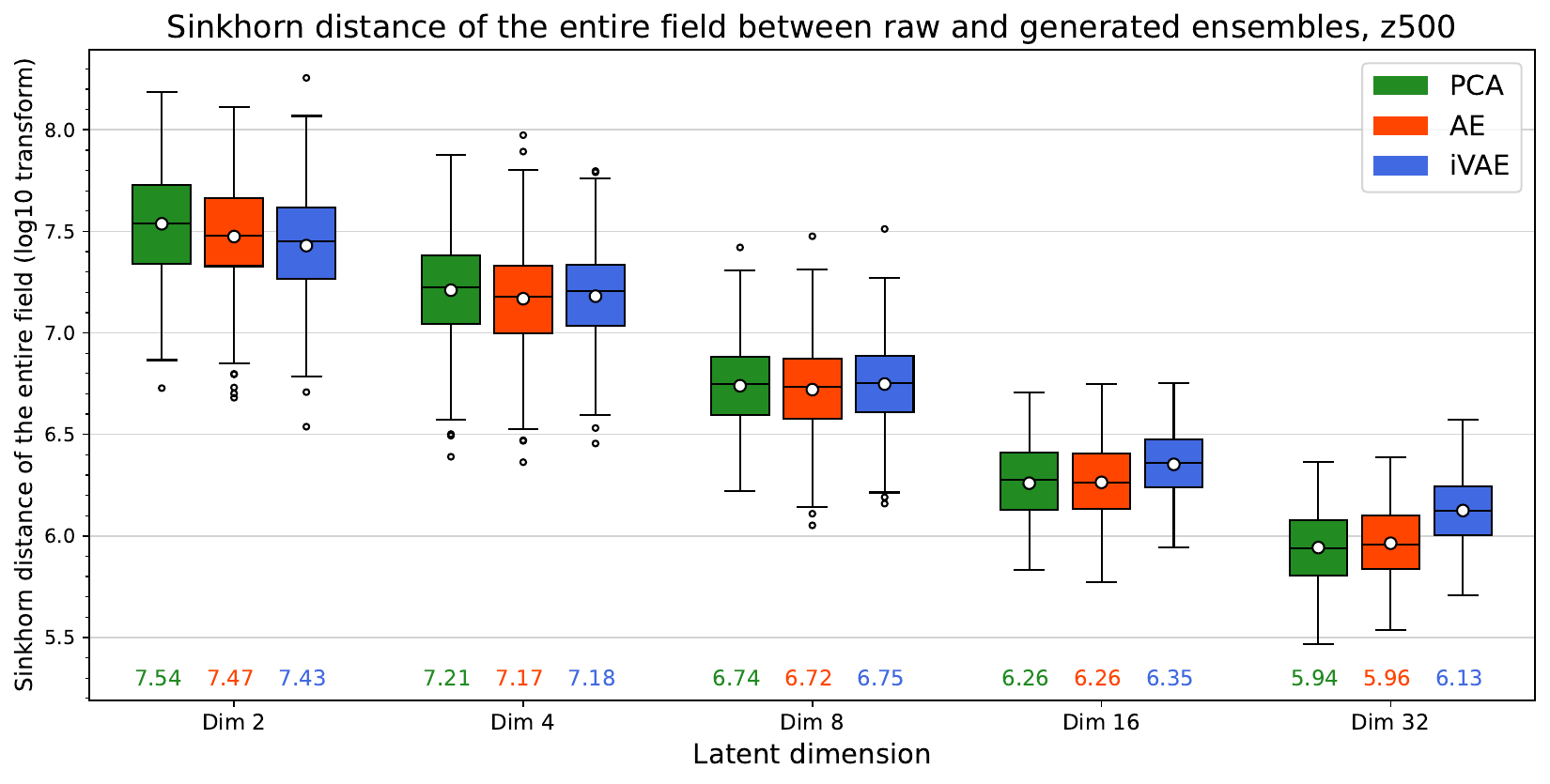}
	\caption{As Figure \ref{fig_wd_t}, but for the geopotential height.}
	\label{fig_wd_z}
\end{figure}

Further evaluation results for two additional weather variables, the V component of 10-m wind speed and the geopotential height at 500 hPa, are presented in Figures \ref{fig_mae_std_box_v}--\ref{fig_wd_box_z}.
In addition to the skill scores, boxplots of the four distance measures for the four weather variables are provided in Figures \ref{fig_ed_t}--\ref{fig_wd_z}.





\begin{figure}
	\centering
	\includegraphics[width=\textwidth]{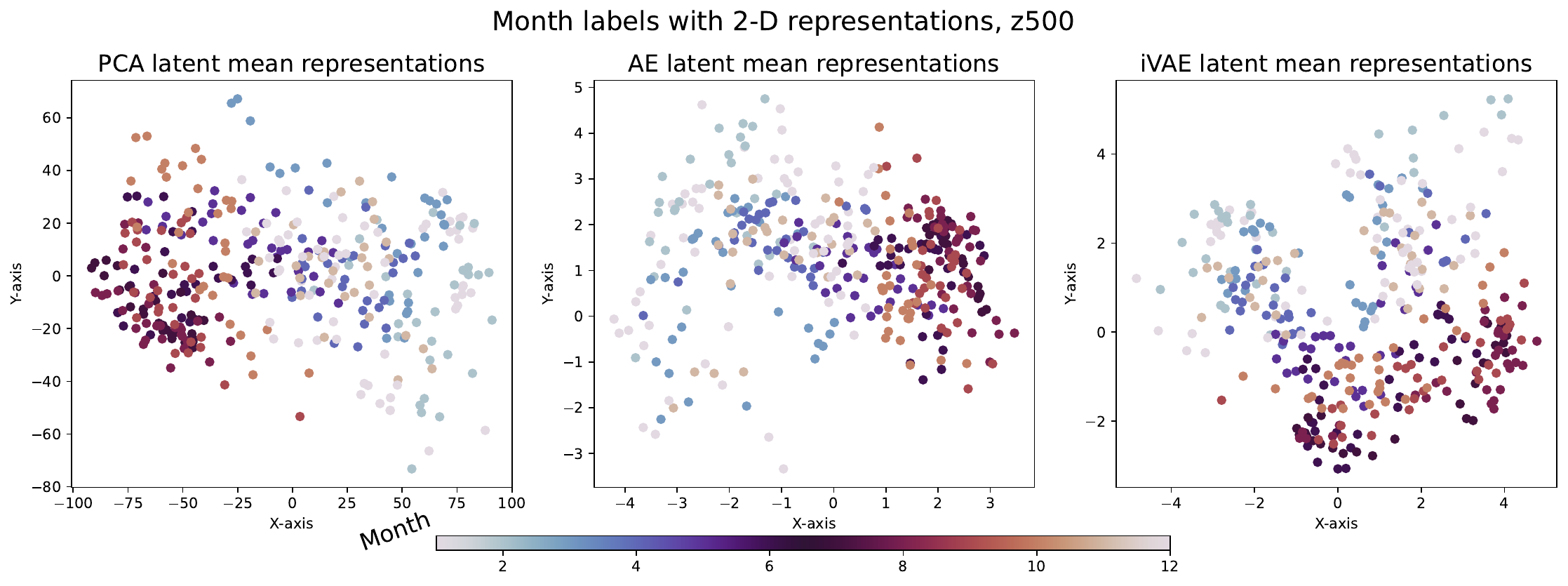}
	\caption{Seasonality visualization of learned 2-D representations from three different methods for geopotential height data, for each day the mean of the latent distribution is plotted, colors indicate month information of the forecast date.}
	\label{fig_scatter_z500}
\end{figure}

\begin{figure}
	\centering
	\includegraphics[width=\textwidth]{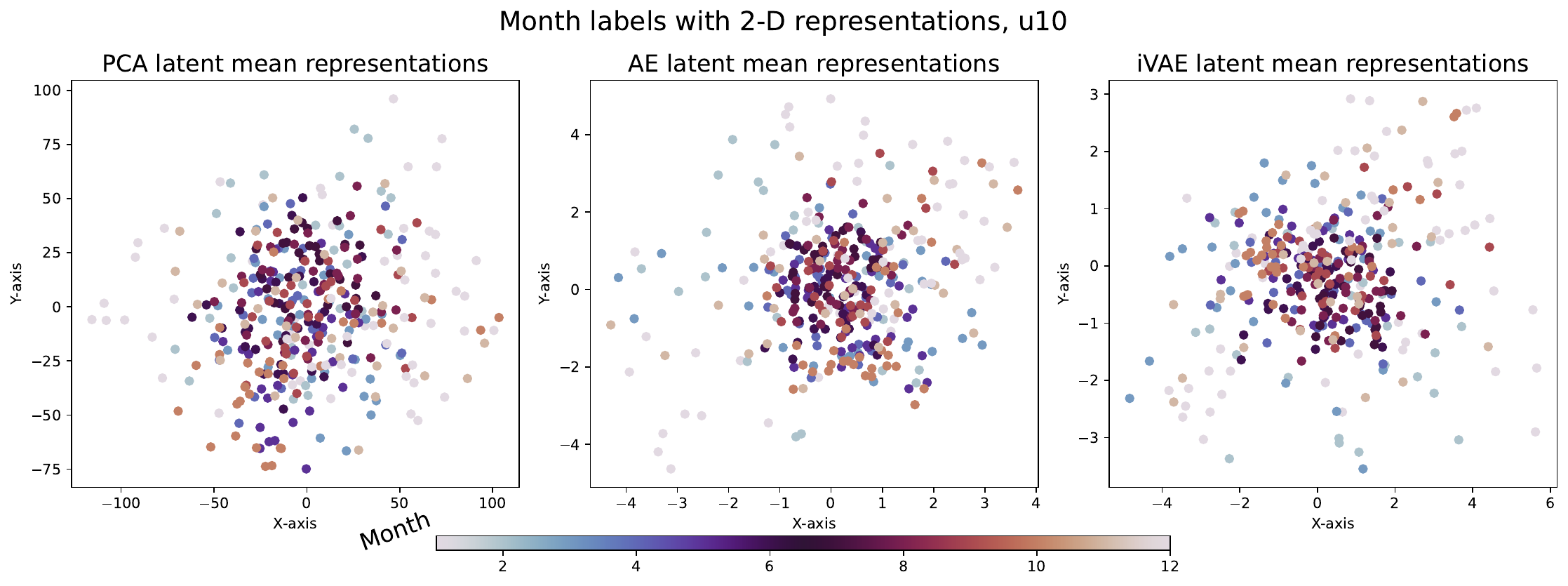}
	\caption{As Figure \ref{fig_scatter_z500}, but for the U component of wind speed.}
	\label{fig_scatter_u10}
\end{figure}

\begin{figure}
	\centering
	\includegraphics[width=\textwidth]{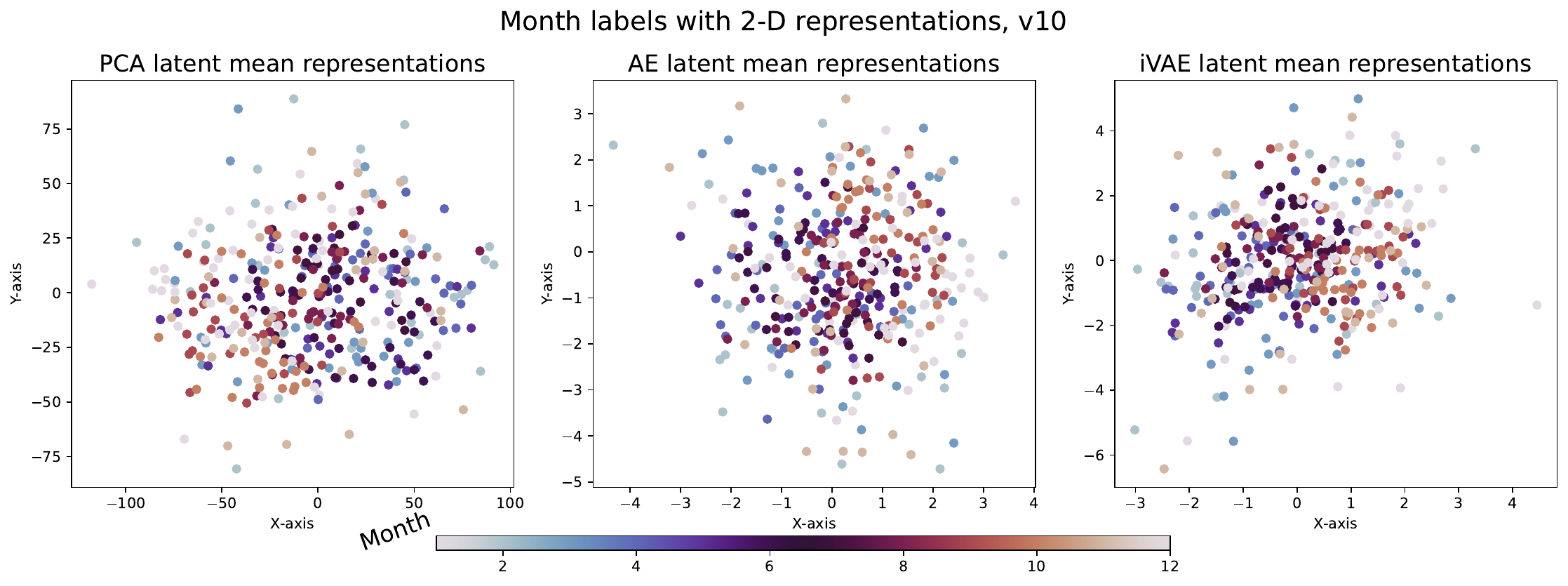}
	\caption{As Figure \ref{fig_scatter_z500}, but for the V component of wind speed.}
	\label{fig_scatter_v10}
\end{figure}

\begin{figure}
	\centering
	\includegraphics[width=0.9\textwidth]{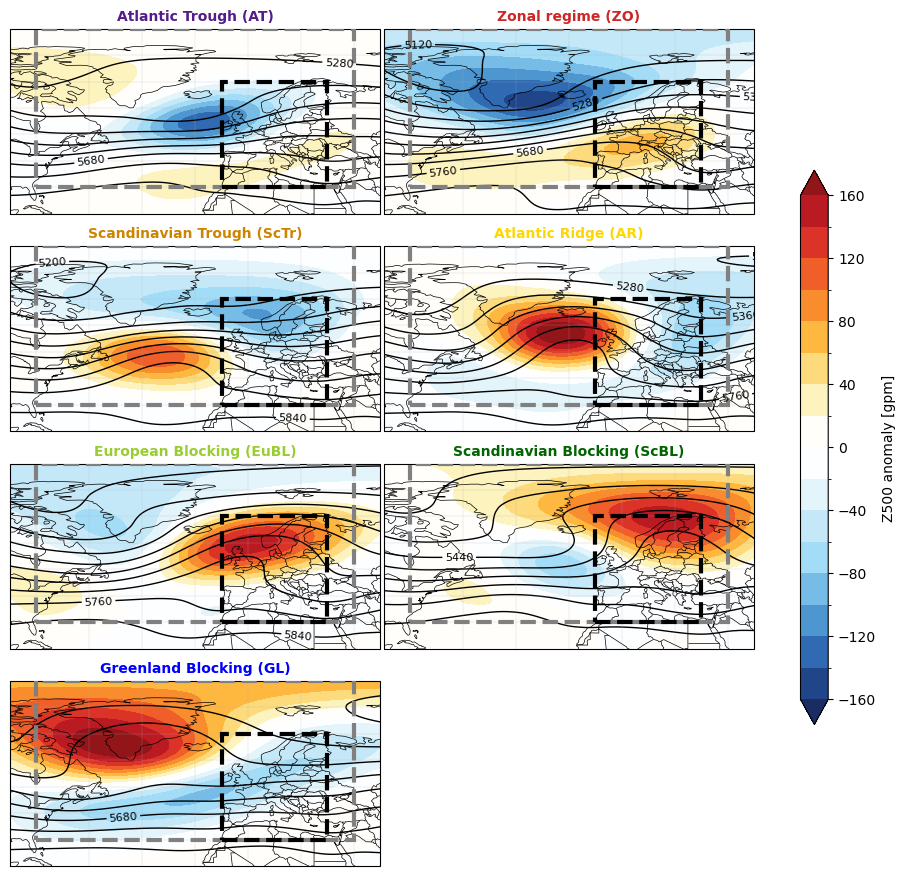}
	\caption{Exemplary fields of geopotential height anomaly at 500 hPa for the 7 weather regimes. The weather regime for a specific day is determined based on the geopotential height anomaly at 500 hPa over the gray dashed square area, while the black dashed square area is considered in this dimensionality reduction study. This figure is adjusted from Figure S1 in the Supplemental Material of \citet{Mockert2024}.}
	\label{fig_wr_def}
\end{figure}

\begin{figure}
	\centering
	\includegraphics[width=\textwidth]{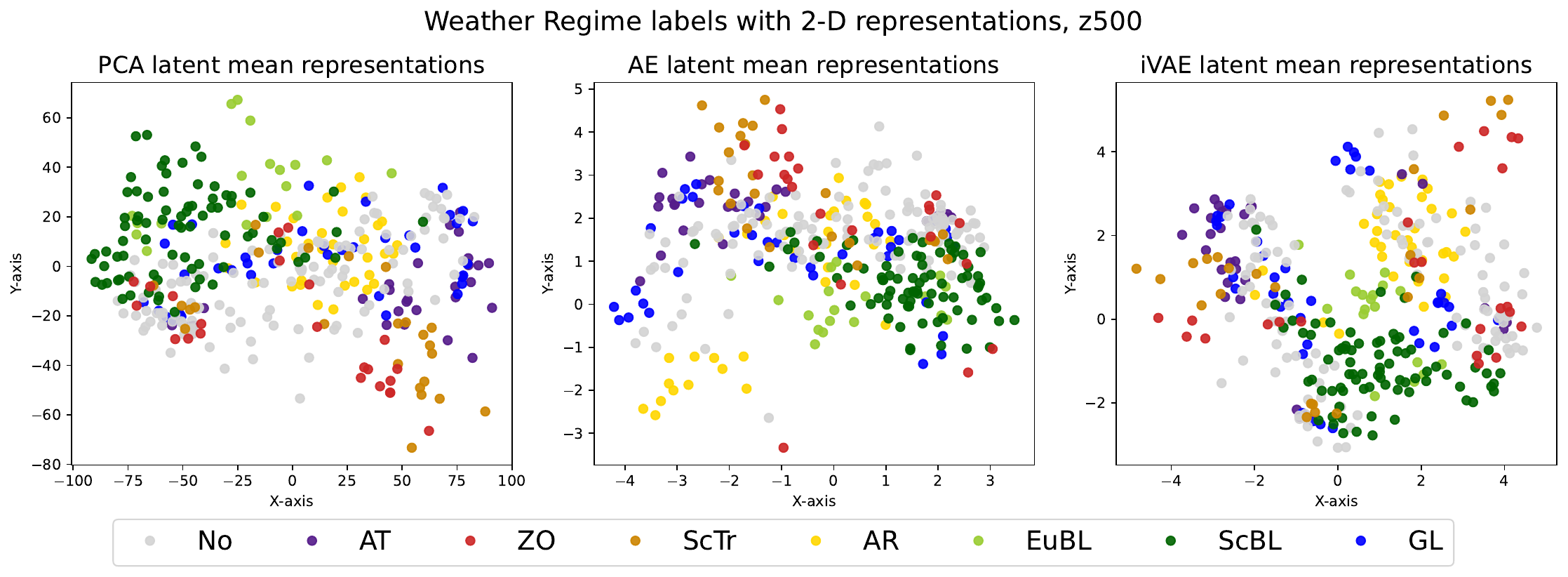}
	\caption{Visualization of learned 2-D representations from three different methods for geopotential height data, labeled and colored by the corresponding weather regime, for each day the mean of the latent distribution is plotted. The gray ``No'' label indicates no dominant weather regime for that specific forecast day.}
	\label{fig_scatter_wr_z500}
\end{figure}

\begin{figure}
	\centering
	\includegraphics[width=\textwidth]{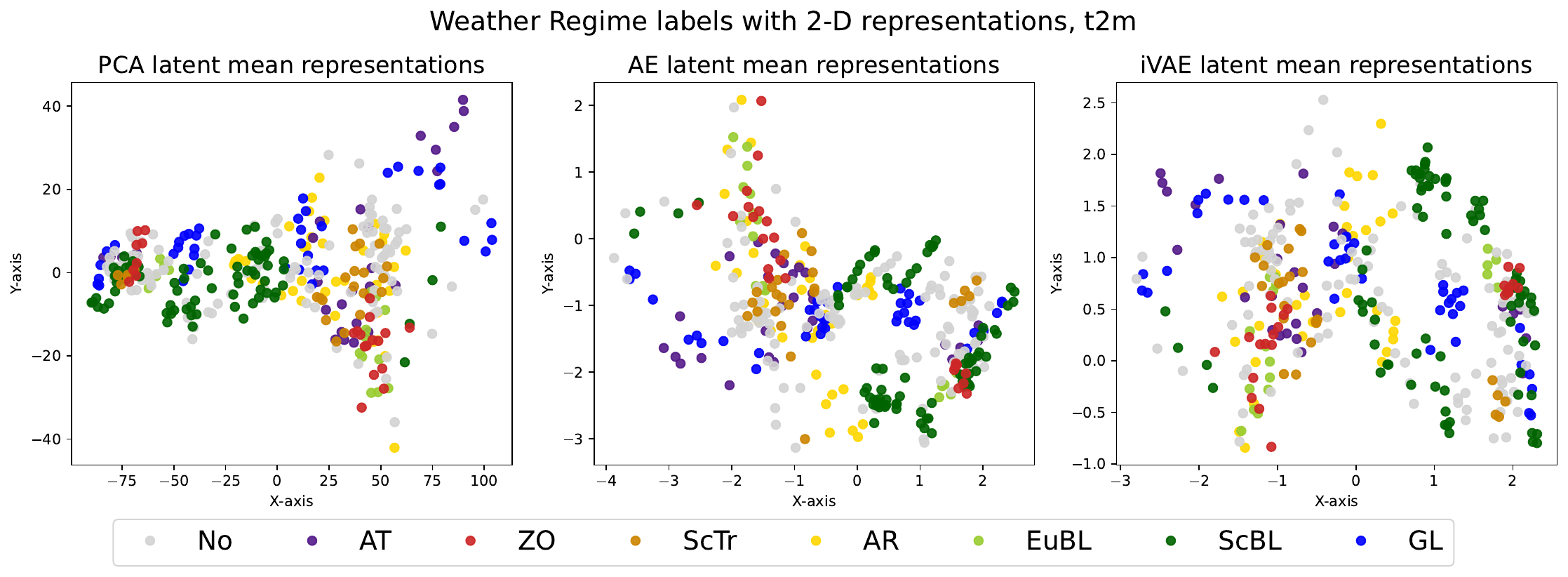}
	\caption{As Figure \ref{fig_scatter_wr_z500}, but for temperature.}
	\label{fig_scatter_wr_t2m}
\end{figure}

\begin{figure}
	\centering
	\includegraphics[width=\textwidth]{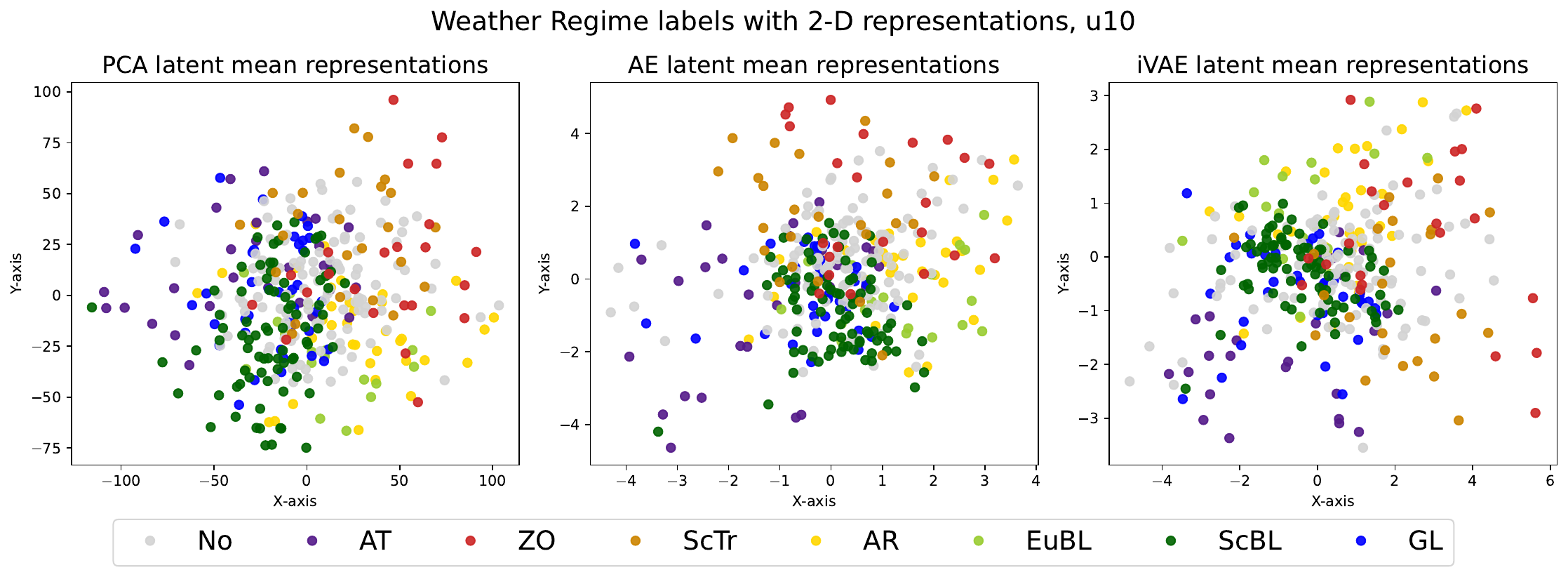}
	\caption{As Figure \ref{fig_scatter_wr_z500}, but for the U component of wind speed.}
	\label{fig_scatter_wr_u10}
\end{figure}

\begin{figure}
	\centering
	\includegraphics[width=\textwidth]{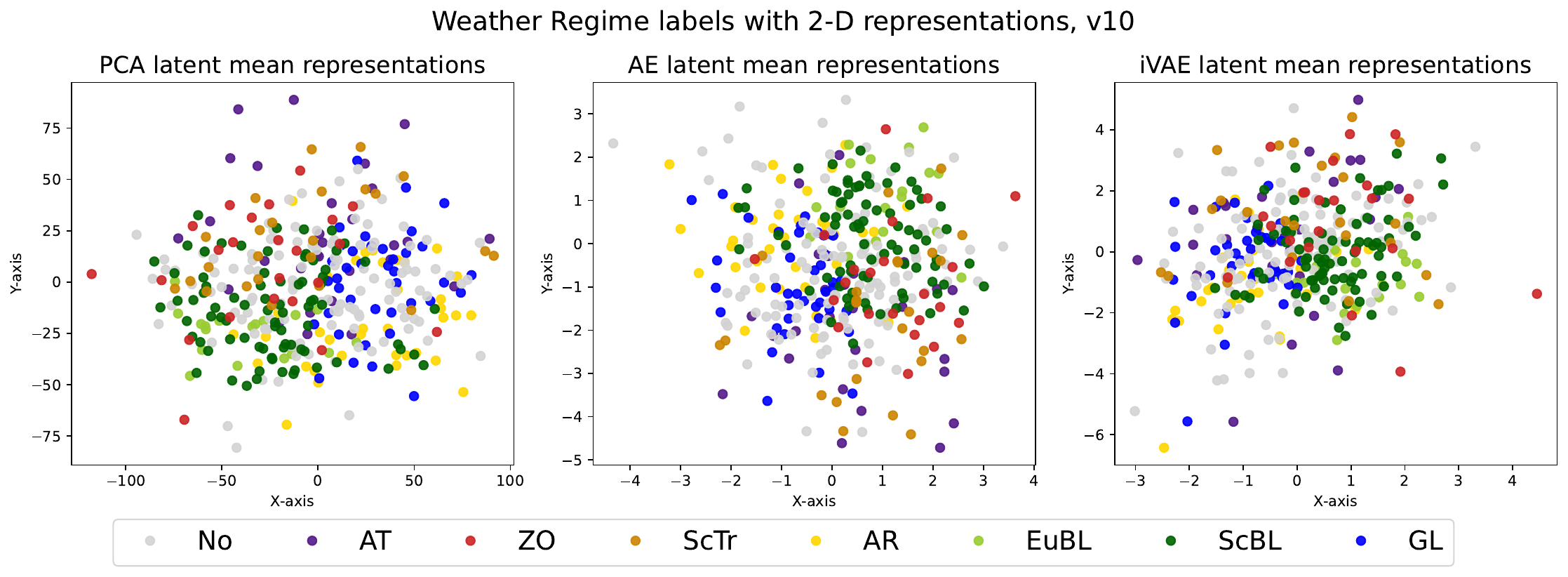}
	\caption{As Figure \ref{fig_scatter_wr_z500}, but for the V component of wind speed.}
	\label{fig_scatter_wr_v10}
\end{figure}

The visualization of connections between the month information of the forecast date and two-dimensional mean representations learned by all three dimensionality reduction approaches, for geopotential height and the U and V components of wind speed, is provided in Figures \ref{fig_scatter_z500}--\ref{fig_scatter_v10}.
The visualization of connections between the specific weather regime at the forecast date and the learned two-dimensional mean representations is provided in Figures \ref{fig_scatter_wr_z500}--\ref{fig_scatter_wr_v10}.
In Figure \ref{fig_wr_def}, we show examples of exemplary fields of geopotential height anomaly at 500 hPa used to determine the weather regime, where the 7 weather regimes are defined by \citet{grams2017balancing}.



\begin{figure}
	\centering
	\includegraphics[width=\textwidth]{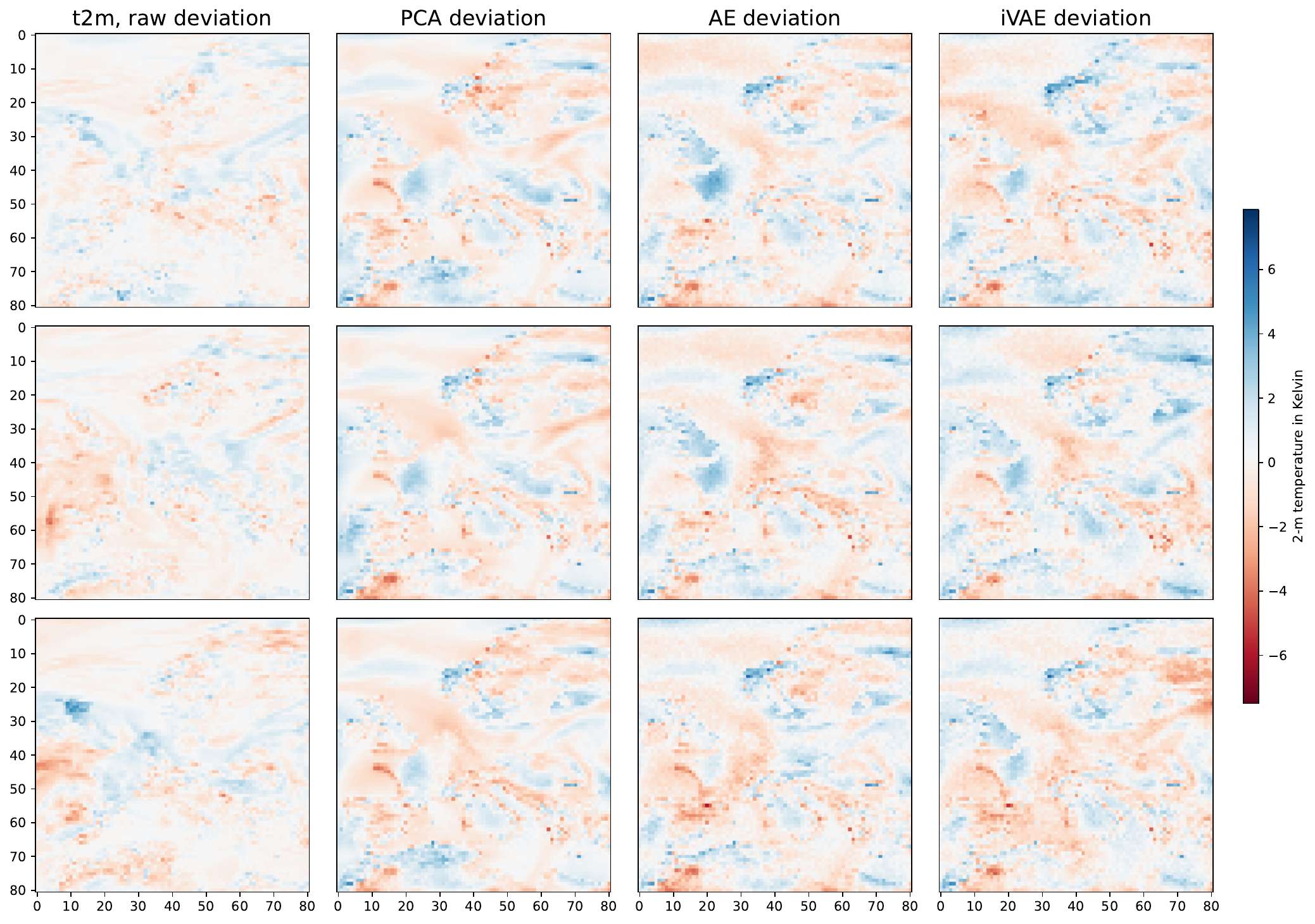}
	\includegraphics[width=\textwidth]{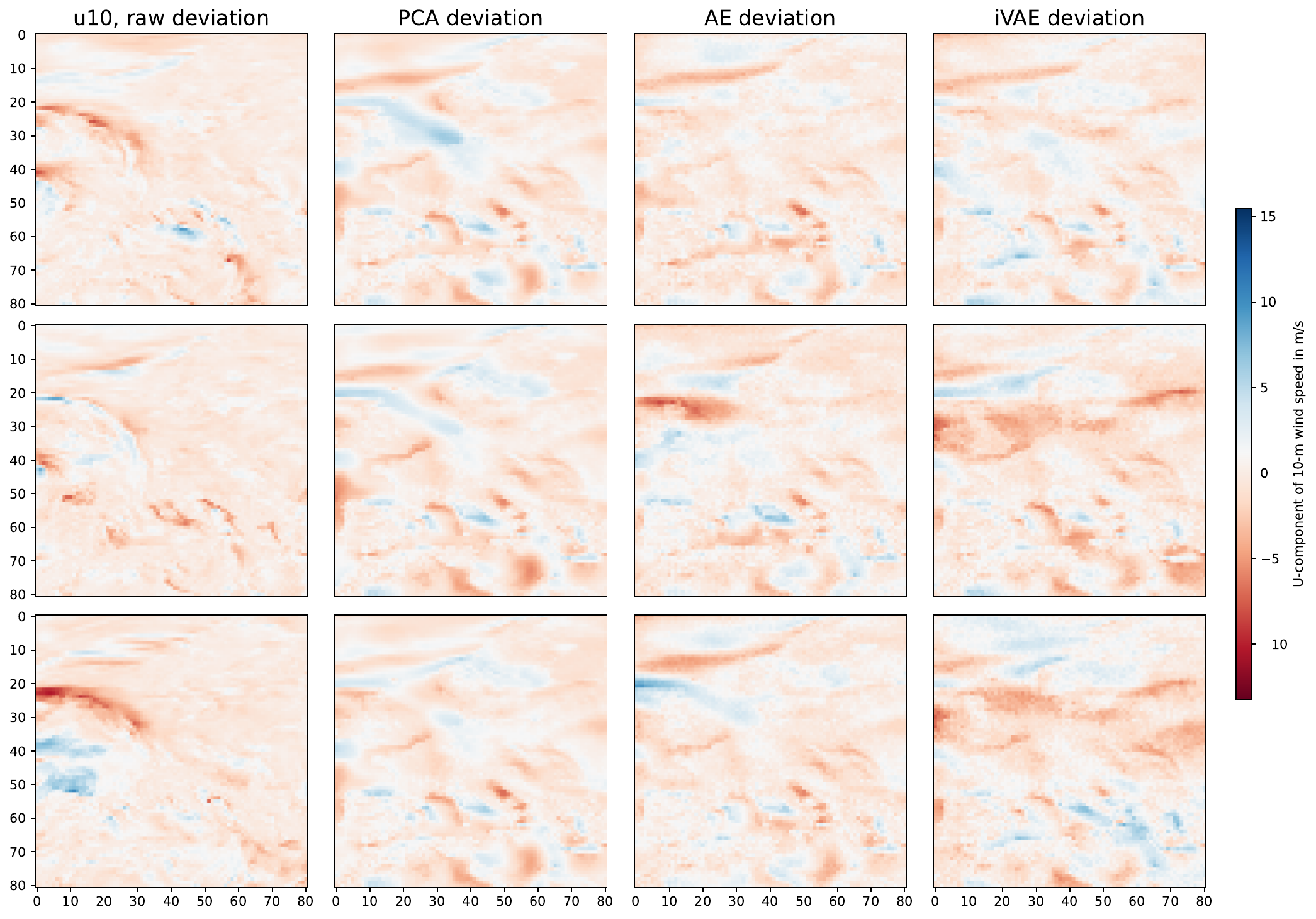}
	\caption{Exemplary raw forecast fields and reconstructed forecast fields of 2-m temperature (top) and the U component of 10-m wind speed (bottom) by different methods, showing deviation of each ensemble member to the raw ensemble mean, with a latent dimension of 32. The rows correspond to different ensemble members for the same forecast day.}
	\label{fig_example_anomaly}
\end{figure}

Figure \ref{fig_example_anomaly} shows exemplary ensemble members of the deviation from the raw ensemble mean with respect to the exemplary figure (Figure 3) in the main text.

\end{document}